\documentclass[]{sensenova}

\usepackage{hyperref}
\usepackage{cleveref}
\usepackage{verbatim}

\usepackage{wrapfig}
\usepackage{graphicx}
\usepackage{floatrow}
\usepackage{subcaption}
\usepackage{listings}
\usepackage{algorithm}
\usepackage{marvosym}

\usepackage{mmstyles}
\usepackage[toc,page,header]{appendix}

\title{Vision as Unified Multimodal Generation}

\author{%
Xiaoyang Han$^{*,1}$, Jianhua Li$^{*,1}$, Kewang Deng$^{*,1}$,
Zukai Chen$^{*,1}$, Xuanke Shi$^{*,1}$, Sihan Wang$^{*,1}$ \\
Boxuan Li$^{*,1}$, Linyan Wang$^{*,1}$, Siyi Xie$^{1,4}$, Xin You$^{1,5}$,
Jinsheng Quan$^{1,6}$, Zhongang Cai$^1$ \\
Haiwen Diao$^{2}$, Ziwei Liu$^{\textrm{\Letter},2}$,
Lei Yang$^{\textrm{\Letter},1}$,
Dahua Lin$^{\textrm{\Letter},1,3}$, Quan Wang$^{*,\textrm{\Letter},1}$ \\[1pt]
\parbox{\textwidth}{\centering\small
    $*$ Core Contributors,
    $\textrm{\Letter}$ Corresponding Authors \\
    $^1$SenseTime Research,
    $^2$Nanyang Technological University,
    $^3$The Chinese University of Hong Kong \\
    $^4$Peking University,
    $^5$Shanghai Jiao Tong University,
    $^6$Zhejiang University
}
}

\abstract{

We formulate computer vision as unified multimodal generation, where heterogeneous visual tasks are expressed through the native text and image generation spaces of a unified multimodal model (UMM), without task-specific architectures.
With this formulation, the single model \textbf{SenseNova-Vision} matches leading task-specialized systems across structured visual understanding, dense geometric prediction, segmentation, and multi-view visual geometry.
Natural-language instructions and optional visual prompts specify the task, target regions or views, and decoding convention.
Responses are then generated as text for symbolic records, images for dense spatial targets, or mixed outputs for compositional tasks.
To enable large-scale training, we convert heterogeneous computer vision annotations into instruction-response examples compatible with these native generation spaces.
This conversion yields the \textbf{SenseNova-Vision Corpus}, a computer-vision instruction-response corpus spanning text, image, and mixed text-and-image targets.
Starting from an off-the-shelf pretrained UMM, SenseNova-Vision is trained primarily on the SenseNova-Vision Corpus, using auxiliary multimodal data as a capability-preserving mixture and requiring no task-specific prediction heads or architectural changes.
The resulting model covers detection, OCR, keypoints, segmentation, depth, surface normals, point maps, and camera pose estimation, and can follow language-defined variants that combine category, color, region, and other visual cues.
These results suggest unified multimodal generation as a scalable route for integrating computer vision into general-purpose foundation models.
The SenseNova-Vision model and SenseNova-Vision Corpus are publicly available.

}

\checkdata[Codebase]{\url{https://github.com/OpenSenseNova/SenseNova-Vision}}
\checkdata[Model and Dataset]{\url{https://huggingface.co/collections/sensenova/sensenova-vision}}

\begin{document}
\maketitle

\begin{center}
    \vspace{-1pt}
    \captionsetup{type=figure,hypcap=false,skip=2pt}
    \includegraphics[width=\textwidth]{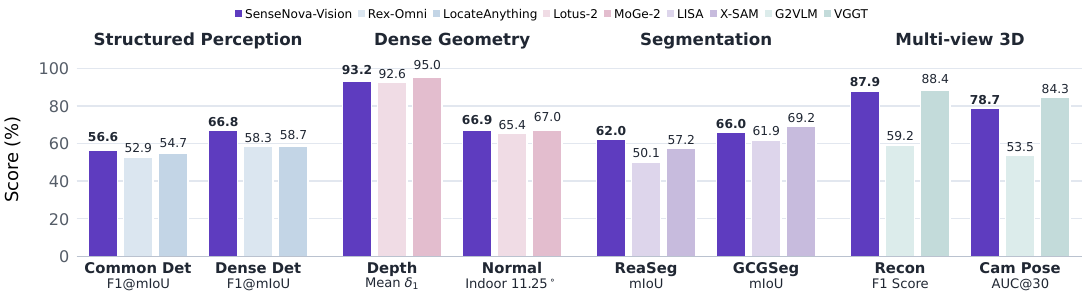}
    \caption{
        Benchmark overview across representative computer vision task families.
        Despite using a single unified multimodal generation interface and no task-specific heads, SenseNova-Vision achieves competitive performance across heterogeneous output formats, including text-serialized records, dense image outputs, mixed text-mask responses, and multi-view geometric predictions.
    }
    \label{fig:benchmark-overview}
    \vspace{-2pt}
\end{center}

\section{Introduction}
\label{sec:intro}

\begin{figure}[t]
    \centering
    \includegraphics[width=\textwidth]{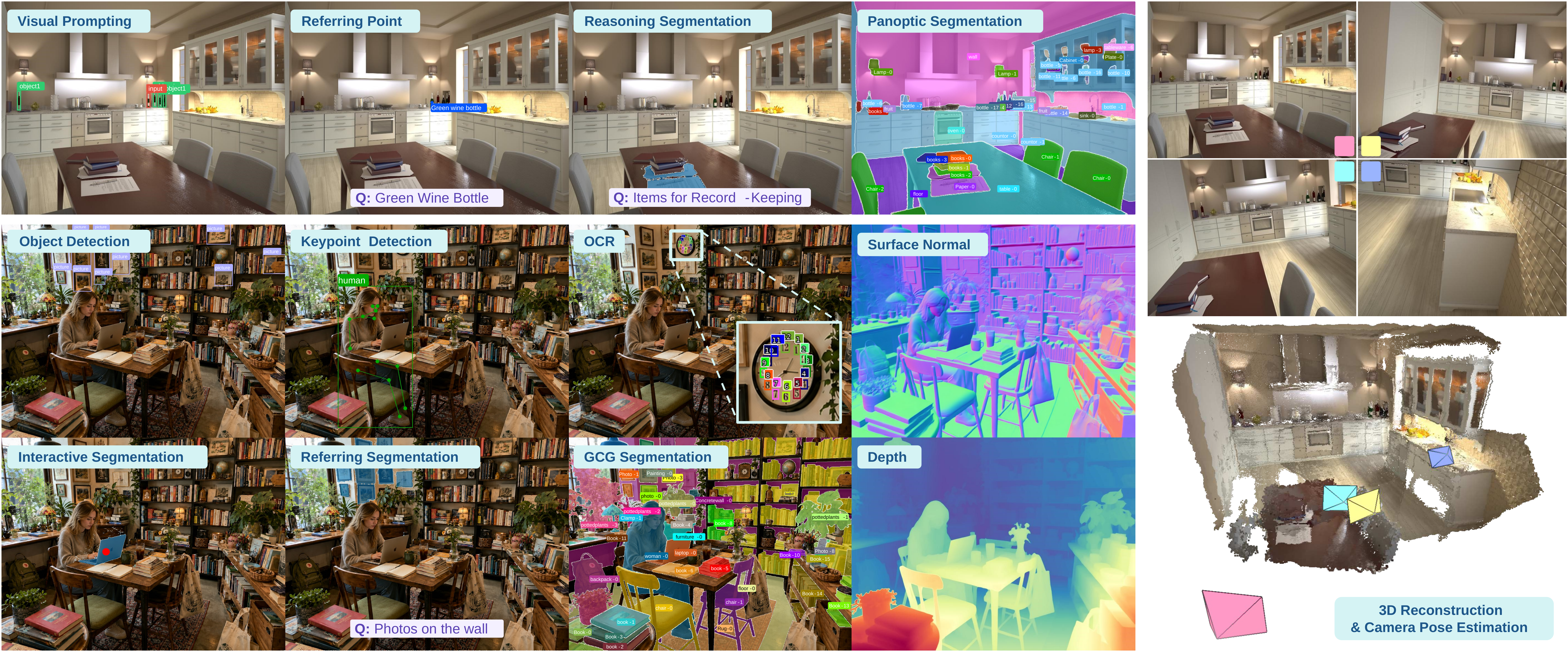}
    \caption{
        SenseNova-Vision integrates diverse computer vision tasks into a single UMM, producing outputs for structured visual understanding, dense geometric prediction, segmentation, and multi-view visual geometry through unified multimodal generation.
    }
    \label{fig:qualitative-overview}
\end{figure}

\begin{figure}[t]
    \centering
    \includegraphics[width=\textwidth]{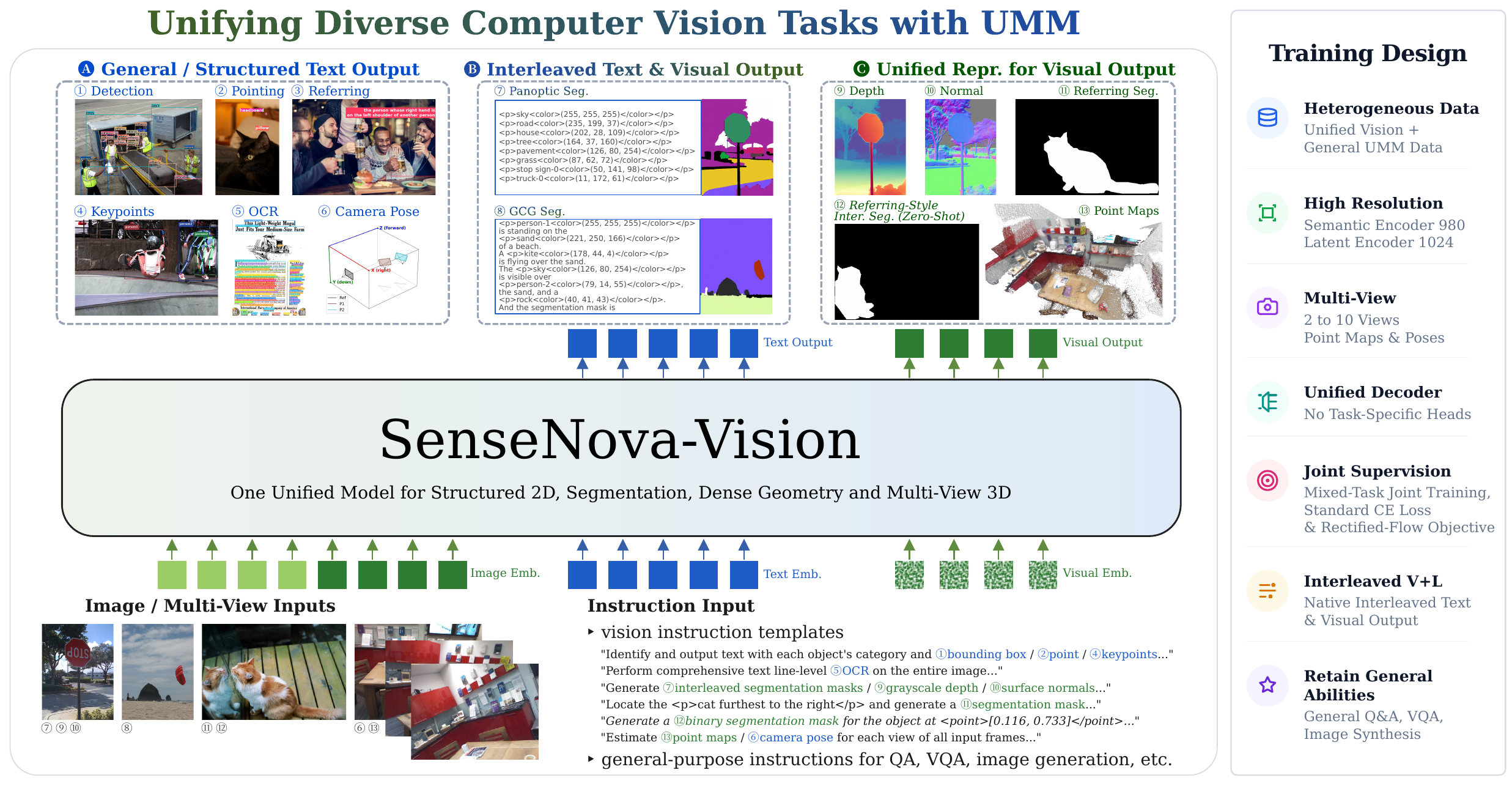}
    \caption{
        Overview of SenseNova-Vision: heterogeneous computer vision annotations are converted into native text, image, and mixed text-image generation targets for joint training in a single UMM, without task-specific heads.
    }
    \label{fig:system-overview}
\end{figure}

Large language models~\cite{gpt3} have shown that diverse language tasks can be consolidated through prompting and generation, and unified multimodal models (UMMs)~\cite{chameleon,showo,janus,bagel,sensenovau1} further extend this paradigm to both text and image generation.
We consider whether the entire spectrum of classical computer vision---from detection to multi-view 3D---can be expressed as unified multimodal generation within a single UMM, without task-specific heads.
Computer vision has made remarkable progress through specialist systems~\cite{detr,maskrcnn,sam,raft,depthanything,vggt} tailored to individual task families.
Their outputs range from boxes and masks to motion fields, dense maps, and 3D geometry, each typically paired with task-specific architectures, losses, decoding rules, and evaluation protocols.
This task-specific organization makes visual supervision difficult to share, reuse, and compose across tasks, motivating a shared and scalable formulation for computer vision.

Prior efforts approach this goal in complementary ways: sequence-format unifications~\cite{ofa,pix2seqv2,unifiedio2} bring diverse annotations into shared model interfaces, but different output types still require serialization and parsing conventions; representation-centric models~\cite{sam,seggpt,depthanything,vggt} generalize within visually coherent task families yet offer no unified output space and limited language control.
In parallel, generative foundation models offer a different substrate: diffusion and image-generation models~\cite{ddpm,latentdiffusion} provide visual generative priors for spatially aligned outputs, while multimodal large language models (MLLMs)~\cite{flamingo,llava} bring language instructions and reasoning into visual perception.
Recent perception work adopts these strengths only partially: image-generative methods~\cite{lotus,diception,visionbanana} handle dense maps but struggle with symbolic records, while MLLM-based systems~\cite{lisa,visionllmv2,xsam} add language and reasoning yet still route dense outputs through task-specific decoders.
Across these routes, symbolic records, dense spatial targets, and mixed outputs are still not expressed within one native multimodal generation framework.

To address this gap, we introduce SenseNova-Vision, built on a simple formulation: heterogeneous computer vision tasks can be cast as unified multimodal generation.
Natural-language instructions specify the task, target, output schema, and decoding convention, while text generation expresses symbolic visual answers such as categories, spatial references, OCR strings, and camera parameters.
Image generation is natural for dense prediction because masks, depth maps, surface normals, and point maps are spatially aligned with the input image and can be represented on the same grid.
Together, text and image generation provide complementary response spaces that can express a broad range of computer vision tasks.
This formulation is, for visual perception, the analogue of what GPT~\cite{gpt3} did for NLP tasks: consolidating heterogeneous specialist supervision into a single generative interface.

Making this formulation trainable at scale requires converting heterogeneous computer vision annotations into instruction-response examples.
We therefore construct the SenseNova-Vision Corpus, spanning structured visual understanding, dense geometric prediction, segmentation, and multi-view visual geometry with decodable text, image, and mixed targets that can be recovered as boxes, masks, depth maps, surface normals, point maps, or camera poses.
Built from the off-the-shelf UMM Bagel~\cite{bagel}, SenseNova-Vision is trained on this corpus to express diverse visual task outputs through the unified multimodal generation interface illustrated in Figure~\ref{fig:system-overview}.
Figure~\ref{fig:qualitative-overview} and Figure~\ref{fig:benchmark-overview} show representative qualitative outputs and quantitative results across four perception families, respectively.
In quantitative evaluations, SenseNova-Vision leads on structured visual understanding, remains close to the strongest dense geometric prediction and segmentation baselines, and approaches the leading specialist in multi-view visual geometry.
This flexibility further allows the same model to follow language-defined task variants that are not explicitly enumerated in the training corpus, combining category, color, region, and other visual cues.

SenseNova-Vision is built on a simple premise: computer vision can become a native generative capability of unified foundation models, moving beyond a collection of isolated task-specific systems.
First, we introduce a unified multimodal generation formulation that casts heterogeneous vision tasks into native input-output spaces of UMMs.
Second, we construct the SenseNova-Vision Corpus, a large-scale computer-vision instruction-response corpus with decodable text, image, and mixed targets.
Third, we train SenseNova-Vision, showing strong results across four perception families and supporting language-defined task variants beyond the training set.
Together, this work points toward a new era of computer vision, where perception is no longer engineered task by task, but becomes a programmable, generative, and extensible capability of unified foundation models, providing a reusable basis for future research.

\FloatBarrier

\section{Related Works}
\label{sec:related_works}

\subsection{Early Unified Vision Models and Interfaces}

Early unified vision models explore how diverse vision and vision-language tasks can be expressed through shared task formats.
This line begins with casting structured visual predictions as sequences, as in Pix2Seq for object detection~\cite{pix2seq}, and then extends to broader task families in Pix2Seq v2~\cite{pix2seqv2} and text-box generation in UniTAB~\cite{unitab}.
OFA~\cite{ofa} and Uni-Perceiver~\cite{uniperceiver,uniperceiverv2} further expand this idea to cross-modal and generalist vision-language modeling.
Unified-IO~\cite{unifiedio} and Unified-IO 2~\cite{unifiedio2} broaden the direction by representing heterogeneous inputs and outputs in a common token space, while Florence-2~\cite{florence2} scales prompt-based sequence generation to a wide range of vision tasks.
Yet their unification still relies heavily on shared sequence formats and task-specific encoding rules, making the interface less natural for dense maps and structurally diverse visual outputs.

\subsection{Task-Family Generalization with Visual Foundation Models}

Another line of work seeks generalization with visual foundation models and pretrained visual representations.
Representative pretraining methods such as MAE~\cite{mae} and DINOv2~\cite{dinov2} provide reusable visual features for recognition, dense prediction, and geometry transfer.
SAM~\cite{sam} shows that promptable segmentation can cover a broad family of mask prediction tasks, while Painter~\cite{painter} and SegGPT~\cite{seggpt} use visual in-context examples to specify different image-to-image perception tasks.
In geometry, Depth Anything~\cite{depthanything} scales depth prediction with strong visual representations and large data, and VGGT~\cite{vggt} extends feed-forward visual modeling to cameras, point maps, reconstruction, and point tracking.
These works generalize strongly within visually coherent task families, but the absence of a unified output space keeps their unification partial.
They also provide limited support for understanding complex, compositional, or open-ended language instructions.

\subsection{Image-Generative Models for Dense Perception}

Diffusion and image generation models provide a generative alternative to conventional dense prediction by producing spatially aligned visual outputs.
Marigold~\cite{ke2024marigold} adapts diffusion priors to monocular depth estimation, while Lotus, Lotus-2, and FE2E~\cite{lotus,lotus2,fe2e2025editor} adapt powerful image generative or editing models for geometric dense prediction such as depth and normal estimation.
DICEPTION~\cite{diception} casts multiple perceptual tasks as conditional image generation in a shared RGB output space, and Visual Bridge~\cite{visualbridge} studies generative visual perception representations.
Vision Banana~\cite{visionbanana} further scales this direction with a stronger image-generation foundation model and lightweight instruction tuning, enabling richer language-conditioned visual predictions while preserving generation capability.
These methods make dense targets compatible with the native image space of generative models, but their unification remains largely image-side.
Since symbolic records, task schemas, camera parameters, and decoding rules are naturally textual, image generation alone cannot cover the symbolic and mixed outputs required by many tasks.

\subsection{MLLM-based Perception and Task Modules}

As powerful foundation models that integrate language and vision, MLLMs bring new opportunities to visual perception.
One line extends MLLMs to structured visual grounding: Kosmos-2~\cite{kosmos2}, Shikra~\cite{shikra}, and Ferret~\cite{ferret} connect language generation with boxes, coordinates, and regions, while Rex-Omni~\cite{rexomni} broadens this direction through next-point prediction.
Another line represents dense spatial outputs through discrete token or logit spaces, as in Text4Seg~\cite{lan2025text4seg} and DenseMLLM~\cite{densemllm}, but preserving fine spatial details requires carefully designed tokenization and decoding schemes.
Decoder-based systems instead attach task-specific modules to MLLMs: LISA~\cite{lisa}, PixelLM~\cite{pixellm}, and X-SAM~\cite{xsam} connect language reasoning to segmentation or mask decoders, while G$^2$VLM~\cite{g2vlm} and VisionLLM v2~\cite{visionllmv2} introduce downstream modules or connections for geometry and broader perception tasks.
This preserves useful inductive biases but keeps output spaces fragmented across task-specific components.

Recent UMMs~\cite{chameleon,showo,janus,bagel,sensenovau1} extend multimodal foundation models from understanding-centered MLLMs to models that natively support both text and image generation.
They typically model language and visual content within shared or coupled generative spaces, allowing a single model to understand instructions, generate text, and synthesize images.
This makes it timely to revisit the unification of computer vision tasks: heterogeneous visual supervision can be expressed as decodable targets in the same native spaces of a UMM.
SenseNova-Vision follows this route by aligning symbolic records, dense spatial targets, and mixed responses with the native text-and-image input-output spaces of a UMM, turning them into decodable training targets for a single model.

Complementary to this line of work, SenseNova-SI \cite{cai2026scaling} studies how curated spatial-intelligence supervision can scale spatial reasoning abilities in multimodal foundation models including UMM backbones. Our work builds on the same broader view that spatial and geometric abilities can be cultivated in foundation models, but focuses on a different question: how heterogeneous computer vision annotations can be converted into native text, image, and mixed generation targets for benchmark-decodable perception outputs.
\section{Data}
\label{sec:data}

Training a UMM on unified computer vision tasks requires casting heterogeneous annotations into the model's native text-and-image generation spaces.
We therefore convert each source annotation into an instruction-response example with one or more visual inputs, a language instruction, and a decodable target response.
This conversion is organized through a data protocol covering four computer vision families: structured visual understanding, dense geometric prediction, segmentation, and multi-view visual geometry.
Sec.~\ref{sec:data_protocol} details the protocol for each family.

\begin{figure}[!tp]
    \centering
    \includegraphics[
        width=\textwidth,
        trim=20 355 18 16,
        clip
    ]{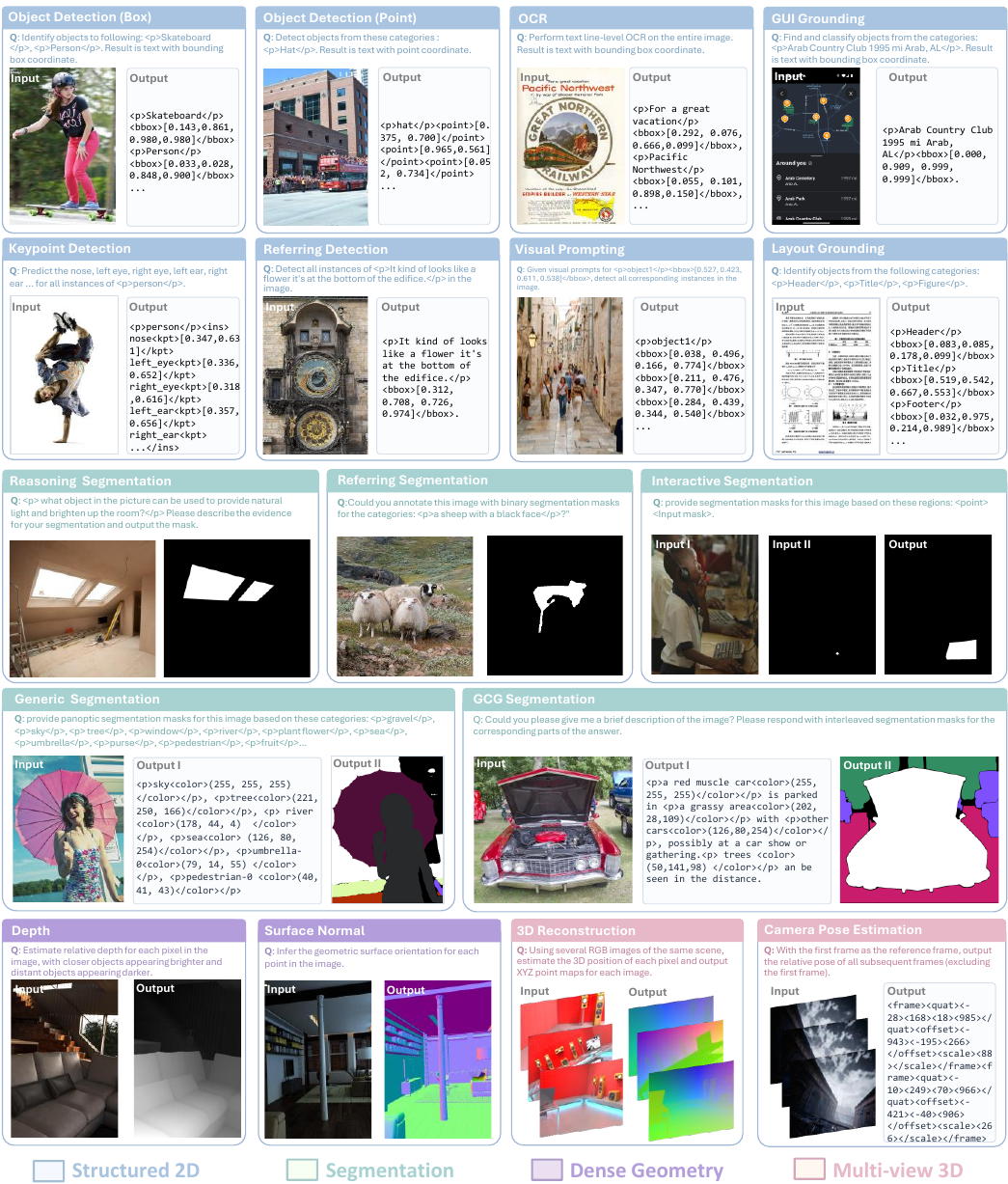}
    \caption{
        Representative SN-VC examples under the data protocol.
        Source annotations from structured visual understanding, dense geometric prediction, segmentation, and multi-view visual geometry are converted into instruction-conditioned text, image, or mixed text-image targets.
    }
    \label{fig:data-examples}
\end{figure}

Following this protocol, we construct the \textbf{SenseNova-Vision Corpus (SN-VC)}, a large-scale computer vision corpus built from public images, with its source composition summarized in Fig.~\ref{fig:snvc-source-composition}.
Available public annotations are directly converted when possible, and additional targets are generated or curated to handle incomplete supervision while improving data diversity.
We release the generated and curated subset as \textbf{SN-VC-50M}, a 50-million-example collection of converted computer vision supervision, and provide source lists, prompt templates, conversion rules, and examples to reproduce the remaining public-source portion of SN-VC, as described in Sec.~\ref{sec:dataset_construction}.

\subsection{Data Protocol}
\label{sec:data_protocol}

The protocol defines a common sample schema for UMM training: one or more visual inputs, a natural-language instruction, and a decodable target response.
The instruction specifies the task intent, output schema, and decoding convention, while the response is represented as text, an image, or a mixed text-image output that can be recovered as benchmark-compatible labels, coordinates, masks, dense maps, or camera parameters.
Fig.~\ref{fig:data-examples} illustrates representative examples produced by this protocol across the four families.
We use lightweight textual markers for structured and segmentation responses, and reserved special tokens for camera-pose records; Appendix Tables~\ref{tab:structure_markers}, \ref{tab:segment_markers}, and~\ref{tab:pose_token_markers} summarize these conventions.

\textbf{Structured visual understanding.}
This family covers structured visual understanding tasks whose outputs can be decoded as sparse symbolic records, including detection, referring, pointing, keypoint localization, OCR, layout understanding, and GUI grounding~\cite{rexomni}.
We represent these annotations as text generation targets: labels, transcripts, and task-specific attributes remain ordinary text, while spatial fields are written as normalized image coordinates.
Lightweight markers such as \texttt{<p>}, \texttt{<bbox>}, and \texttt{<point>} delimit phrases and coordinate fields.
The generated response is parsed back into typed benchmark records, so detection, grounding, OCR, pointing, keypointing, layout, and GUI tasks share the same text-generation space while being separated by language instructions and textual schemas.

\textbf{Dense geometric prediction.}
This family covers dense, pixel-aligned geometric targets, including depth maps and surface normals.
These targets share the spatial layout of the input image, making conditional image generation a natural representation.
The instruction specifies the requested geometric quantity and decoding convention, while the response image stores the corresponding dense signal in a deterministic visual encoding.
For depth, valid metric values are converted to inverse depth and rendered as normalized grayscale images; surface normals are rendered as RGB maps whose channels encode the normal components.
Generated images can then be decoded back into metric-compatible depth or normal maps for evaluation.

\textbf{Segmentation.}
Following the task taxonomy of X-SAM~\cite{xsam}, this family covers single-target and multi-region segmentation.
Segmentation combines semantic region selection with pixel-level spatial prediction, and we choose the response format according to whether the instruction asks for one target or multiple regions.
For single-target tasks such as referring, reasoning, and interactive segmentation, the instruction identifies the target region and the response is a binary mask image with fixed foreground and background colors.
Interactive segmentation additionally provides visual prompts, such as points, boxes, scribbles, or masks, together with the input image.
For multi-region tasks, such as generic segmentation (semantic and panoptic segmentation), we use a mixed text-image response: the text component lists regions and uses the \texttt{<color>} marker to specify RGB palette values in prompts or generated legends, while the image component renders the corresponding color-coded mask.
Grounded conversation generation (GCG) segmentation further exercises this format: given an open-ended instruction, the model first produces region descriptions and color assignments, then renders the mask according to the generated legend.
This design lets language handle category names, referring expressions, reasoning-derived regions, and generated region descriptions, while the image channel preserves dense pixel-level supervision.

\textbf{Multi-view visual geometry.}
This family follows feed-forward visual geometry settings such as VGGT~\cite{vggt} and uses an ordered image set as visual input.
The instruction specifies the view order, reference coordinate frame, and requested output types.
Dense scene geometry is represented as image outputs in the form of per-view dense XYZ point maps; each point map stores aligned and normalized 3D coordinates in its RGB channels.
Camera pose outputs are represented as structured sequences.
Each target view is encoded relative to the reference frame as a quaternion rotation, a translation direction, and a scale, with reserved tokens such as \texttt{<frame>}, \texttt{<quat>}, \texttt{<offset>}, and \texttt{<scale>} delimiting the boundaries of view entries and pose fields.
This mixed response format keeps dense point-map reconstruction in the image space and discrete geometric metadata in the text space.

\subsection{Corpus Construction}
\label{sec:dataset_construction}

Following the protocol above, we construct SN-VC by converting public computer vision datasets into instruction-response examples.
SN-VC is intended as the full reproducible corpus: images are drawn from public sources, and available annotations are converted directly when they already match a decodable target format.
When supervision is incomplete, unavailable, or insufficiently diverse for a target family, we generate or curate additional targets ourselves.
SN-VC-50M denotes the released 50-million-example subset containing these generated and curated targets, while the rest of SN-VC can be reproduced from the released source lists, prompt templates, conversion rules, and examples.
For datasets that overlap with our evaluation benchmarks, we preserve the official benchmark splits and exclude the corresponding evaluation images and annotations from training.

\begin{figure}[t]
    \centering
    \begin{subfigure}[t]{0.59\textwidth}
        \centering
        \includegraphics[width=\linewidth,height=0.31\textheight,keepaspectratio]{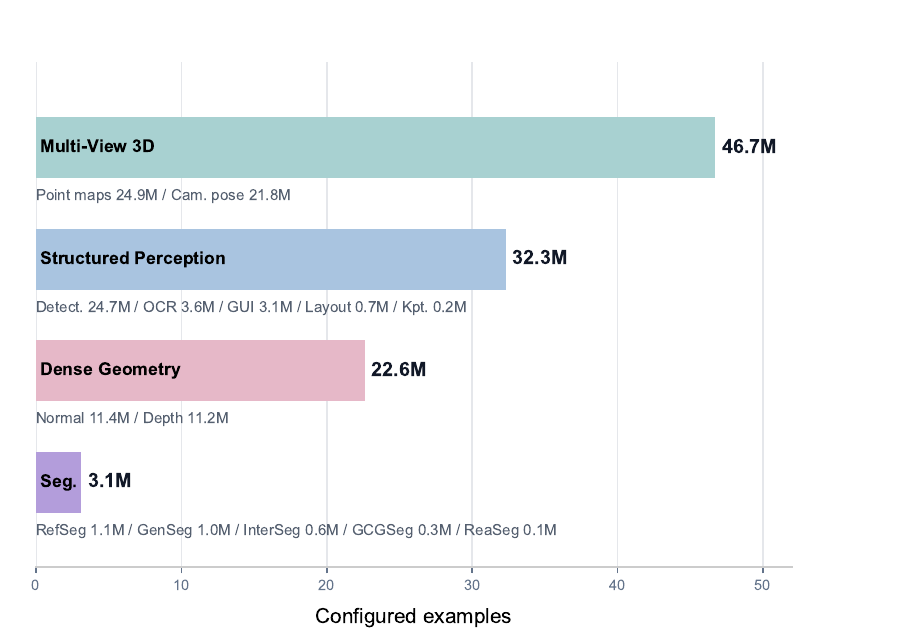}
        \caption{SN-VC source composition.}
        \label{fig:snvc-source-composition}
    \end{subfigure}%
    \begin{subfigure}[t]{0.4\textwidth}
        \centering
        \includegraphics[width=\linewidth,height=0.31\textheight,keepaspectratio]{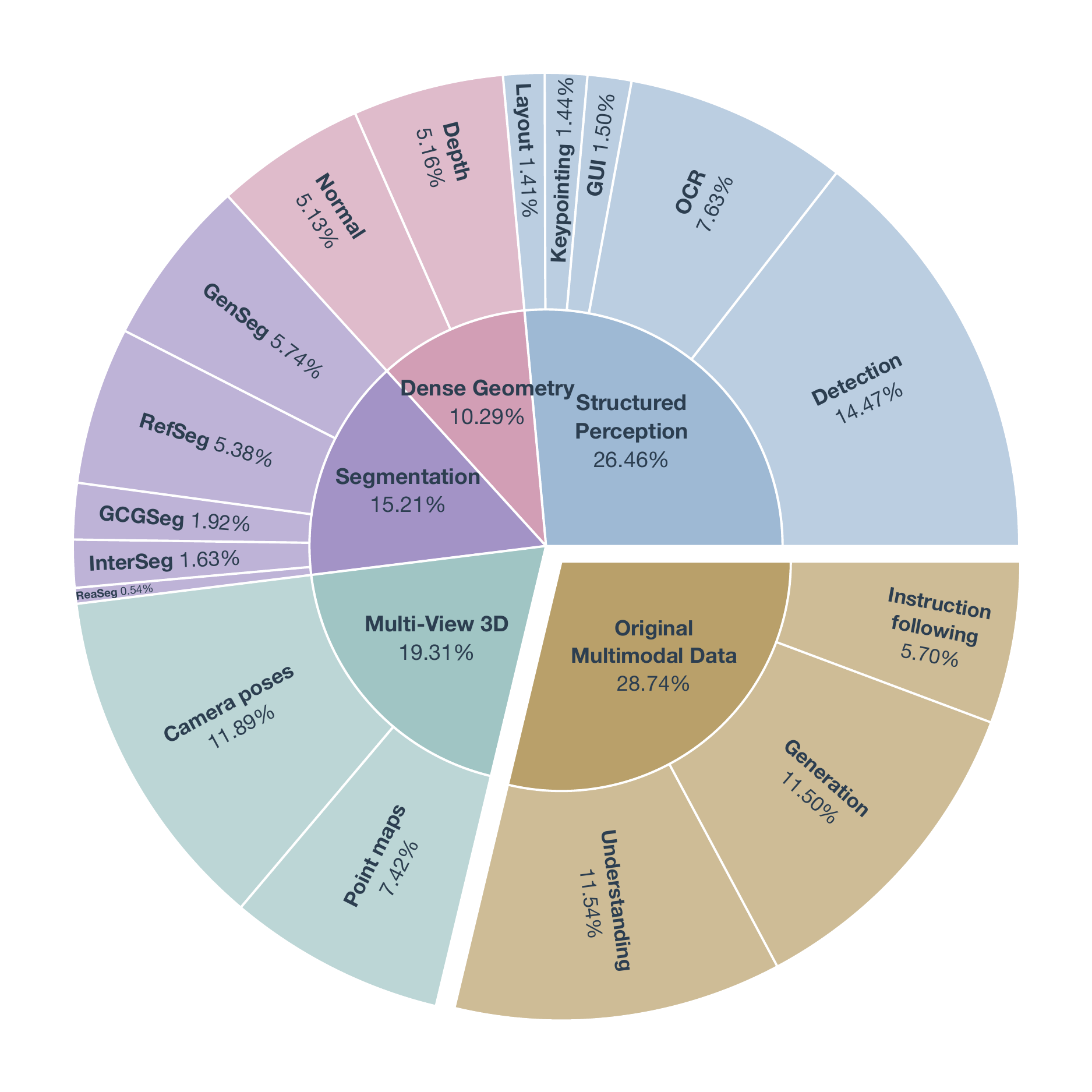}
        \caption{Training mixture composition.}
        \label{fig:training-mixture-composition}
    \end{subfigure}
    \caption{
        SN-VC source composition and training mixture.
        Left: configured source examples by converted family and subtype; right: realized sample-type proportions during training.
    }
    \label{fig:snvc-and-training-composition}
\end{figure}

\textbf{Corpus organization.}
We organize SN-VC into four source families according to the converted supervision they provide.
Fig.~\ref{fig:snvc-source-composition} summarizes the configured counts and subtypes of source examples.
Structured visual understanding sources cover detection-centered annotations together with GUI, OCR, layout, and keypoint supervision.
Dense geometric prediction sources provide depth and surface-normal supervision.
Segmentation sources cover mask-centric tasks, including referring, generic, interactive, grounded conversation generation (GCG), and reasoning segmentation.
Multi-view visual geometry sources contribute reconstruction targets and camera-pose annotations.
Appendix~\ref{app:data_protocol_construction} lists all datasets and construction details for each source family.
When the same source image appears in multiple tasks, we keep the corresponding converted examples as independent samples, since each task is defined by its own instruction and target response.

\textbf{Instruction-response conversion.}
Each source annotation is converted by a task template into an instruction-response training sample.
The visual input is chosen according to the task context: a single image for standard image tasks, an image with auxiliary visual prompts for interactive tasks, or an ordered image set for multi-view tasks.
The instruction states the task goal and expected output convention, and we instantiate multiple instruction variants for each task to improve prompt robustness.
The target response is produced deterministically from the source annotation: text-oriented tasks use normalized schemas, image-oriented tasks render masks or geometric maps, and mixed tasks place text and image components in a fixed order.
Representative prompt-response examples are shown in Fig.~\ref{fig:data-examples}.

\textbf{SN-VC-50M target curation.}
Some public sources contain incomplete supervision or annotations that cannot be directly used as decodable multimodal targets, so we additionally generate or curate targets for these cases.
For structured visual understanding, we draw on the data-construction pipeline of Rex-Omni~\cite{rexomni} to construct part of the detection and OCR data.
For dense geometric prediction, we use MoGe-2~\cite{wang2025moge2} to densify incomplete supervision and expand data diversity by generating additional depth and normal targets, followed by validity and scene-content filtering.
For segmentation, we curate mixed text-image targets such as grounded conversation generation segmentation, where region descriptions and color legends must be aligned with mask images.
For multi-view visual geometry, we complete sparse depth with LingBot-Depth~\cite{lingbotdepth} and filter examples with invalid depth, missing camera information, or inconsistent view metadata.
These generated and curated examples form SN-VC-50M, while the remaining SN-VC examples can be reconstructed from public datasets using the released source lists, templates, and conversion scripts.
Appendix Table~\ref{tab:open-source-datasets} summarizes the released SN-VC-50M task families and frame counts.

\section{Training}
\label{sec:training}

We train {SenseNova-Vision} by adapting Bagel-7B-MoT~\cite{bagel} to the unified vision-task corpus constructed in Sec.~\ref{sec:dataset_construction}.
Instead of training a multimodal model from scratch, we aim to endow an existing UMM with benchmark-compatible computer vision abilities while mitigating the degradation of its open-ended capabilities, including image understanding, instruction following, and image generation.
This section describes the mixed-task fine-tuning strategy, the high-resolution and multi-view training setup, and the training hyperparameters.

\textbf{Mixed-task fine-tuning.}
We perform supervised fine-tuning from the Bagel checkpoint on the SenseNova-Vision Corpus together with general-purpose multimodal data spanning visual question answering (VQA), text-to-image, and image-to-image tasks.
In a unified understanding and generation framework, this data mixture allows our model to learn representations of benchmark-readable outputs for structured visual understanding, dense geometric prediction, segmentation, and multi-view visual geometry, while mitigating degradation of the broad capabilities of the base UMM.

We adopt a joint sampling strategy that draws mini-batches from a weighted mixture of converted computer vision samples and auxiliary multimodal samples; the realized training mixture is shown in Fig.~\ref{fig:training-mixture-composition}.
Each mini-batch may contain samples drawn from multiple task categories and therefore produces a mixture of text and visual supervision targets.
During training, text and visual targets are interleaved in the same optimization process despite their different objectives.
Text-form outputs, including detection boxes, OCR strings, keypoints, and camera parameters, are tokenized and optimized with the standard cross-entropy (CE) loss under the next-token-prediction paradigm.
Visual outputs, including masks, depth maps, normal maps, and point maps, are encoded into a VAE latent space and optimized with the rectified-flow training objective inherited from Bagel.
In this way, heterogeneous computer vision targets are learned through the native text and image decoders of the base model, without introducing task-specific prediction heads.

The mixed-task joint training strategy enables the model to learn shared parameters across interleaved task formats and output modalities, potentially improving its generalization to zero-shot task variants.
Sec.~\ref{sec:text-prompted-interseg} gives further analysis.

\textbf{High-resolution and multi-view training.}
Compared with the VAE pathway, SigLIP2~\cite{siglip2} provides stronger semantic conditioning over input images, which benefits language-guided and region-level perception.
For image-input tasks requiring fine spatial conditioning, especially segmentation, we keep the SigLIP2 input resolution up to 980 pixels for both understanding and image-conditioned generation.
This preserves high-resolution conditioning whenever generation depends on an input image, whereas Bagel uses a lower SigLIP2 conditioning limit on the generation side.

For multi-view visual geometry, each data sample corresponds to a multi-view scene.
Due to memory constraints, each training sample is formed by randomly selecting at most 10 views from the corresponding scene.
For point map reconstruction, the selected views are aligned to the first view, center-normalized, and invalid pixels are mapped to a distant sky box.
For camera pose estimation, we reserve the final 2,009 vocabulary entries of the base model and repurpose them as a dedicated set of special tokens.
Among these tokens, 2,001 encode quantized pose parameters, including quaternion-based rotations, unit translation vectors, and scales, while the remaining 8 act as structural placeholders, as detailed in Appendix Table~\ref{tab:pose_token_markers}.
These special tokens are used exclusively for camera pose estimation, while other structured outputs remain serialized using ordinary text tokens.

\textbf{Training setup.}
We perform SFT with the VAE visual encoder frozen, while allowing all other modules and connectors to be updated.
We use the AdamW optimizer with a learning rate of $2.5\times10^{-5}$ and no weight decay.
We follow the method used in Bagel to pack training mini-batches, with 32K--36K tokens per rank and a maximum context window of 32K per sample.
The dropout rates for text, ViT and VAE input tokens are set to 0.05, 0.1 and 0.1, respectively.
The model is trained for 50K steps including 500 warm-up steps, and an EMA ratio of 0.995; the EMA checkpoint is used for evaluation.
All other configurations remain the same as in the SFT stage of Bagel.

\section{Experiments}
\label{sec:experiments}

We evaluate whether SenseNova-Vision can cover a broad set of computer vision tasks through unified multimodal generation.
The evaluation is organized into four task families: structured visual understanding, dense geometric prediction, segmentation, and multi-view visual geometry.
We further compare SenseNova-Vision with recent generalist visual models and evaluate multimodal understanding and image generation to assess whether training on computer vision tasks preserves the base model's general capabilities.
Finally, we provide additional analyses of convergence behavior, qualitative results, and language-defined task variants beyond standard benchmark settings.

All tasks are formulated with natural-language instructions.
Textual outputs are parsed into benchmark-specific structured formats, such as boxes, points, recognized text, keypoints, and camera parameters.
Image outputs are decoded into masks, depth maps, normal maps, or 3D point maps using deterministic conversion rules.
This protocol allows heterogeneous computer vision tasks to be evaluated after generation through task-specific benchmark metrics.

\subsection{Structured Visual Understanding}

Structured visual understanding evaluates tasks whose outputs can be represented as structured textual predictions, such as bounding boxes, points, recognized text, and keypoint coordinates.
As shown in Table~\ref{tab:structured-results}, the evaluation covers box- and point-based localization and grounding on COCO-Common~\cite{lin2014microsoft}, HumanRef~\cite{zhang2024humanref}, RefCOCOg val/test~\cite{yu2016modeling,kazemzadeh-etal-2014-referitgame}, LVIS~\cite{gupta2019lvis}, Dense200~\cite{rexomni}, and VisDrone~\cite{du2019visdrone}, OCR localization on HierText~\cite{long2022towards,long2023icdar} and ICDAR15~\cite{ICDAR2015}, GUI grounding on ScreenSpot-V2~\cite{wu2024atlas}, and keypoint detection on COCO-Kpt.~\cite{lin2014microsoft}.

These tasks test whether the model can produce benchmark-compatible coordinate-level predictions through a unified text generation interface.
They are challenging for serialized generation because dense scenes require long object lists, stable ordering, and precise coordinates.
SenseNova-Vision achieves strong overall performance, especially on dense, long-tailed, small-object, referring, and OCR localization benchmarks.

\begin{table}[t]
    \centering
    \small
    \setlength{\tabcolsep}{2pt}
    \resizebox{\textwidth}{!}{
        \begin{tabular}{l|cccccccccc}
            \toprule
            \multirow{3}{*}{Method}
            & \multicolumn{6}{c}{Object Detection}
            & \multicolumn{2}{c}{OCR}
            & \multicolumn{1}{c}{GUI}
            & \multicolumn{1}{c}{Keypoint} \\
            & \multicolumn{1}{c}{COCO-Com.}
            & \multicolumn{1}{c}{HR/RefCOCOg V/T}
            & \multicolumn{1}{c}{LVIS}
            & \multicolumn{1}{c}{Dense200}
            & \multicolumn{2}{c}{VisDrone}
            & HierText
            & ICDAR15
            & ScreenSpot-V2
            & COCO-Kpt. \\
            \cmidrule(lr){2-2} \cmidrule(lr){3-3} \cmidrule(lr){4-4} \cmidrule(lr){5-5} \cmidrule(lr){6-7} \cmidrule(lr){8-9} \cmidrule(lr){10-10} \cmidrule(lr){11-11}
            & bbox
            & bbox
            & bbox
            & bbox
            & bbox & point
            & bbox & bbox
            & bbox
            & point \\
            \midrule
            Grounding DINO-Swin-T~\cite{liu2024grounding} & \textbf{56.6} & 25.2 / 45.9 / 46.8                   & 38.8              & 33.1             & 38.5              & --                & --               & --               & --    & --               \\
            Bagel~\cite{bagel}                            & 50.2             & 74.6 / 76.4 / \underline{77.8}       & 46.8              & 42.4             & 23.0              & 36.9              &   7.1             &    15.8      & 81.1 & --               \\
            Qwen3-VL-8B-Instruct~\cite{qwen2025qwen3vl}   & 46.6             & 70.4 / 72.3 / 72.6                   & 43.2              & 13.5             & 28.7              & 35.7              & 22.4            & 25.4            & \underline{90.5} & --               \\
            Qwen3.5-9B~\cite{qwen2026qwen35}              & 49.3             & 71.7 / 72.1 / 72.6                   & 43.2              & 27.5             & 26.8              & 41.7              & 19.6            & 11.4            & \textbf{92.2} & --               \\
            LocateAnything~\cite{locateanything}          & \underline{54.7}             & 78.7 / \underline{76.7} / 77.6       & \underline{50.7}  & \underline{58.7} & \underline{39.9}  & \underline{60.4}  & \underline{29.1} & 26.4            & 85.5 & --               \\
            Rex-Omni~\cite{rexomni}                   & 52.9             & \underline{79.9} / 73.6 / 74.3       & 46.9              & 58.3             & 35.8              & 58.9              & 28.0            & \underline{28.1} & 88.4 & \underline{32.6} \\
            \textbf{SenseNova-Vision}                             & \textbf{56.6}             & \textbf{80.2} / \textbf{79.6} / \textbf{80.5} & \textbf{54.8}     & \textbf{66.8}    & \textbf{43.3}  & \textbf{62.9}       & \textbf{31.2}   & \textbf{49.5}   & 85.9 & \textbf{34.6}    \\
            \bottomrule
        \end{tabular}
    }
    \caption{
        Quantitative comparison of structured visual understanding.
        Performance is assessed using F1@mIoU for box-based detection, referring, and OCR localization tasks, F1@Point for VisDrone point localization, click accuracy for GUI grounding, and F1@mOKS for keypoint localization.
        Higher values indicate better performance for all metrics.
    }
    \label{tab:structured-results}
\end{table}

\subsection{Dense Geometric Prediction}

Dense geometric prediction evaluates pixel-aligned geometric outputs, including monocular depth estimation and surface normal estimation.
Depth maps are decoded from generated depth images and evaluated using affine-invariant depth metrics, while surface normal maps are recovered from color-coded images and evaluated by angular error.
As shown in Table~\ref{tab:dense-geometry-results}, we report depth results on NYUv2~\cite{silberman2012indoor}, KITTI~\cite{geiger2013vision}, ETH3D~\cite{schoeps2017multi}, ScanNet~\cite{dai2017scannetrichlyannotated3dreconstructions}, and DIODE~\cite{vasiljevic2019diode}, together with normal estimation results on ScanNet, iBims-1~\cite{koch2018evaluation}, and NYUv2.

These tasks test whether dense geometric maps can be produced as image outputs without task-specific depth or normal prediction heads.
SenseNova-Vision achieves strong performance across both depth and normal estimation, outperforming recent generation-based baselines on several benchmarks and remaining competitive with geometry-specialized models.

\begin{table}[t]
    \centering
    \small
    \setlength{\tabcolsep}{3pt}
    \begin{tabular}{l|ccccc|ccc}
        \toprule
        \multirow{3}{*}{Method} & \multicolumn{5}{c|}{Depth} & \multicolumn{3}{c}{Normal} \\
        & \multicolumn{5}{c|}{abs rel. $\downarrow$ / $\delta_1 \uparrow$} & \multicolumn{3}{c}{mean err. $\downarrow$ / $\delta_{11.25} \uparrow$} \\
        \cmidrule(lr){2-6} \cmidrule(lr){7-9}
        & NYUv2 & KITTI & ETH3D & ScanNet & DIODE
        & ScanNet & iBims-1 & NYUv2 \\
        \midrule
        DSINE~\cite{bae2024dsine}                   & --                              & --                               & --                             & --                               & --                    & 16.2 / 61.0                         & 17.1 / 67.4                      & 16.4 / 59.6                        \\
        DepthAnything~\cite{depthanything}          & 4.3 / \textbf{98.1}              & 7.6 / 94.7                          & 12.7 / 88.2                        & 4.3 / 98.1                         & 26.0 / 75.9             & --                              & --                           & --                             \\
        DepthAnything V2~\cite{depthanythingv2}     & 4.5 / 97.9                        & 7.4 / 94.6                          & 13.1 / 86.5                        & 4.2 / 97.8                         & 26.5 / 73.4             & --                              & --                           & --                             \\
        $\ast$MoGe-2~\cite{wang2025moge2}          & \textbf{3.5} / 98.0& \textbf{5.5} / \textbf{97.7}& \textbf{3.4} / \textbf{98.8}& \textbf{3.4} / \textbf{98.3}& \textbf{23.0} / \textbf{82.3}& \textbf{12.8} / \textbf{68.4}& \textbf{14.7} / \textbf{70.4}& \textbf{14.7} / \textbf{62.3}\\
        \midrule
        Marigold~\cite{ke2024marigold}              & 5.5 / 96.4                        & 9.9 / 91.6                           & 6.5 / 95.9                         & 6.4 / 95.2                           & 30.8 / 77.3 & 21.3 / 45.6                         & 18.5 / 64.7                      & 20.9 / 50.5                        \\
        DICEPTION~\cite{diception}                  & 6.1 / 96.0                        & 6.9 / 94.9                           & 5.0 / 97.5                         & 7.2 / 94.4                           & 28.9 / 72.2             & 18.8 / 53.6                         & --                           & 18.3 / 52.9                        \\
        FE2E~\cite{fe2e2025editor}                  & 4.1 / 97.7& 6.6 / \textbf{96.0}& \textbf{3.8} / \textbf{98.7}& 4.4 / 97.5                           & 22.8 / \textbf{81.2}& 13.8 / 67.2& \textbf{15.1} / \textbf{70.6}& 16.2 / 59.6\\
        Lotus-2~\cite{lotus2}                       & 4.1 / 97.6& 6.7 / 94.5                           & 4.6 / 98.1& 4.2 / 97.6& 22.1 / 75.2 & 14.2 / 66.8                         & 15.4 / 70.4          & 16.9 / 59.0                        \\
        \textbf{SenseNova-Vision}       & \textbf{4.0} / \textbf{98.1}& \textbf{5.9} / 95.9& 4.3 / 97.4& \textbf{3.9} / \textbf{98.0}& \textbf{20.6} / 76.4& \textbf{12.8} / \textbf{68.9}     & 15.4 / 69.1        & \textbf{14.4} / \textbf{62.7}\\
        \bottomrule
    \end{tabular}
    \caption{
        Quantitative comparison of dense geometric prediction.
        The upper block reports geometry-specialized models, while the lower block compares generation-based methods.
        Methods denoted with an asterisk ($\ast$) have been re-evaluated to ensure a direct and consistent comparison with our method.
    }
    \label{tab:dense-geometry-results}
\end{table}

\subsection{Segmentation}

Segmentation evaluates mask prediction under semantic, referring, reasoning, grounded, and interactive guidance.
Generated segmentation images are decoded into benchmark masks using the color palettes, visual prompts, or target specifications defined by each instruction.
As shown in Table~\ref{tab:segmentation-results}, we report generic, referring, reasoning, Grounded Conversation Generation (GCG)~\cite{rasheed2024glamm}, and interactive segmentation results using the corresponding benchmark metrics.

These tasks test two abilities: selecting the intended language-conditioned target and producing benchmark-compatible masks.
SenseNova-Vision achieves competitive overall performance among unified segmentation and multimodal baselines, with strong results on reasoning and GCG segmentation.
Specialized segmentation models remain stronger on several generic and referring segmentation metrics, where segmentation-specific pretrained mask models such as SAM~\cite{sam} or Mask2Former~\cite{mask2former} can provide strong mask priors.

\begin{table}[t]
    \centering
    \small
    \setlength{\tabcolsep}{6pt}
    \begin{tabular}{l|ccccc}
        \toprule
        \multirow{2}{*}{Method} & Gen. Seg. & Ref. Seg. & Rea. Seg. & GCG Seg. & Inter. Seg. \\
        & Pan. / Sem. & RefCOCO / + / g & Val / Test & Val / Test & Point / Box \\
        \midrule
        LISA-7B~\cite{lisa}               & --                                  & 74.9 / 65.1 / 67.9                                  & 52.9 / 47.3             & 62.0 / 61.7                         & --                               \\
        PSALM~\cite{zhang2024psalm}       & \textbf{55.9} / \textbf{66.6}       & 83.6 / 72.9 / 73.8                                  & --                      & --                                  & \underline{64.3} / 67.3          \\
        Text4Seg~\cite{lan2025text4seg}   & --                                  & 79.2 / 72.8 / 74.0                                  & 59.1 / 57.1             & --                                  & --                               \\
        LENS~\cite{zhu2026lens}           & --                                  & \underline{84.2} / \textbf{79.4} / \underline{81.2} & \underline{62.1} / 57.2 & --                                  & --                               \\
        ConverSeg~\cite{conversational}   & --                                  & 79.4 / 74.3 / 74.9                                  & 61.9 / 57.0             & --                                  & --                               \\
        X-SAM~\cite{xsam}                 & \underline{54.7} / \underline{66.5} & \textbf{85.1} / \underline{78.0} / \textbf{83.8}    & 56.6 / \underline{57.8} & \textbf{69.4} / \textbf{69.0}       & \textbf{65.4} / \underline{70.0} \\
        \textbf{SenseNova-Vision}           & 48.8 / 64.0                         & 81.3 / 76.0 / 80.3                                  & \textbf{63.2} / \textbf{60.7}    & \underline{65.7} / \underline{66.2} & 60.9 / \textbf{73.9}             \\
        \bottomrule
    \end{tabular}
    \caption{
        Quantitative comparison of segmentation.
        For Gen. Seg., we report PQ for panoptic segmentation (Pan.) and mIoU for semantic segmentation (Sem.).
        For Ref. Seg., we report cIoU, defined as the ratio of total true positives to total union.
        For Rea. Seg., we report gIoU; for GCG Seg. and Inter. Seg., we report mIoU.
        Higher values indicate better performance for all metrics.
    }
    \label{tab:segmentation-results}
\end{table}

\subsection{Multi-View Visual Geometry}

Multi-view visual geometry evaluates geometric prediction from multiple input images.
We focus on multi-view point map reconstruction and camera pose estimation, with results for both tasks reported in Table~\ref{tab:multiview-results}.
For reconstruction on 7Scenes~\cite{shotton2013scene} and ETH3D~\cite{schoeps2017multi}, we report accuracy and completeness following VGGT~\cite{vggt}, together with F1-score using the thresholds from Depth Anything 3~\cite{depthanything3}; for camera pose estimation on RealEstate10K (Re10K)~\cite{zhou2018stereo} and CO3Dv2~\cite{reizenstein2021commonobjects3dlargescale}, we report relative rotation accuracy (RRA), relative translation accuracy (RTA), and AUC under the 30-degree threshold.
All methods are re-evaluated to ensure a fair and direct comparison with SenseNova-Vision.

These tasks test whether the model can align information across multiple views and produce view-specific geometric outputs for reconstruction and camera pose estimation.
SenseNova-Vision achieves strong results among generalist geometric approaches, especially on ETH3D reconstruction and camera pose estimation.
Compared with feed-forward geometric models such as VGGT and Depth Anything 3, a performance gap remains on several metrics, highlighting the continued benefit of geometry-focused training and geometric inductive biases.

\begin{table}[t]
    \centering
    \small
    \setlength{\tabcolsep}{6pt}
    \begin{tabular}{l|cc|cc}
        \toprule
        \multirow{3}{*}{Method} & \multicolumn{2}{c|}{Multi-View Reconstruction} & \multicolumn{2}{c}{Camera Pose} \\
        & \multicolumn{2}{c|}{Acc.$\downarrow$ / Comp.$\downarrow$ / F1$\uparrow$} & \multicolumn{2}{c}{RRA@30$\uparrow$ / RTA@30$\uparrow$ / AUC@30$\uparrow$} \\
        \cmidrule(lr){2-3} \cmidrule(lr){4-5}
        & \multicolumn{1}{c}{7Scenes} & \multicolumn{1}{c|}{ETH3D} & \multicolumn{1}{c}{Re10K} & \multicolumn{1}{c}{CO3Dv2} \\
        \midrule
        DUSt3R~\cite{dust3r}                     & 0.026 / 0.034 / 87.1                            & 0.359 / 0.531 / 66.6                            & 99.8 / 84.9 / 67.6                             & 97.7 / 93.4 / 78.3                            \\
        DepthAnything3~\cite{depthanything3}     & \textbf{0.020} / \textbf{0.026} / \textbf{90.5} & 0.228 / 0.212 / 76.6                            & \textbf{100.0} / \textbf{96.4} / \textbf{89.6} & \textbf{99.3} / \textbf{98.0} / \textbf{91.8} \\
        VGGT~\cite{vggt}                         & 0.023 / 0.032 / 88.4                            & \textbf{0.177} / \textbf{0.155} / \textbf{80.9} & \textbf{100.0} / 93.5 / 79.3                   & 98.3 / 96.6 / 89.2                            \\
        MoRe~\cite{MoRe4d2026}                   & 0.038 / 0.039 / 77.1                            & 0.348 / 0.318 / 62.7                            & \textbf{100.0} / 94.0 / 79.1                   & 98.4 / 96.3 / 83.0                            \\
        \midrule
        MapAnything~\cite{keetha2026mapanything} & \textbf{0.027} / 0.029 / 87.8                   & 0.400 / 0.524 / 67.0                            & \textbf{100.0} / 93.5 / \textbf{80.7}          & 95.5 / 91.6 / 70.9                            \\
        G2VLM~\cite{g2vlm}                       & 0.084 / 0.056 / 59.2                            & 0.784 / 0.553 / 36.7                            & 99.8 / 77.5 / 51.8                             & 96.3 / 92.0 / 55.2                            \\
        \textbf{SenseNova-Vision}                & 0.028 / \textbf{0.026} / \textbf{87.9}          & \textbf{0.301} / \textbf{0.175} / \textbf{72.2} & 99.8 / \textbf{94.2} / 77.3                    & \textbf{97.4} / \textbf{95.4} / \textbf{80.1} \\
        \bottomrule
    \end{tabular}
    \caption{Quantitative comparison of multi-view point map reconstruction and camera pose estimation.
        We evaluate feed-forward geometric models (top) alongside generalist geometric approaches (bottom).
        MapAnything is classified within the latter category, as it accepts images with optional geometric inputs and fuses the encoded features with a multi-view transformer.
    }
    \label{tab:multiview-results}
\end{table}

\subsection{Comparison with Generalist Vision Models}

The previous sections mainly compare SenseNova-Vision with strong task-specialized systems within each vision task family.
We further compare with recent generalist visual models that span multiple visual capabilities, as shown in Table~\ref{tab:generalist-visual-models}.
This comparison evaluates how broadly a single model can cover heterogeneous visual tasks under unified multimodal generation, beyond performing well on individual benchmarks.

We select Youtu-VL~\cite{youtuvl} and Vision Banana~\cite{visionbanana} as representative recent generalist visual models, since they extend vision task coverage from complementary directions.
Youtu-VL represents a vision-language understanding route, so we compare on benchmarks aligned with structured and semantic perception, including detection, referring segmentation, semantic segmentation, and depth.
Vision Banana represents an image-generation-centered route, so we compare on benchmarks aligned with image-space prediction, including segmentation, depth, and surface normals.

\begin{table}[t]
    \centering
    \small
    \setlength{\tabcolsep}{4pt}

    \textbf{(a) Comparison with Youtu-VL}
    \vspace{3pt}

    \begin{tabular}{l|c|c|c|c}
        \toprule
        \multirow{3}{*}{Method} & Detection & Sem. Seg. & Ref. Seg. & Depth \\
        & mAP $\uparrow$ & mIoU $\uparrow$ & cIoU $\uparrow$ & $\delta_1 \uparrow$ \\
        \cmidrule(lr){2-2} \cmidrule(lr){3-3} \cmidrule(lr){4-4} \cmidrule(lr){5-5}
        & COCO & Cityscapes & RefCOCO / + / g & NYUv2 \\
        \midrule
        Youtu-VL                  & 47.1          & 70.4          & 80.7 / \textbf{76.2} / 76.5 & 90.4 \\
        \textbf{SenseNova-Vision} & \textbf{53.7} & \textbf{71.2} & \textbf{81.3} / 76.0 / \textbf{80.3} & \textbf{98.1} \\
        \bottomrule
    \end{tabular}

    \vspace{8pt}

    \textbf{(b) Comparison with Vision Banana}
    \vspace{3pt}

    \setlength{\tabcolsep}{3pt}
    \resizebox{\linewidth}{!}{%
        \begin{tabular}{l|c|c|c|cccc|ccc}
            \toprule
            \multirow{3}{*}{Method} & Sem. Seg. & Ref. Seg. & Rea. Seg. & \multicolumn{4}{c|}{Depth} & \multicolumn{3}{c}{Normal} \\
            & mIoU $\uparrow$ & cIoU $\uparrow$ & gIoU $\uparrow$ & \multicolumn{4}{c|}{$\delta_1 \uparrow$} & \multicolumn{3}{c}{mean err. $\downarrow$} \\
            \cmidrule(lr){2-2} \cmidrule(lr){3-3} \cmidrule(lr){4-4} \cmidrule(lr){5-8} \cmidrule(lr){9-11}
            & Cityscapes & RefCOCOg & ReasonSeg & KITTI & NYUv2 & DIODE & ETH3D & NYUv2 & ScanNet & DIODE \\
            \midrule
            Vision Banana                     & 69.9 & 73.8 & 79.3$^\dagger$ & 91.5$^\ddagger$ & 94.8$^\ddagger$ & 91.7$^\ddagger$ & 93.5$^\ddagger$ & 17.8 & 15.1 & 13.8 \\
            SenseNova-Vision                  & 71.2 & 80.3 & 63.2 & 95.9 & 98.1 & 76.4 & 97.4 & 14.4 & 12.8 & 15.3 \\
            \bottomrule
        \end{tabular}
    }

    \caption{
        Quantitative comparison with recent generalist visual models under the metrics reported in their original papers.
        For Vision Banana, $^\dagger$ denotes a Gemini-assisted ReasonSeg result,
        and $^\ddagger$ marks depth scores reported under an absolute-depth protocol,
        whereas SenseNova-Vision uses affine-invariant depth evaluation;
        these entries are included only as reference comparisons.
    }
    \label{tab:generalist-visual-models}
\end{table}

The comparison shows that recent generalist visual models already go beyond single-task specialization, but their coverage remains shaped by their native output modality.
SenseNova-Vision performs strongly against Youtu-VL on structured, semantic, and depth benchmarks, and remains competitive with Vision Banana on image-space segmentation and dense prediction benchmarks.
This broader coverage comes from unified multimodal generation over text, image, and mixed outputs, which better matches the heterogeneous output forms required by computer vision tasks.

Beyond comparisons with other generalist vision models, we further evaluate whether SenseNova-Vision retains the pretrained UMM's general multimodal abilities after mixed-task fine-tuning.
For multimodal understanding, SenseNova-Vision obtains 79.0 on MMVP~\cite{mmvp}, compared with 83.3 for Bagel~\cite{bagel}.
For text-to-image generation, SenseNova-Vision obtains 0.85 on GenEval~\cite{geneval}, compared with 0.82 for Bagel.
Overall, SenseNova-Vision preserves core multimodal abilities while expanding to a broad range of visual tasks.

\subsection{Qualitative Results and Additional Analysis}
\label{sec:qualitative-behavioral-analysis}

Beyond standard benchmark metrics, we further analyze training dynamics and qualitative behaviors of SenseNova-Vision, including convergence trends, a focused referring-style interactive segmentation variant, and broader free-form language-to-mask probes.

\subsubsection{Convergence Analysis}
\label{sec:convergence-analysis}

\begin{wrapfigure}[20]{r}{0.50\textwidth}
    \centering
    \includegraphics[width=\linewidth]{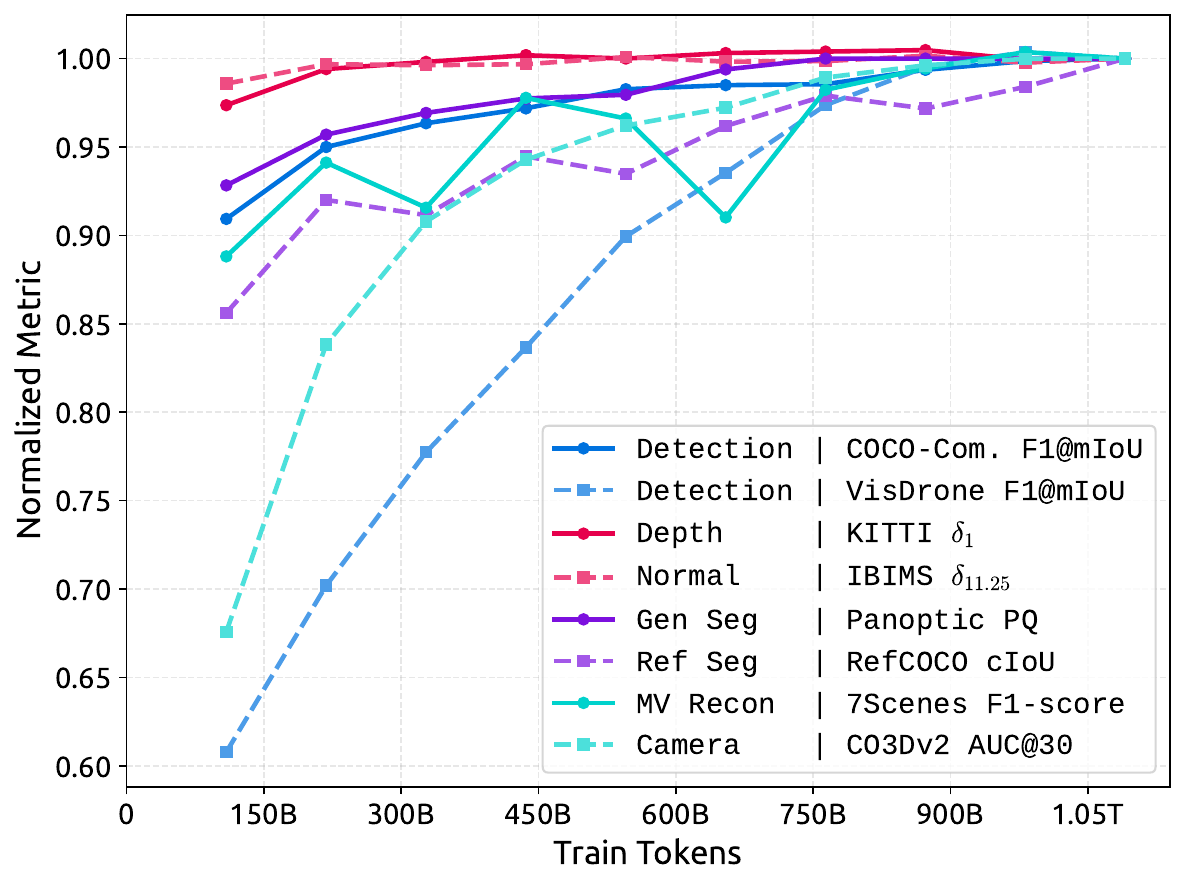}
    \caption{
        Normalized convergence curves across representative tasks.
        Each metric is normalized by its final-step value to compare relative convergence trends during training.
    }
    \label{fig:convergence-curve}
\end{wrapfigure}

Figure~\ref{fig:convergence-curve} visualizes normalized metric curves throughout training to compare learning progress and convergence speed across diverse computer vision tasks.
Depth and surface normal estimation converge fastest, likely because their targets are spatially aligned with the input image and may be close to image generation or editing patterns already seen during pretraining.
Multi-view reconstruction is spatially similar to depth prediction but requires alignment across multiple views, leading to a more moderate convergence speed.
Camera pose estimation converges more slowly because it requires cross-view alignment, newly introduced pose tokens, and deeper cross-modal understanding of geometric structure.
Common detection and generic segmentation show intermediate convergence, as both rely on semantic recognition and spatial alignment; detection progresses slightly faster, possibly because object localization is already partly covered by the pretrained model's visual-language experience.
Referring segmentation converges more slowly than generic segmentation because it places stronger demands on language-conditioned semantic grounding.
Dense detection is the slowest task, suggesting that crowded small-object localization requires precise cross-modal discrimination and detailed image understanding when many regions must be serialized.
Overall, these trends suggest that visual abilities do not converge uniformly but follow a staged learning pattern, with later convergence on tasks that require deeper alignment between fine-grained visual evidence, language intent, and spatial structure.

\subsubsection{Overall Qualitative Results}
\label{sec:overall-qualitative-analysis}

\begin{figure}[p]
    \centering
    \includegraphics[width=\textwidth,height=0.91\textheight,keepaspectratio]{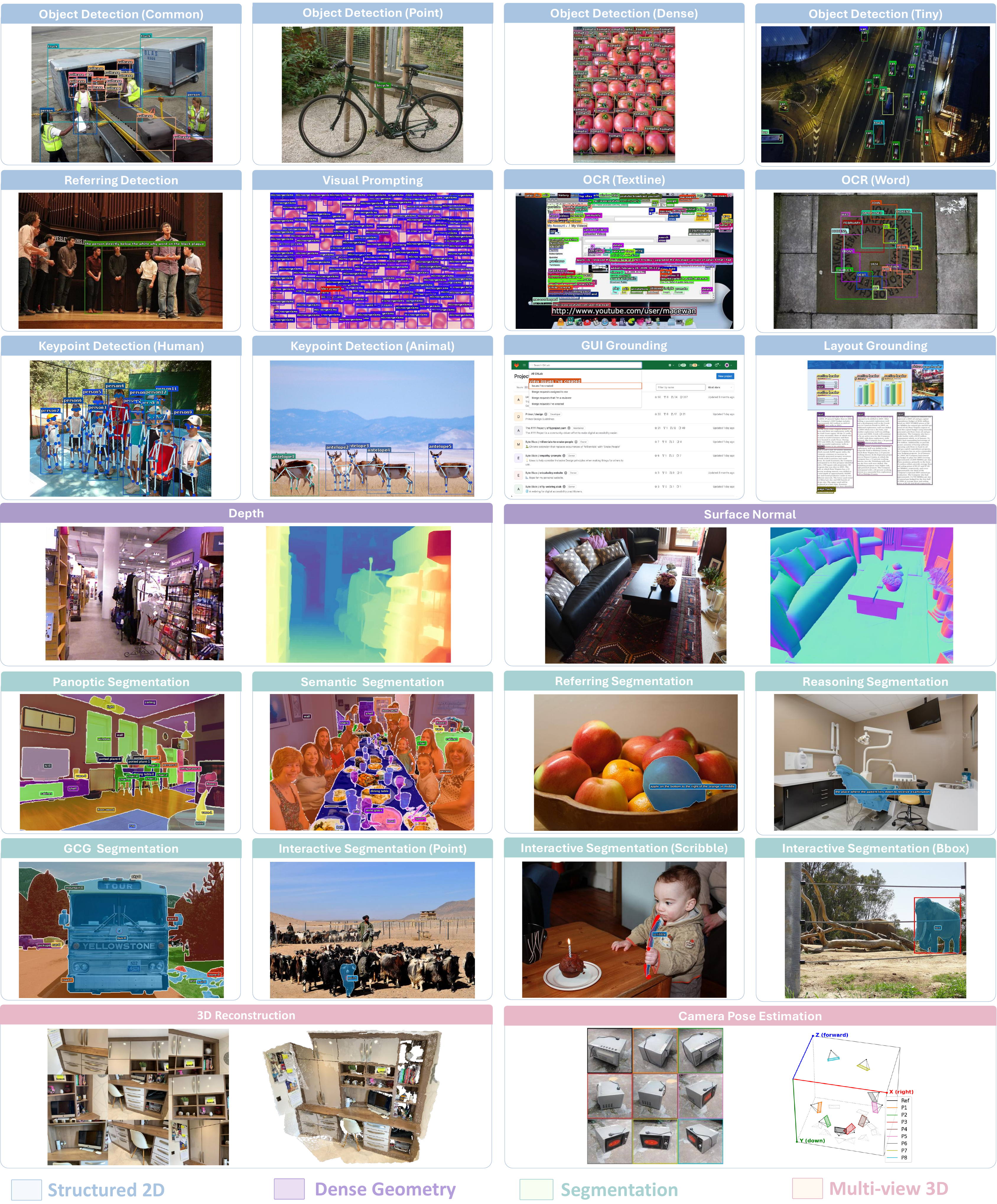}
    \caption{
        Qualitative results of SenseNova-Vision across representative computer vision tasks.
        All examples are generated through the same language-conditioned multimodal generation interface.
    }
    \label{fig:overall-qualitative-results}
\end{figure}

Figure~\ref{fig:overall-qualitative-results} presents qualitative examples across the visual task families supported by SenseNova-Vision.
Across structured visual understanding, dense geometric prediction, segmentation, and multi-view visual geometry, the model follows the requested task prompts and produces outputs that can be parsed or decoded into the corresponding task representations.

The examples show complete structured predictions in crowded, small-object, and document-like scenes.
For dense and spatial outputs, the generated depth and normal maps preserve major scene structures, the predicted masks are semantically aligned with the requested targets, and the multi-view outputs show reasonable cross-view geometric consistency.
These examples illustrate that SenseNova-Vision can preserve task-specific output structure across heterogeneous visual tasks within a unified generation framework.

\subsubsection{Referring-Style Interactive Segmentation}
\label{sec:text-prompted-interseg}

Interactive segmentation is usually conditioned on visual prompts such as rendered points, boxes, scribbles, or masks.
For a text-and-image generation interface, however, a sparse point cue can be expressed more compactly as text: a normalized coordinate specifies the location directly, whereas an image prompt encodes the same point through a full visual condition with substantial redundant information.
We therefore examine a referring-style interactive segmentation task in which the model receives a text-encoded point coordinate and generates the corresponding binary mask.
This task combines three training domains: image-conditioned interactive segmentation provides the mask-generation objective, referring segmentation provides the text-span interface for target specification, and structured visual understanding tasks such as detection, grounding, and pointing provide coordinate-level localization knowledge.

\begin{figure}[t]
    \centering
    \includegraphics[width=\linewidth]{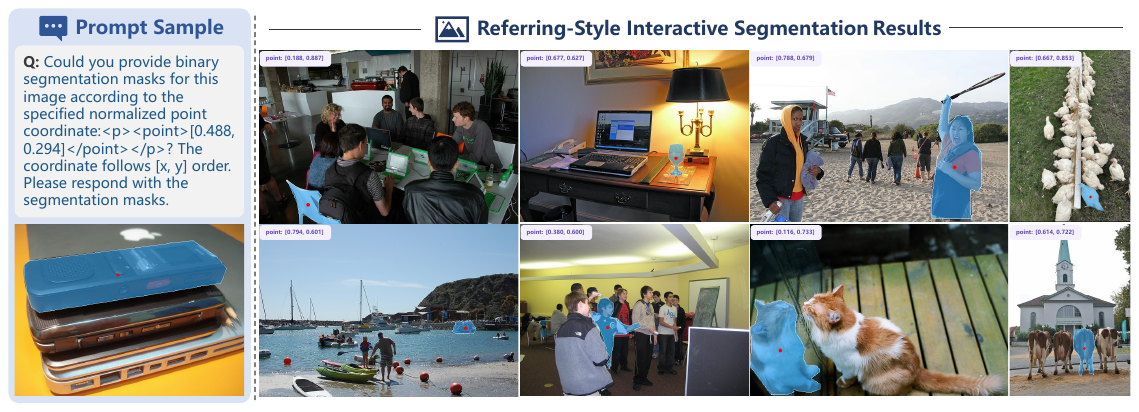}
    \vspace{-10pt}
    \caption{
        Qualitative examples of referring-style interactive segmentation with text-encoded point cues.
        SenseNova-Vision interprets normalized coordinates embedded in the referring span and generates binary masks for the indicated targets.
    }
    \label{fig:referring-style-interactive-segmentation}
\end{figure}

Concretely, standard referring segmentation uses \texttt{<p>referring expression</p>} to specify the target region through language.
We keep this referring-style span but replace the semantic expression with a normalized coordinate cue, written as \texttt{<p><point>[0.xxx, 0.xxx]</point></p>}.
This exact coordinate-based segmentation prompt is not included in the segmentation training protocol, but its components are familiar from different domains: the \texttt{<p>} span follows referring segmentation, and the normalized coordinate follows structured localization annotations.
Unlike an image prompt, which is already spatially aligned with the input image, a text-encoded point requires the model to convert numerical coordinates into image-space locations before generating the mask.
Figure~\ref{fig:referring-style-interactive-segmentation} shows qualitative examples of this composed prompt format.
Across the displayed cases, SenseNova-Vision follows the text-specified point accurately, selects the intended target among nearby or same-class objects, and generates binary masks even for small regions.

These results suggest that coordinate understanding learned from structured prediction can be interpolated into mask generation through the unified text-and-image interface.
The behavior is a concrete example of a language-defined task variant beyond the explicit training protocols, pointing toward the potential of unified multimodal models to recombine supervision from different computer vision domains.

\subsubsection{Free-Form Language-to-Mask Generation}
\label{sec:free_form_mask_generation}

Beyond the focused coordinate-to-mask variant above, we use segmentation as a testbed for broader free-form multimodal generation.
Dense geometric prediction tasks and multi-view visual geometry tasks are largely constrained by deterministic geometric targets, while structured visual understanding produces text outputs that may already benefit from the language flexibility of the pretrained UMM.
Segmentation lies between these cases: it requires dense spatial outputs, yet the target type, region grouping, instance organization, and mask representation protocol can all be flexibly specified through language.
During training, however, segmentation supervision still follows predefined task domains and fixed annotation protocols, as in conventional segmentation datasets.

This setting differs from conventional open-vocabulary segmentation, which mainly expands the category vocabulary of segmentation targets.
Here, the probes vary the data distribution, task definition, mask representation protocol, and target type.
We construct them by adapting task instructions from detection, grounding, segmentation, and OCR into non-canonical mask-generation prompts.
These probes test whether mask generation can move beyond fixed segmentation protocols and recombine capabilities learned from other task domains.

\begin{figure}[t]
    \centering
    \begin{subfigure}{\textwidth}
        \centering
        \includegraphics[width=\linewidth]{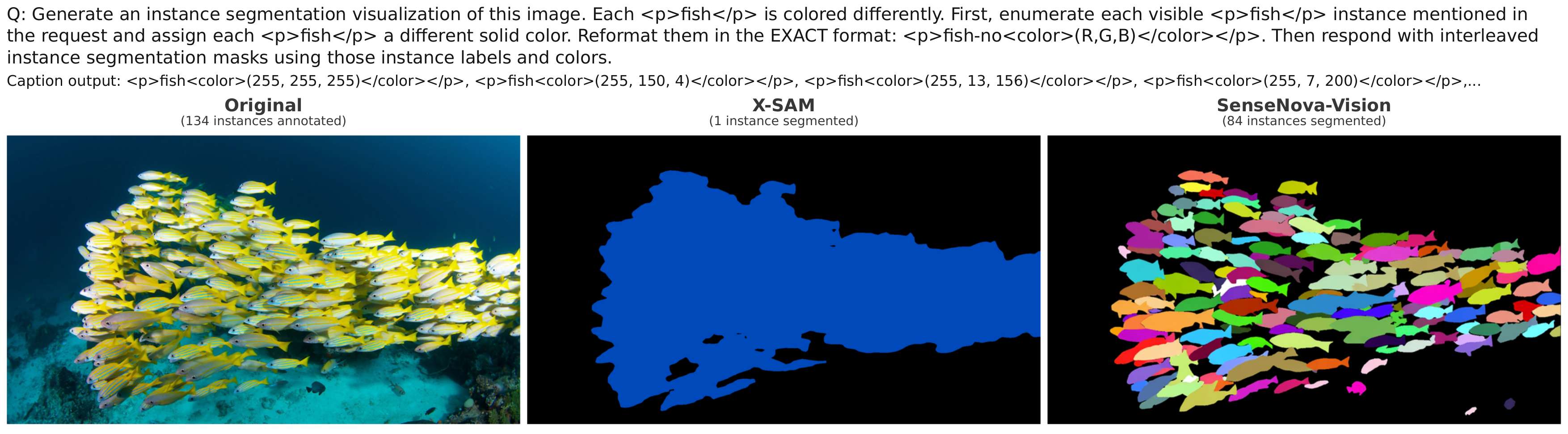}
        \caption{Crowded scenes with compact objects from Dense200.}
        \label{fig:dense-instance-segmentation-dense200}
    \end{subfigure}
    \vspace{-4pt}
    \begin{subfigure}{\textwidth}
        \centering
        \includegraphics[width=\linewidth]{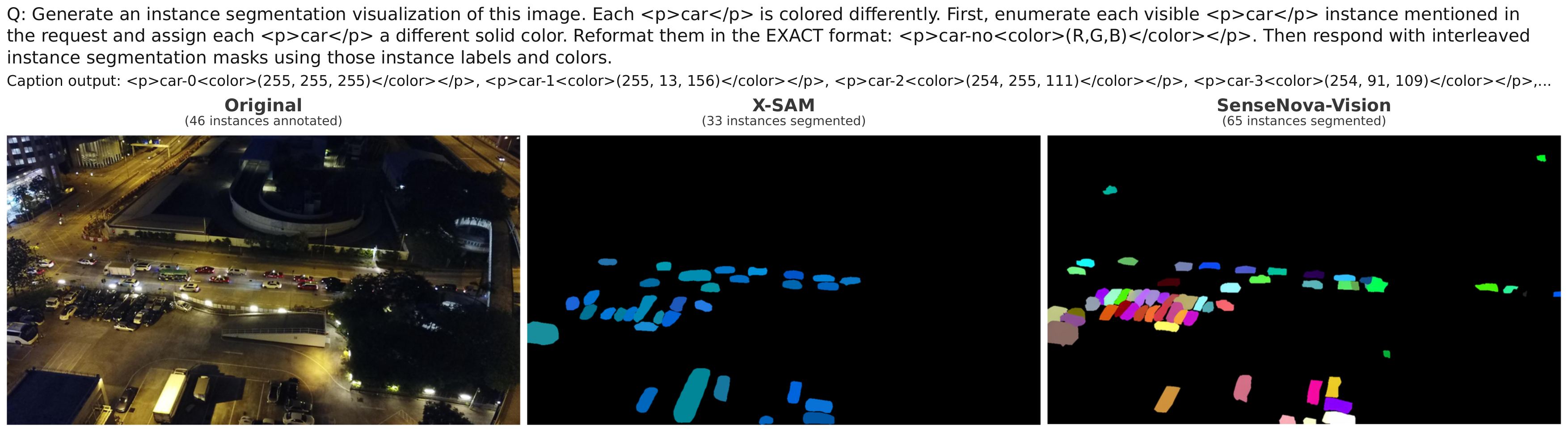}
        \caption{Aerial scenes with tiny objects from VisDrone.}
        \label{fig:dense-instance-segmentation-visdrone}
    \end{subfigure}
    \caption{
        \textbf{Dense instance segmentation with adapted detection-task instructions.}
        The prompts ask the model to enumerate visible instances of a target category and render different instances with distinct colors.
        (a) Despite heavy overlap and occlusion in a crowded scene, SenseNova-Vision separates a large fraction of individual instances and assigns distinct regions to adjacent same-category objects.
        (b) In the cluttered aerial night scene, the model recovers numerous tiny objects while distinguishing categories such as cars and trucks.
        Separate colors are assigned to individual targets, enabling instance-level decoding from the generated mask image.
        For comparison, X-SAM visualization is obtained by refilling the predicted masks for easier visual inspection.
    }
    \label{fig:dense-instance-segmentation-examples}
\end{figure}

The first probe focuses on dense instance segmentation using detection-task instructions adapted to request instance-level mask outputs.
Dense and small-object scenes are common in detection data but remain costly to annotate with instance-level masks.
On Dense200~\cite{rexomni} and VisDrone~\cite{du2019visdrone}, SenseNova-Vision extracts up to nearly one hundred compact or tiny objects in dense and aerial scenes, assigning each instance a distinct color that enables separate decoding of individual targets, as shown in Fig.~\ref{fig:dense-instance-segmentation-examples}.
These examples suggest that unified training can broaden the data distribution of mask generation to include dense and small-object cases that are better covered by detection data.

\begin{figure}[t]
    \centering
    \begin{subfigure}{0.49\textwidth}
        \centering
        \includegraphics[width=\linewidth,height=0.34\textheight,keepaspectratio]{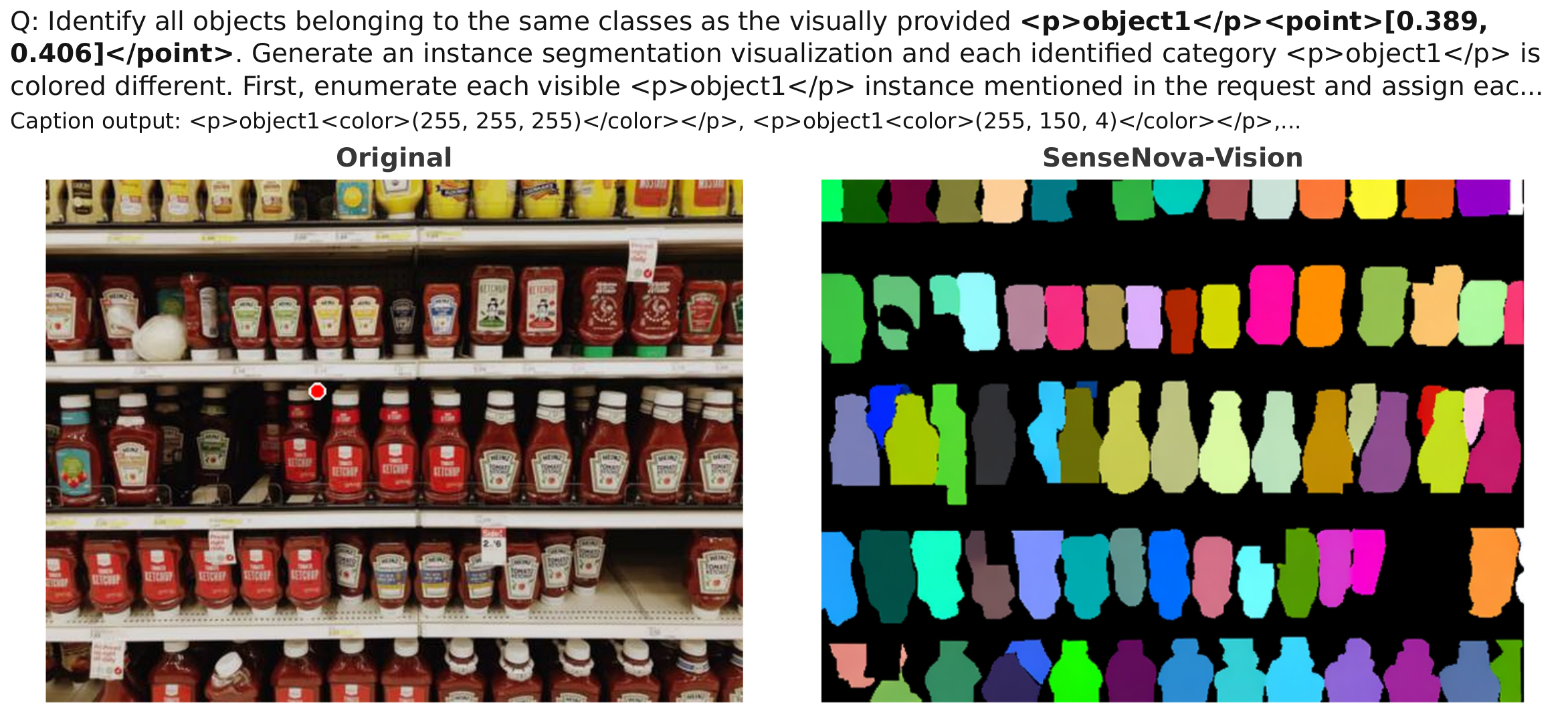}
        \caption{VGD segmentation with point reference cue.}
        \label{fig:vgd-segmentation-point-prompt}
    \end{subfigure}
    \hfill
    \begin{subfigure}{0.49\textwidth}
        \centering
        \includegraphics[width=\linewidth,height=0.34\textheight,keepaspectratio]{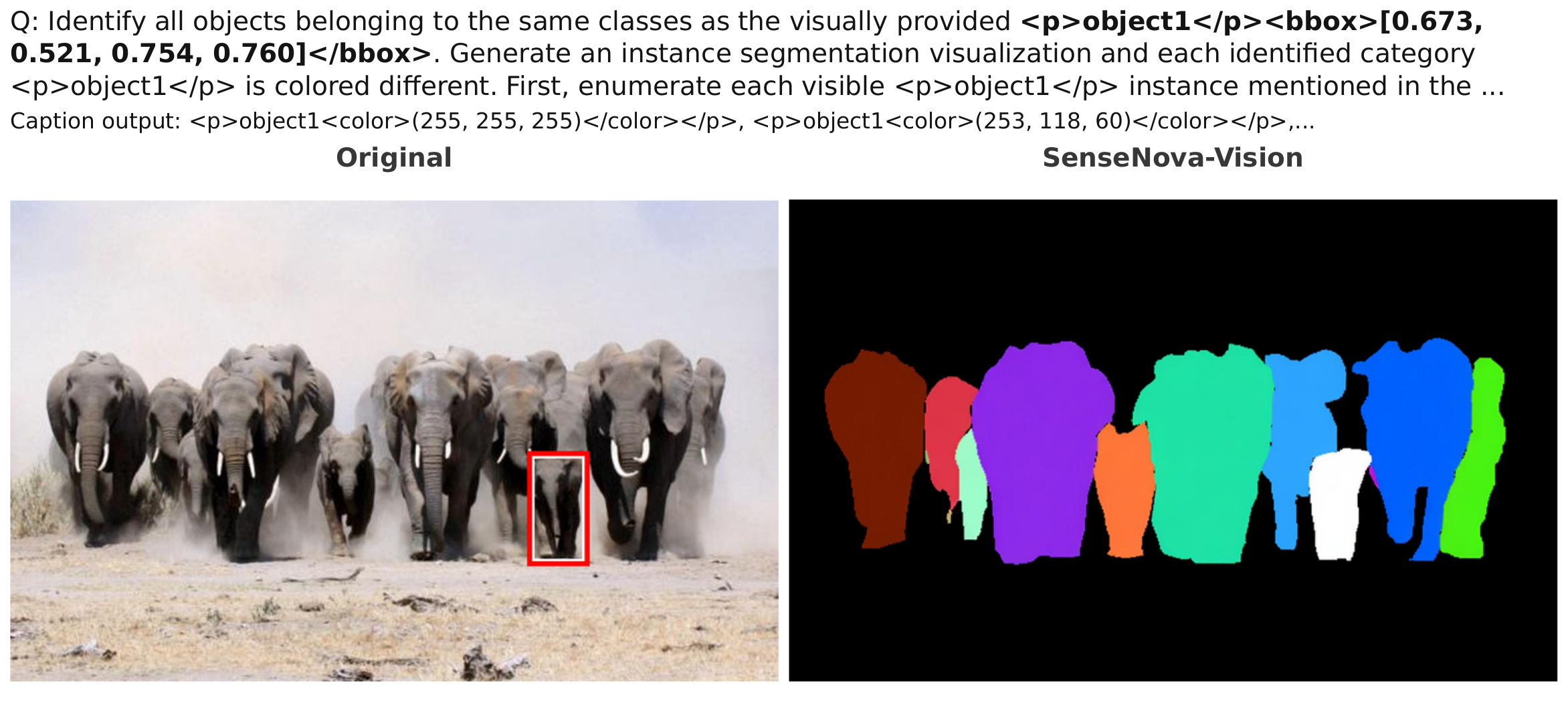}
        \caption{VGD segmentation with box reference cue.}
        \label{fig:vgd-segmentation-box-prompt}
    \end{subfigure}
    \vspace{-8pt}
    \caption{
        \textbf{Visual Grounded (VGD) segmentation with text-specified reference cues.}
        The prompt specifies a reference instance with a point or box written as textual coordinates and asks the model to segment other same-class instances.
        (a) The point cue selects one bottle, and SenseNova-Vision segments matching bottles across different shapes and sizes.
        (b) With a box cue on one elephant, the model separates nearby same-class instances despite overlap and occlusion.
    }
    \label{fig:vgd-segmentation-examples}
\end{figure}

Another probe follows the visual-prompt grounding setting, but uses text-specified reference cues for Visual Grounded (VGD) segmentation~\cite{xsam}.
A reference point or box written as textual coordinates indicates one instance, and the model must infer the corresponding visual match and segment other same-class targets in the image.
As shown in Fig.~\ref{fig:vgd-segmentation-examples}, SenseNova-Vision uses the text-specified point or box to identify the reference instance and generate masks for corresponding same-class targets.
This extends the task definition by adapting nearby grounding supervision into a new segmentation task that is not explicitly enumerated in the training data.

\begin{figure}[t]
    \centering
    \begin{subfigure}{0.49\textwidth}
        \centering
        \includegraphics[width=\linewidth]{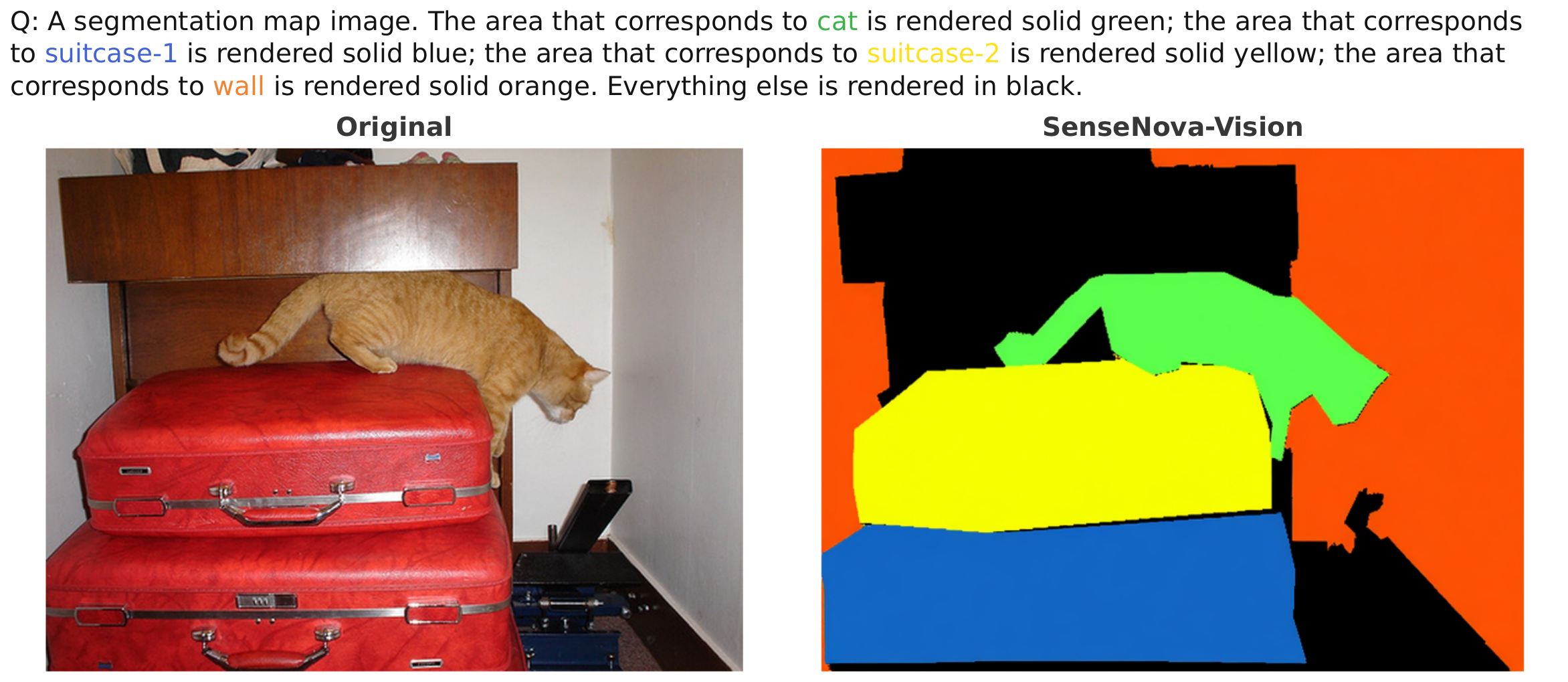}
        \label{fig:freeform-color-mask-case1}
    \end{subfigure}
    \hfill
    \begin{subfigure}{0.49\textwidth}
        \centering
        \includegraphics[width=\linewidth]{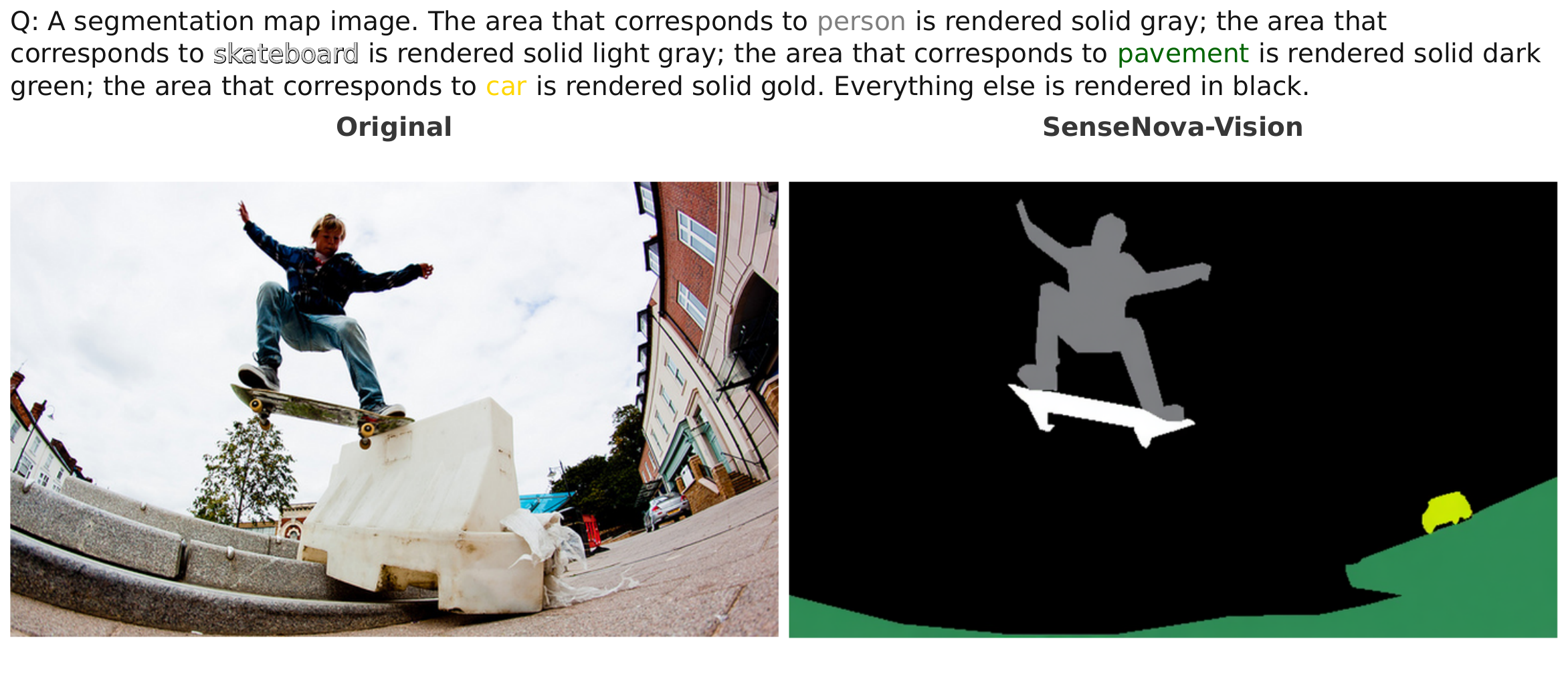}
        \label{fig:freeform-color-mask-case2}
    \end{subfigure}
    \vspace{-4pt}
    \caption{
        \textbf{Free-form color-coded mask generation.}
        The prompt describes region-color correspondences in natural language, without relying on the fixed class-color tags or exact RGB values used in training data.
        Both examples show mask-like outputs whose regions approximately follow the requested color assignments, illustrating language control over the mask representation protocol.
    }
    \label{fig:freeform-color-mask-generation}
\end{figure}

\begin{figure}[t]
    \centering
    \begin{subfigure}{0.49\textwidth}
        \centering
        \includegraphics[width=\linewidth]{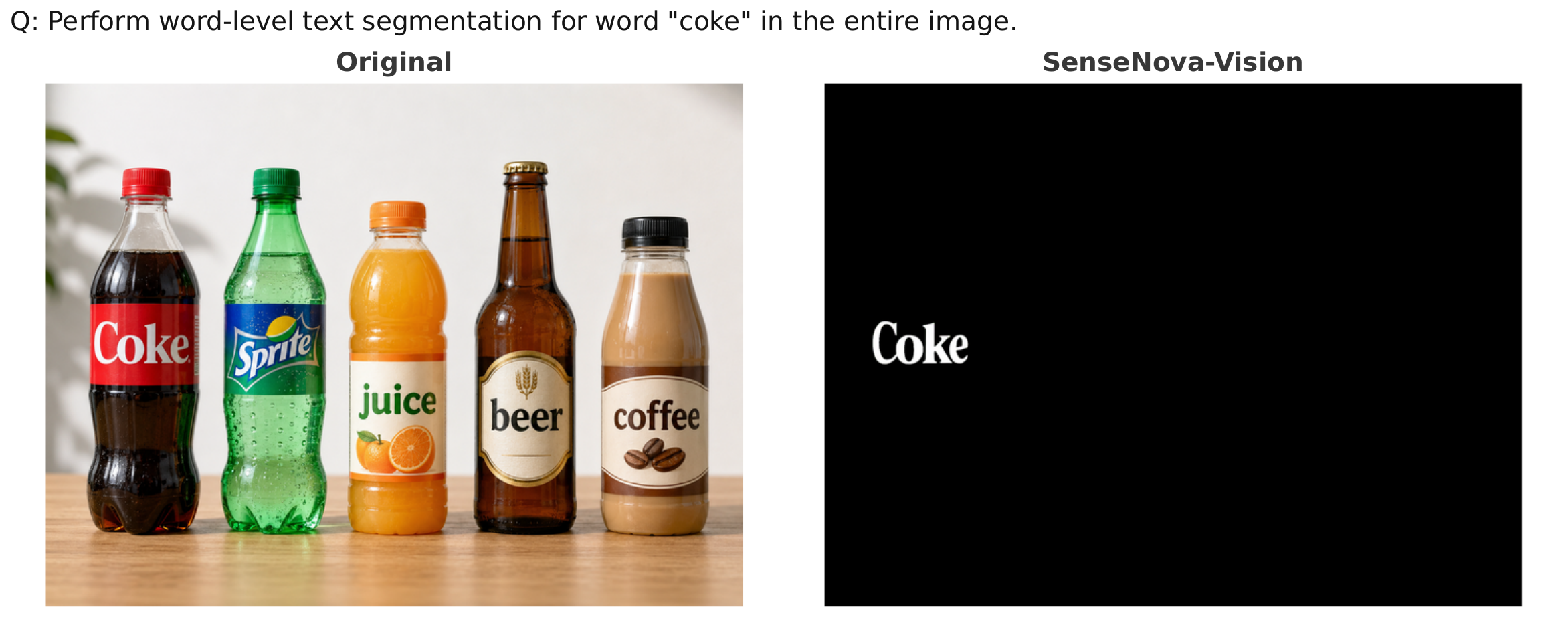}
        \caption{Word-level text segmentation.}
        \label{fig:text-segmentation-word}
    \end{subfigure}
    \hfill
    \begin{subfigure}{0.49\textwidth}
        \centering
        \includegraphics[width=\linewidth]{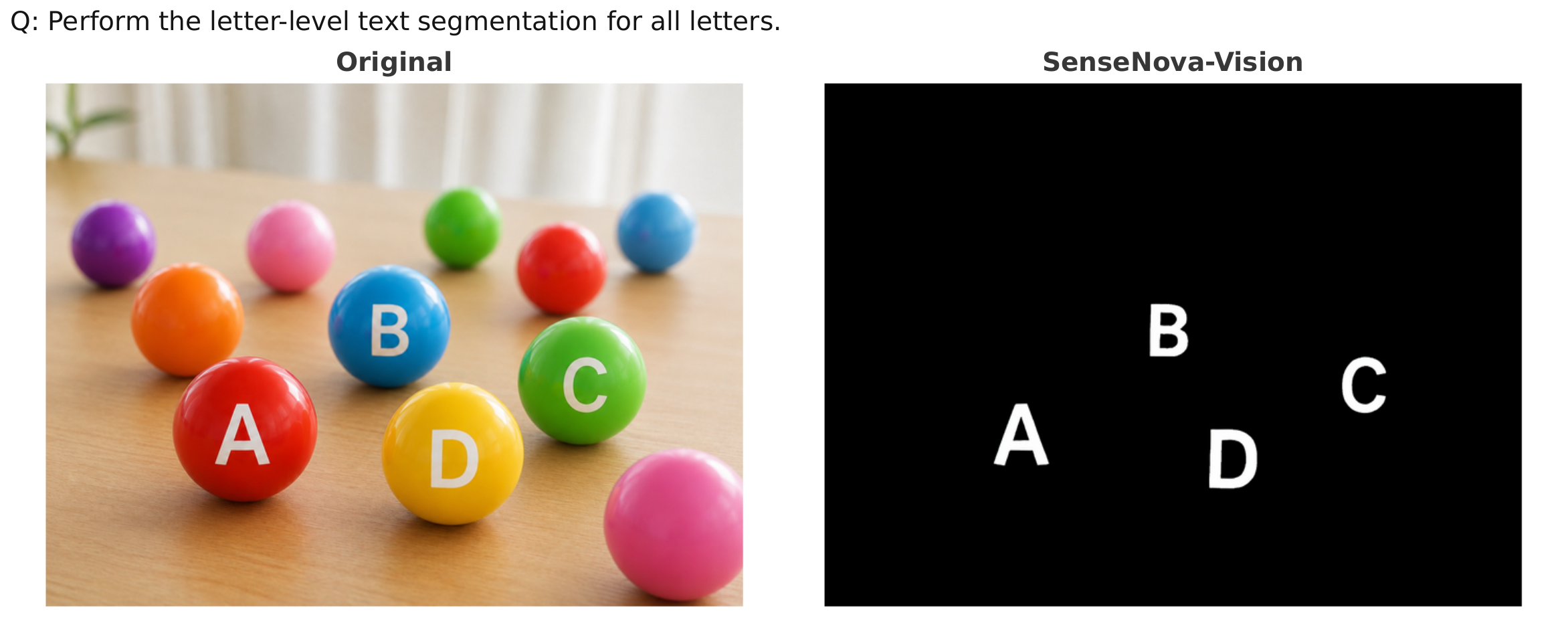}
        \caption{Letter-level text segmentation.}
        \label{fig:text-segmentation-letters}
    \end{subfigure}
    \vspace{-4pt}
    \caption{
        \textbf{Text segmentation from free-form language prompts.}
        (a) A word-level prompt asks the model to segment the queried word ``coke'' as a text-shaped region.
        (b) A letter-level prompt asks the model to segment all visible letters, requiring finer-grained localization and decomposition of text regions.
        In both cases, the generated masks align with the requested text targets.
    }
    \label{fig:text-segmentation-examples}
\end{figure}

Free-form color-coded mask generation probes whether language can define the mask representation protocol itself.
Here, color assignments are given through free-form language, which tests whether the model can handle user-described palettes after training on fixed class-color tags and exact RGB values.
As shown in Fig.~\ref{fig:freeform-color-mask-generation}, SenseNova-Vision generates mask-like outputs whose regions roughly follow the requested color assignments, although the colors and boundaries are not always exact.
This shows that language can control the mask representation protocol, not only the target regions to be segmented.

Text segmentation treats textual content, such as words or characters, as spatial mask targets and has been studied in scene text segmentation~\cite{xu2021rethinking,yu2024eaformer}.
Although SenseNova-Vision is not trained with text segmentation masks, OCR localization supervision can be recombined with mask generation to produce text-shaped regions.
As shown in Fig.~\ref{fig:text-segmentation-examples}, the model segments the queried word ``coke'' and can also generate masks for individual visible letters.
The target type of segmentation therefore broadens from conventional object or stuff regions to lexical units, including fine-grained or disconnected regions that jointly require recognition, localization, and mask rendering.

Overall, these qualitative probes suggest that free-form language-to-mask generation can extend segmentation along several dimensions: data distribution, task definition, mask representation protocol, and target type.
Although still qualitative and imperfect, these behaviors indicate that new mask-generation tasks can emerge from recombining recognition, localization, grounding, OCR, and rendering capabilities.

More importantly, even when supervision is collected from separate task domains, joint training under our formulation strengthens cross-modal correspondences, allowing the same underlying information to be represented and used flexibly across modalities.
Referring-style interactive segmentation and VGD segmentation show that spatial locations can be specified through either text or image cues and aligned with mask outputs.
OCR localization and text segmentation represent textual-spatial information in complementary forms: symbolic text with locations and grid-based visual masks.
Free-form color-coded mask generation shows that the model can generalize from RGB-specified mask protocols to natural-language color descriptions while connecting both to color-coded mask outputs, reflecting its ability to understand and use different expressions of color.
Together, these cross-modal correspondences point to unified multimodal generation as a way to jointly organize visual, linguistic, and spatial information within one model.

\section{Conclusion}
\label{sec:conclusion}

In this work, we present unified multimodal generation as a formulation for computer vision, analogous to the role of GPT-style generative modeling in NLP.
We present the SenseNova-Vision Corpus by casting heterogeneous computer vision annotations into text, image, and mixed text-and-image generation targets, and train SenseNova-Vision within the same formulation.
This conversion enables large-scale training of a single UMM across structured visual understanding, dense geometric prediction, segmentation, and multi-view visual geometry without task-specific heads.
The resulting model matches leading task-specialized systems and supports language-defined task variants, suggesting a scalable route for absorbing computer vision supervision into general-purpose foundation models.
The generative modeling of diverse 2D and 3D perception tasks may further cultivate implicit spatial understanding, connecting this paradigm naturally to the rapidly emerging frontiers of physical intelligence.

This formulation opens several directions for future work.
First, stronger in-context learning could further reduce the boundaries between task domains, enabling new visual tasks to be specified by examples, prompts, or mixed demonstrations beyond predefined datasets.
Second, extending unified multimodal generation from images to video would bring temporal dynamics and web-scale video supervision into foundation-model training.
Finally, scaling the corpus and model capacity, together with deeper integration with the strongest language models, may allow general-purpose foundation models to absorb richer visual and spatial knowledge from computer vision, moving toward world models that can perceive, reason about, and interact with the physical world.

\clearpage

\bibliographystyle{plainnat}
\bibliography{main}

\clearpage


\beginappendix
\lstset{
    basicstyle=\ttfamily\footnotesize,
    frame=single,
    float=htbp,
    breaklines=true,
    sensitive=false, 
    tabsize=2,
    lineskip=-2pt,    
    aboveskip=2pt,    
    belowskip=2pt,    
    xleftmargin=4pt   
}
\renewcommand{\topfraction}{0.95}
\renewcommand{\bottomfraction}{0.9}
\renewcommand{\textfraction}{0.05}
\renewcommand{\floatpagefraction}{0.85}
\setcounter{topnumber}{5}
\setcounter{bottomnumber}{3}
\setcounter{totalnumber}{8}
\setlength{\abovecaptionskip}{4pt}
\setlength{\belowcaptionskip}{2pt}
\captionsetup[table]{skip=4pt}
\section{SenseNova-Vision Corpus Construction Details}
\label{app:data_protocol_construction}

This appendix documents the construction details behind the SenseNova-Vision Corpus and expands the data protocol in Sec.~\ref{sec:data}.
Each training example is organized as one or more visual inputs, a natural-language instruction, and a target response represented as text, an image, or a mixed text-and-image output.
It then details data composition, processing pipelines, prompt design, and release resources for each task family.

\subsection{Structured visual understanding}
\textbf{Data Composition}
Structured visual understanding includes bounding-box detection, point detection, visual-prompt detection, referring detection, OCR localization, layout analysis, GUI grounding, and keypoint detection.
Because bounding-box, point, and visual-prompt detection share closely related source annotations, we summarize their integrated datasets in Table~\ref{tab:detection-visual-prompt-datasets}.
Other structured tasks, including OCR, layout analysis, GUI grounding, keypoint detection, and referring detection, are constructed from dedicated datasets, as shown in Table~\ref{tab:other-detection-datasets}.
The corresponding processing pipelines and data engines are described below.

\begin{table}[htbp]
    \centering
    \small
    \setlength{\tabcolsep}{5pt}
    \renewcommand{\arraystretch}{0.88}
    \begin{tabular}{l|l|l|l}
        \toprule
        Dataset & Task & Frames & Source \\
        \midrule
        APTv2~\cite{yang2023aptv2benchmarkinganimalpose} & BBox/Point & 28.0K/14.7K & Public/B2PC \\
        BDD100K~\cite{yu2020bdd100k} & BBox/Point/Visual & 70.0K/68.0K / 44.1K & Public/B2PC/Public \\
        Blood Cell~\cite{blood-cell-detection-1ekwu_dataset} & BBox/Visual & 0.2K/0.1K & Public/Public \\
        CARPK~\cite{Hsieh_2017_ICCV} & BBox/Visual & 1.0K/0.6K & Public/Public \\
        CrowdHuman~\cite{shao2018crowdhumanbenchmarkdetectinghuman} & BBox/Visual & 3.4K/2.1K & Public/Public \\
        DOTAv2~\cite{Ding_2022} & BBox/Point/Visual & 1.8K/1.7K/1.1K & Public/B2PC/Public \\
        DeepFashion~\cite{liuLQWTcvpr16DeepFashion} & BBox/Point & 191.0K/112.6K & Public/B2PC \\
        EgoObjects~\cite{zhu2023egoobjects} & BBox/Point & 78.0K/49.9K & Public/B2PC \\
        FAIR1M~\cite{SUN2022116} & BBox/Point/Visual & 16.0K/16.0K/10.1K & Public/B2PC/Public \\
        FSC147~\cite{amininaieni2023openworldtextspecifiedobjectcounting} & BBox/Point/Visual & 1.8K/3.6K/1.1K & GDE/Public/GDE \\
        FiftyOne~\cite{shubh303_dense_object_detection_fiftyone_2026} & BBox/Visual & 8.0K/5.0K & Public/Public \\
        Fish~\cite{solawetz_fish_dataset_2020} & Visual & 0.4K & Public \\
        Football~\cite{football-player-detection-kucab_dataset} & BBox/Visual & 0.8K/0.5K & Public/Public \\
        GroceryStore~\cite{klasson2019hierarchical} & BBox/Visual & 1.8K/1.1K & GDE/GDE \\
        HomeObjects-3k~\cite{Jocher_Ultralytics_Datasets_2025} & BBox/Visual & 2.2K/1.4K & Public/Public \\
        HumanParts~\cite{li2019detector} & BBox/Point & 12.0K/7.0K & Public/B2PC \\
        ImageNetPart~\cite{he2022partimagenet} & BBox/Point & 16.0K/10.2K & Public/B2PC \\
        Industrial Site Safety~\cite{chappieut_industrial_site_safety_detection_v1_2026} & BBox/Visual & 0.3K/0.2K & Public/Public \\
        LVIS Fruits \& Vegetables~\cite{henningheyen_lvis_fruits_and_vegetables_2026} & BBox/Visual & 6.7K/6.9K & Public/Public \\
        Locount~\cite{Cai2020Locount} & BBox/Visual & 34.0K/21.4K & Public/Public \\
        METU-ALET~\cite{METU_ALET} & BBox/Visual & 2.0K/1.3K & Public/Public \\
        NuImages~\cite{Fong2025nuScenesRP} & BBox/Point & 60.0K/55.0K & Public/B2PC \\
        OWOD~\cite{open_world_dense_object_detection} & BBox/Visual & 8.0K/5.0K & Public/Public \\
        Objects365~\cite{shao2019objects365} & BBox/Point/Visual & 1742.0K/1077.1K/1428.8K & Public/B2PC/Public \\
        PACO-LVIS~\cite{ramanathan2023paco} & BBox/Point & 45.0K/26.9K & Public/B2PC \\
        PixMo-Points~\cite{Deitke2024MolmoAP} & BBox/Point & 0.1K/1090.6K & GDE/Public \\
        S2TLD~\cite{yang2022scrdet++} & BBox/Visual & 5.0K/3.2K & Public/Public \\
        SA-1B~\cite{sam} & BBox/Point/Visual & 3119.0K/1949.4K/3116.6K & GDE/B2PC/GDE \\
        SKU110K~\cite{goldman2019dense} & BBox/Visual & 28.0K/17.6K & Public/Public \\
        Shoes~\cite{kukreti_shoe_dataset_2021} & BBox & 0.1K & Public \\
        TinyPerson~\cite{yu2020scale} & BBox/Visual & 1.5K/0.4K & Public/Public \\
        V3Det-OVD~\cite{wang2023v3det} & BBox/Point & 116.0K/60.8K & Public/B2PC \\
        VisDrone~\cite{zhu2021detection} & BBox/Point/Visual & 6.4K/6.4K/4.0K & Public/B2PC/Public \\
        WiderPerson~\cite{zhang2019widerperson} & BBox/Visual & 9.0K/5.7K & Public/Public \\
        Pill~\cite{Jocher_Ultralytics_Datasets_2024} & BBox/Visual & 0.1K/0.1K & Public/Public \\
        Sheep~\cite{aerial-sheep_dataset} & BBox/Visual & 3.6K/2.3K & Public/Public \\
        \bottomrule
    \end{tabular}
    \caption{
        Datasets for bounding-box, point and visual-prompt detection.
        Frames and sources follow the order of tasks.
        “GDE” means annotations from our grounding data engine; “B2PC” stands for points converted from bounding boxes via our pipeline.
    }
    \label{tab:detection-visual-prompt-datasets}
\end{table}

\begin{table}[htbp]
    \centering
    \small
    \setlength{\tabcolsep}{4pt}
    \renewcommand{\arraystretch}{0.88}
    \begin{tabular}{l|l|l|l||l|l|l|l}
        \toprule
        \multicolumn{4}{c||}{Task: Ref / OCR} & \multicolumn{4}{c}{Task: Keypoint / Layout / GUI} \\
        \cmidrule(lr){1-4} \cmidrule(lr){5-8}
        Dataset & Task & Frames & Source & Dataset & Task & Frames & Source \\
        \midrule
        HumanRef~\cite{zhang2024humanref} & Referring & 64.0K & Public
        & AP-10K~\cite{yu2021ap} & Keypoint & 9.7K & Public \\
        Objects365~\cite{shao2019objects365} & Referring & 3589.0K & RGDE
        & APT36K~\cite{yang2022apt} & Keypoint & 35.0K & Public \\
        OpenImages~\cite{OpenImages} & Referring & 4034.0K & RGDE
        & COCO2017~\cite{lin2014microsoft} & Keypoint & 56.0K & Public \\
        RefCOCO/+/g~\cite{yu2016modeling,kazemzadeh-etal-2014-referitgame} & Referring & 579.0K & Public
        & CrowdPose~\cite{li2018crowdpose} & Keypoint & 10.0K & Public \\
        RexVerse~\cite{jiang2024chatrextamingmultimodalllm} & Referring & 1632.0K & Public
        & Human-Art~\cite{ju2023human} & Keypoint & 33.0K & Public \\
        BLIP3-OCR-200M~\cite{Xue2024BLIP3AF} & OCR & 1582.0K & OCRE
        & MPII~\cite{MPII} & Keypoint & 17.0K & Public \\
        HierText~\cite{long2022towards,long2023icdar} & OCR & 32.8K & Public
        & MacaquePose V1~\cite{labuguen2021macaquepose} & Keypoint & 1.3K & Public \\
        ICDAR2013~\cite{ICDAR2013} & OCR & 0.2K & Public
        & OCHuman~\cite{zhang2019pose2seg} & Keypoint & 3.2K & Public \\
        ICDAR2015~\cite{ICDAR2015} & OCR & 1.8K & Public
        & CDLA~\cite{buptlihang2021cdla} & Layout & 5.0K & Public \\
        ICDAR2019~\cite{ICDAR2019} & OCR & 11.0K & Public
        & DocLayNet Core~\cite{pfitzmann2022doclaynet} & Layout & 69.0K & Public \\
        LSVT2019~\cite{sun2019lsvt} & OCR & 116.0K & Public
        & PubLayNet~\cite{zhong2019publaynet} & Layout & 335.0K & Public \\
        MTWI~\cite{he2018mtwi} & OCR & 40.0K & Public
        & TabRecSet~\cite{yang2023large} & Layout & 32.0K & Public \\
        RCTW~\cite{shi2017rctw} & OCR & 16.0K & Public
        & TableBank~\cite{li2019tablebank} & Layout & 260.0K & Public \\
        ReCTS~\cite{ReCTS} & OCR & 76.0K & Public
        & OS-Atlas~\cite{wu2024atlas} & GUI & 3101.6K & Public \\
        SROIE~\cite{SROIE} & OCR & 1.2K & Public
        & ShowUI Desktop~\cite{lin2024showui} & GUI & 1.5K & Public \\
        SynthText~\cite{Gupta16} & OCR & 1716.0K & Public & & & & \\
        TextOCR~\cite{singh2021textocr} & OCR & 42.0K & Public & & & & \\
        WildReceipt~\cite{sun2021spatial} & OCR & 2.4K & Public & & & & \\
        \bottomrule
    \end{tabular}
    \caption{Datasets covering referring detection, OCR localization, keypoint detection, document layout analysis and GUI understanding. “RGDE” and “OCRE” refer to outputs from our referring grounding and OCR data engines.}
    \label{tab:other-detection-datasets}
\end{table}

\begin{figure}[!t]
    \centering
    \includegraphics[width=\linewidth]{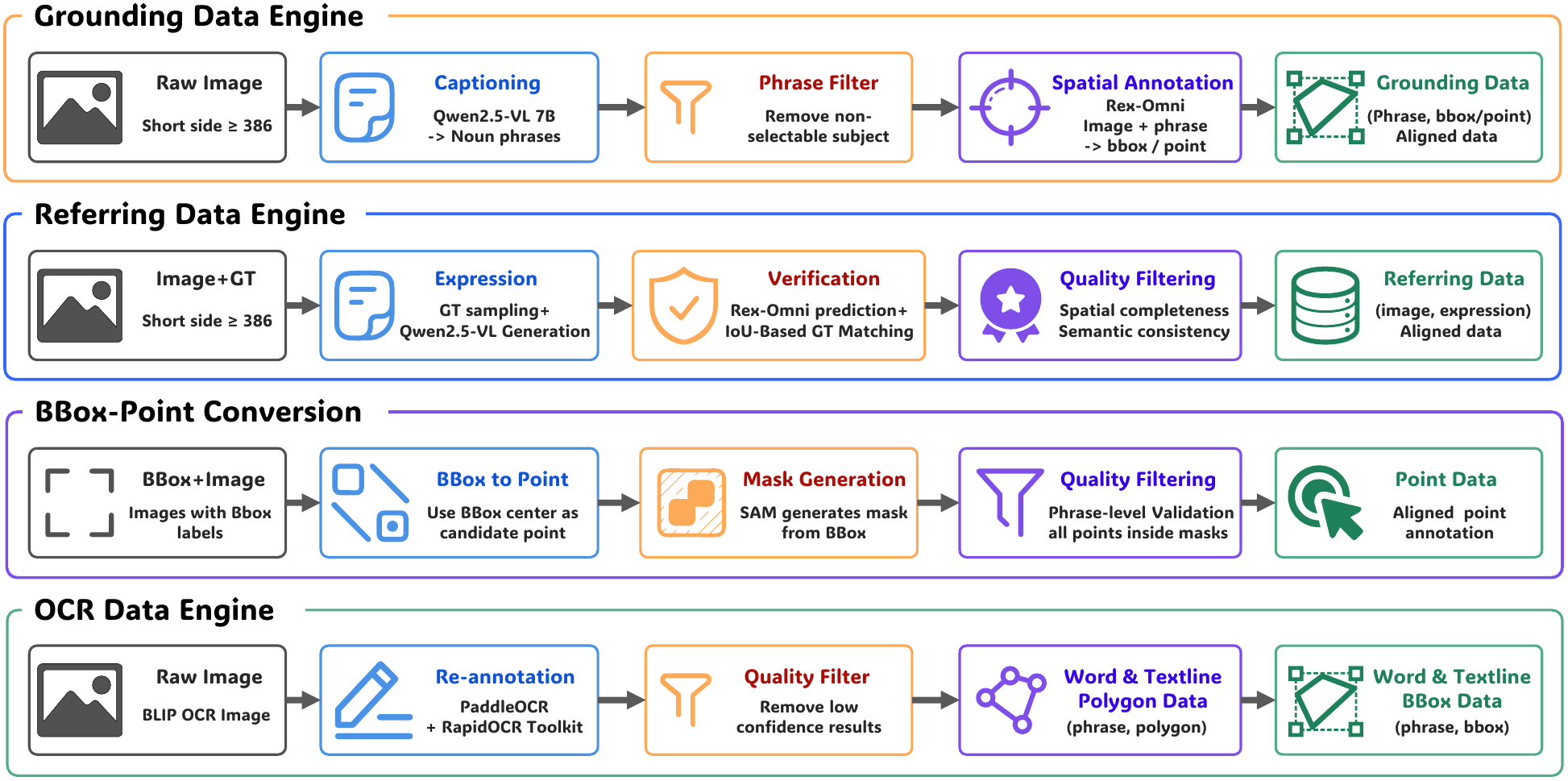}
    \caption{Overview of the proposed data construction pipelines, including the grounding data engine, referring grounding data engine, BBox-to-Point data conversion and OCR data engine for generating large-scale, high-quality spatial annotations.}
    \label{fig:structured-data-engine}
\end{figure}

\textbf{Data Processing}
We use relative image coordinates for all detection-related tasks, normalized by image width and height.
Each coordinate is rounded to three decimal places, and values outside the image boundary are clipped, such that the final coordinates remain within $[0.000, 0.999]$.
This unified representation is applied to bounding boxes, points and keypoints, reducing the discrepancy caused by different image resolutions and annotation formats.
To enhance the diversity and quality of the training data, particularly for challenging scenarios and multi-task learning, we develop several data engines that generate large-scale, high-quality annotations and facilitate the learning of visual-spatial correspondences.
The overall pipelines are illustrated in Figure~\ref{fig:structured-data-engine}.

\textbf{Prompt Design}
Following the prompt construction paradigm of InternVL~\cite{Chen2023InternVS}, each sample is organized as the dictionary structure template below.

\begin{lstlisting}
{
    "id": Sample_id,
    "image": [Input_image],
    "conversations": [
    {"from": "human", "value": "<image>Instruction"},
    {"from": "gpt", "value": "Target_text"}
  ]
}
\end{lstlisting}

\texttt{Sample\_id} uniquely identifies each training sample, while \texttt{Input\_image} records the path to the input RGB image.
The placeholder \texttt{<image>} marks where the data loader inserts the corresponding visual input, and \texttt{Instruction} specifies the task and the required output format.
\texttt{Target\_text} denotes the textual result containing task-specific semantic labels and spatial annotations such as bounding boxes, points, polygons, or keypoints.
For OCR-related samples, they additionally include the recognized text associated with each localized region.
To better illustrate task types in \texttt{Instruction} and standardize \texttt{Target\_text} formats, we adopt several lightweight markers for auxiliary representation, as shown in Table~\ref{tab:structure_markers}.

\begin{table}[!htbp]
  \centering
  \small
  \setlength{\tabcolsep}{5pt}
  \renewcommand{\arraystretch}{0.88}
  \begin{tabular}{l|l|p{0.52\textwidth}}
    \toprule
    Marker & Field & Function \\
    \midrule
    \texttt{\detokenize{<p>...</p>}} & Phrase or label & Object category, referring expression, OCR text, region description, or generated caption span. \\
    \midrule
    \texttt{\detokenize{<bbox>...</bbox>}} & Box & Normalized bounding-box coordinates. \\
    \midrule
    \texttt{\detokenize{<point>...</point>}} & Point & Normalized point coordinates. \\
    \midrule
    \texttt{\detokenize{<kpt>...</kpt>}} & Keypoint & Named keypoint coordinates. \\
    \midrule
    \texttt{\detokenize{<ins>...</ins>}} & Instance & Distinct individual identities. \\
    \midrule
    \texttt{\detokenize{<polygon>...</polygon>}} & Polygon & Polygon-based OCR result representation. \\
    \bottomrule
  \end{tabular}
  \caption{Representative delimiters used in structured instructions and target responses. }
  \label{tab:structure_markers}
\end{table}

For the \texttt{Instruction} field, one template is randomly selected from the corresponding task-specific pool during dataset construction.
We select a template for each distinct task, and the complete collection is summarized in Table~\ref{tab:detection-prompt-templates}.

\begin{table}[!htbp]
\centering
\footnotesize
\setlength{\tabcolsep}{3pt}
\renewcommand{\arraystretch}{0.88}
\begin{tabular}{p{0.18\textwidth}|p{0.74\textwidth}}
\toprule
Task & Instruction \\
\midrule
BBox Detection
& Identify objects from categories: \texttt{<p>category</p>}. Output detection results as text, each entry with category and bounding box coordinates. This format differs from depth/segmentation image outputs. \\
\midrule
Point Detection
& Detect objects from categories: \texttt{<p>category</p>}. Instead of visual outputs like depth maps or segmentation masks, return text detection list with labels and point coordinates. \\
\midrule
Referring Detection
& Detect all \texttt{<p>expression</p>} in the image. Instead of modified image outputs, return structured text with class labels and bounding box coordinates for each target. \\
\midrule
Visual Prompt Detection
& Given reference objects \texttt{<p>object1</p><bbox>[x0,y0,x1,y1]</bbox>}, detect all matching instances. Output structured text with category labels and bounding boxes in $[x_0,y_0,x_1,y_1]$. \\
\midrule
OCR
& Perform full text-line-level OCR on the whole image. Output structured list with each text line’s bounding box and extracted text content. \\
\midrule
Layout Analysis
& Detect objects from categories: \texttt{<p>category</p>}. Return text detection list with labels and bounding coordinates, no visual mask/depth outputs. \\
\midrule
GUI Grounding
& Detect all \texttt{<p>category</p>} instances. Output structured text with object class and precise location coordinates, distinct from image-format vision outputs. \\
\midrule
Keypoint Detection
& Predict \texttt{<keypoint names>} for all \texttt{<p>category</p>} instances. Output structured list with category and keypoint coordinates in $[x,y]$. \\
\bottomrule
\end{tabular}
\caption{Instruction templates for structured visual understanding tasks.}
\label{tab:detection-prompt-templates}
\end{table}

\textbf{Released Datasets}
Since most source datasets for structured visual understanding are publicly available, we release the generated or converted training examples produced by our data engines.

\subsection{Dense geometric prediction}
\textbf{Data Composition}
Public depth and normal datasets often share the same underlying data sources.
All datasets adopted in this task are summarized in Table~\ref{tab:depth-normal-datasets}.
Our method requires dense supervision, which is natively provided by synthetic datasets.
As LiDAR datasets typically contain sparse and imperfect annotations, we use MoGe-2~\cite{wang2025moge2} to generate dense pseudo labels.
Extra public datasets from related domains are also introduced to strengthen generalization and real-world performance, with dense labels generated consistently by MoGe-2.

\begin{table}[htbp]
  \centering
  \small
  \renewcommand{\arraystretch}{0.88}
  \setlength{\tabcolsep}{6pt}
    \begin{tabular}{l|l|l|l|l|l}
      \toprule
      Annotation & Dataset & Task & Domain & Frames & Source \\
      \midrule
      \multirow{6}{*}{Dense}
        & Hypersim~\cite{roberts2021hypersimphotorealisticsyntheticdataset}       & Depth/Normal & Indoor      & 61K/73K     & Synthetic \\
        & Virtual KITTI~\cite{gaidon2016virtualworldsproxymultiobject}  & Depth        & Outdoor     & 17K         & Synthetic \\
        &  InteriorVerse~\cite{zhu2022learningbasedinverserenderingcomplex}  & Normal       & Indoor      & 27K         & Synthetic \\
        & IRS~\cite{wang2021irslargenaturalisticindoor}            & Depth/Normal & Indoor      & 74K/103K    & Synthetic \\
        & TartanAir~\cite{wang2020tartanairdatasetpushlimits}      & Depth/Normal & Outdoor     & 118K/118K   & Synthetic \\
        & SceneNet RGB-D~\cite{mccormac2017scenenetrgbd5mphotorealistic} & Depth/Normal & Indoor      & 2273K/2380K & Synthetic \\
      \midrule
      \multirow{2}{*}{Sparse}
        & Taskonomy~\cite{zamir2018taskonomydisentanglingtasktransfer} & Depth/Normal & Indoor & 2000K/2000K & LiDAR \\
        & ScanNet++~\cite{yeshwanth2023scannethighfidelitydataset3d} & Depth/Normal & Indoor & 813K/813K   & LiDAR \\
      \midrule
      \multirow{3}{*}{None}
        & COCO 2017~\cite{lin2015microsoftcococommonobjects}  & Depth/Normal & In-the-wild & 78K/79K     & -- \\
        & SA-1B~\cite{sam}     & Depth/Normal & In-the-wild & 4442K/4462K & -- \\
        & Objects365~\cite{shao2019objects365} & Depth/Normal & In-the-wild & 1317K/1317K & -- \\
      \bottomrule
    \end{tabular}
  \caption{
    Summary of datasets used for depth and normal estimation tasks.
  }
  \label{tab:depth-normal-datasets}
\end{table}

\textbf{Data Processing}
For depth data processing, we set the valid range from 0.1\,m to 80\,m.
Dense annotations containing over 1 percent invalid pixels are discarded, and a 10 percent threshold is adopted for pseudo labels generated by MoGe-2~\cite{wang2025moge2}.
We employ inverse depth maps as supervision signals and convert all data to standardized three-channel RGB images for depth estimation.
For surface-normal data, we perform filtering only on MoGe-2 pseudo labels and exclude images with sky coverage above 10 percent.
The X, Y and Z components of valid normal values are assigned to the corresponding RGB channels to form the final training data.

\textbf{Prompt Design}
Dense geometric prediction is formulated as an image-to-image task, following the prompt construction paradigm of InternVL~\cite{chen2024internvl}.
\noindent
\begin{lstlisting}
{
    "id": Sample_id,
    "image": [Input_image, Output_image],
    "conversations": [
      {"from": "human", "value": "<image>Instruction"},
      {"from": "gpt", "value": "<image>"}
    ]
}
\end{lstlisting}
In contrast to structured visual understanding, dense geometric prediction uses image responses as target outputs.
 \texttt{Output\_image} refers to the path of the encoded depth or surface-normal target image.
Table~\ref{tab:depth-normal-prompt-templates} lists several illustrative examples of template instructions.

\begin{table}[htbp]
  \centering
  \footnotesize
  \setlength{\tabcolsep}{3pt}
  \renewcommand{\arraystretch}{0.88}
  \begin{tabular}{p{0.11\textwidth}|p{0.78\textwidth}}
    \toprule
    Task & Instruction \\
    \midrule
    Depth &
    Estimate relative depth for each pixel in the image.
    Closer objects should appear brighter and distant objects darker.
    The output is a grayscale image with pixel values ranging from 0 to 255. \\
    \midrule
    Normal &
    Estimate surface normals and encode them as an RGB image.
    The R, G and B channels correspond to the X, Y and Z direction components, respectively. \\
    \bottomrule
  \end{tabular}
  \caption{Instruction templates for dense geometric prediction.}
  \label{tab:depth-normal-prompt-templates}
\end{table}

\textbf{Released Datasets}
Since densely annotated datasets can be easily converted via our provided processing pipeline, we only release the training examples generated by MoGe-2~\cite{wang2025moge2}, including datasets with sparse annotations or no annotations in Table~\ref{tab:depth-normal-datasets}.

\subsection{Segmentation}
\textbf{Data Composition}
We split segmentation datasets into two groups according to their output representation formats.
As listed in Table~\ref{tab:seg-datasets-bin}, the first group covers referring segmentation, reasoning segmentation, and interactive segmentation tasks, which output binary masks containing a single foreground target region per image.
The second group, summarized in Table~\ref{tab:seg-datasets-pan}, consists of generic segmentation (semantic and panoptic segmentation) and grounded conversation generation (GCG) segmentation.
These tasks produce comprehensive panoptic masks that assign dense semantic and instance labels to all spatial regions throughout the entire image.

\begin{table}[htbp]
  \centering
  \small
  \setlength{\tabcolsep}{3pt}
  \renewcommand{\arraystretch}{0.88}
  \begin{tabular}{p{0.21\textwidth}|p{0.11\textwidth}|p{0.08\textwidth}||p{0.21\textwidth}|p{0.11\textwidth}|p{0.08\textwidth}}
    \toprule
    Dataset & Task & Frames & Dataset & Task & Frames \\
    \midrule
    RefCOCO/+/g~\cite{KazemzadehOrdonezMattenBergEMNLP14} & Referring & 545.7K
    & CIHP~\cite{gong2018instance} & Referring & 15.0K \\
    RefClef~\cite{liao2020realtimecrossmodalitycorrelationfiltering} & Referring & 220.7K
    & ATR~\cite{CO-CNN} & Referring & 11.3K \\
    GRefCOCO~\cite{GRES} & Referring & 353.3K
    & LIP~\cite{Gong_2017_CVPR} & Referring & 14.3K \\
    DOORS~\cite{article} & Referring & 6.5K
    & FAT-single/mixed \cite{tremblay2018falling}& Referring & 19.5K \\
    NDISPark~\cite{visapp21} & Referring & 0.1K
    & Fashionpedia \cite{jia2020fashionpedia}& Referring & 13.3K \\
    MinneApple~\cite{Hani_2020} & Referring & 0.5K
    & PartImageNet/Whole~\cite{he2021partimagenet} & Referring & 34.8K \\
    EYTH~\cite{Urooj_2018_CVPR} & Referring & 0.3K
    & WaterOVS~\cite{waterovs_huggingface} & Referring & 2.8K \\
    PST900~\cite{shivakumar2019pst900} & Referring & 0.3K
    & RaidaR-rainy/sunny~\cite{jin2021raidarrichannotatedimage} & Referring & 2.4K \\
    PSTRGB~\cite{Shivakumar2019PST900RC} & Referring & 1.4K
    & FSS-1000~\cite{Li_2020} & Referring & 3.7K \\
    SUIM~\cite{islam2020suim} & Referring & 2.5K
    & DAVIS 2017 \cite{Caelles_arXiv_2019} & Referring & 2.4K \\
    MyFood~\cite{aicrowd_food_recognition_challenge} & Referring & 1.0K
    & OCID-VLG \cite{tziafas2023language} & Referring & 3.0K \\
    CO-SKEL~\cite{8099896}& Referring & 0.3K
    & PIC \cite{liu2022pic}& Referring & 8.7K \\
    YouTube VOS 2022~\cite{xu2018youtubevoslargescalevideoobject} & Referring & 11.7K
    & LaPa \cite{liu2020new}& Referring & 8.1K \\
    MVTec D2S~\cite{10.1007/978-3-030-01249-6_35} & Referring & 2.5K
    & DeepFashion2 \cite{DeepFashion2} & Referring & 20.8K \\
    VizWiz-FewShot~\cite{tseng2022vizwizfewshotlocatingobjectsimages} & Referring & 3.0K
    & MattingHumanHalf \cite{aisegment2019matting}& Referring & 6.9K \\
    Trans10K~\cite{xie2020segmentingtransparentobjectswild} & Referring & 2.7K
    & ReaSeg \cite{lisa,yang2023improved}& Reasoning & 132.6K \\
    & & & Coco\_interactive\_psalm \cite{zhang2025psalm} & Interactive & 400.0K \\
    \bottomrule
  \end{tabular}
  \caption{Binary-mask segmentation datasets used for training.}
  \label{tab:seg-datasets-bin}
\end{table}

\begin{table}[htbp]
  \centering
  \small
  \setlength{\tabcolsep}{3pt}
  \renewcommand{\arraystretch}{0.88}
  \begin{tabular}{p{0.21\textwidth}|p{0.11\textwidth}|p{0.1\textwidth}||p{0.21\textwidth}|p{0.11\textwidth}|p{0.1\textwidth}}
    \toprule
    Dataset & Task & Frames & Dataset & Task & Frames \\
    \midrule
    COCONut-XL~\cite{coconut2024cvpr} & Generic & 587.3K
    & IDD-1/2~\cite{varma2019idd} & Generic & 14.0K \\
    Cityscapes~\cite{Cordts2016Cityscapes, Cordts2015Cvprw} & Generic/GCG & 3.0K/2.1K
    & IDDAv3~\cite{alberti2020idda} & Generic/GCG & 6.2K/1.1K \\
    Hypersim~\cite{roberts:2021} & Generic/GCG & 46.8K/30.4K
    & Mapillary Vistas~\cite{garg2025mapillaryvistasvalidationfinegrained} & Generic & 18.0K \\
    EntityV2~\cite{Qi_2023_EntitySeg} & Generic/GCG & 31.7K/6.1K
    & NuScenes~\cite{fong2021panoptic} & Generic & 65.5K \\
    Trashcan~\cite{hong2020trashcansemanticallysegmenteddatasetvisual} & Generic & 5.9K
    & 51WORLD~\cite{51world_dataone_synthetic_2025} & Generic/GCG & 11.6K/3.5K \\
    Pidray~\cite{zhang2022pidraylargescalexraybenchmark} & Generic & 29.5K
    & StreetHazards~\cite{hendrycks2019anomalyseg} & Generic/GCG & 6.2K/1.0K \\
    ZeroWaste-f~\cite{Bashkirova_2022_CVPR} & Generic & 2.9K
    & KITTI~\cite{step_2021} & Generic/GCG & 5.0K/3.5K \\
    LVIS~\cite{gupta2019lvis} & Generic & 56.7K
    & TAS500~\cite{metzger2021finegraineddatasetefficientsemantic} & Generic/GCG & 0.4K/0.3K \\
    UDD5/6~\cite{chen2018large} & Generic/GCG & 0.2K/0.1K
    & TTPLA~\cite{abdelfattah2020ttpla} & Generic & 1.2K \\
    LoveDA~\cite{Wang2021LoveDAAR} & Generic/GCG & 2.5K/1.6K
    & VIPSeg~\cite{miao2021vspw} & Generic/GCG & 66.8K/59.5K \\
    GranDf~\cite{Rasheed_2024_CVPR} & GCG & 1.0K
    & RefCOCOg~\cite{kazemzadeh-etal-2014-referitgame} & GCG & 19.3K \\
    PSG~\cite{yang2022psg} & GCG & 27.8K
    & Flickr30k~\cite{plummer2015flickr30k} & GCG & 148.2K \\
    \bottomrule
  \end{tabular}
  \caption{Generic and GCG segmentation datasets used for training.}
  \label{tab:seg-datasets-pan}
\end{table}

\textbf{Data Processing}
For all datasets presented in Table~\ref{tab:seg-datasets-bin}, we standardize all segmentation annotations into binary masks.
In these standardized binary masks, pixel values of $(255,255,255)$ denote foreground target regions, while pixels with $(0,0,0)$ correspond to image background areas.
For datasets summarized in Table~\ref{tab:seg-datasets-pan}, we first parse instance-level segmentation annotations within each image to generate individual masks for all existing object instances.
Each instance is then assigned a unique RGB color following a predefined color sampling strategy.
All colored instance masks are finally blended onto a pure black background to construct the final panoptic segmentation mask.
Specifically, we predefine 200 color anchors in the discrete RGB cube $[0,255]^3$ using greedy farthest-point sampling, which iteratively maximizes the minimum Euclidean distance from each new anchor to the previously selected anchors, thus improving visual distinguishability among instance colors.
The resulting anchors are sorted by their RGB values and assigned sequential indices.
For an image with $K$ annotated instances, we deterministically sample $K+1$ approximately evenly spaced colors from the ordered palette, assign the first $K$ colors to instances in processing order, and reserve the final black color $(0,0,0)$ for the background.

\textbf{Prompt Design}
For binary-mask segmentation tasks, including referring segmentation and reasoning segmentation, we employ the prompt template detailed below.
Unlike the previous tasks, we adopt \texttt{Binary\_mask} to store the file paths of binary masks.

\begin{lstlisting}
{
    "image": Input_image,
    "seg": Binary_mask,
    "conversations": [
        {"from": "human",
         "value": "<image>Instruction"},
        {"from": "gpt",
         "value": "Sure, <SEG>.",
        }
    ]
}
\end{lstlisting}

For interactive segmentation in the binary-mask group, we extend the prompt template with an extra visual prompt image. \texttt{Prompt\_image} denotes the file path of the visual prompt image, and \texttt{Prompt\_type} specifies its category, where \texttt{<prompt\_type>} acts as the corresponding symbolic marker.

\begin{lstlisting}
{
    "image": [Input_image, Prompt_image],
    "seg": Binary_mask,
    "visual_prompt_type": Prompt_type,
    "conversations": [
        {"from": "human",
         "value": "<image>Instruction<prompt_type><image>"},
        {"from": "gpt",
         "value": "Sure, <SEG>.",
        }
    ]
}
\end{lstlisting}

For generic segmentation and GCG segmentation, we employ a distinct prompt template as detailed below.

\begin{lstlisting}
{
    "id": Sample_id,
    "image": Input_image,
    "caption": Description,
    "seg": Panoptic_mask,
    "num_instances": Instance_count, (Only for Generic Segmentation)
    "conversations": [
        {"from": "human",
         "value": "<image>Instruction"},
        {"from": "gpt",
         "value": "Target_text<SEG>",}
    ]
}
\end{lstlisting}

\texttt{Panoptic\_mask} denotes the file path of the color-coded target mask.
Optional \texttt{Instance\_count} records the number of instances in the mask. \texttt{Description} captures descriptive textual content for the input image. \texttt{Target\_text} holds the target textual outputs, which enumerate all individual instances alongside their assigned distinct colors.
Inspired by ConsistCompose~\cite{Shi_2026_CVPR}, we adopt a color-instance binding format, where each instance is explicitly paired with a unique color.
Specifically, we use lightweight markers to standardize input and output formatting, as shown in Table~\ref{tab:segment_markers}.

\begin{table}[htbp]
  \centering
  \small
  \setlength{\tabcolsep}{5pt}
  \renewcommand{\arraystretch}{0.88}
  \begin{tabular}{l|l|p{0.52\textwidth}}
    \toprule
    Marker & Field & Function \\
    \midrule
    \texttt{\detokenize{<p>...</p>}} & Phrase or label & Instance category. \\
    \midrule
    \texttt{\detokenize{<box>}} & Prompt\_type & Bounding box visual prompt type. \\
    \midrule
    \texttt{\detokenize{<point>}} & Prompt\_type & Point visual prompt type. \\
    \midrule
    \texttt{\detokenize{<mask>}} & Prompt\_type & Mask visual prompt type. \\
    \midrule
    \texttt{\detokenize{<scribble>}} & Prompt\_type & Scribble visual prompt type. \\
    \midrule
    \texttt{\detokenize{<color>...</color>}} & Color & Instance-specific RGB color value. \\
    \bottomrule
  \end{tabular}
  \caption{Representative delimiters used in segmentation tasks.}
  \label{tab:segment_markers}
\end{table}

Segmentation tasks are formulated with task-specific instructions that specify the segmentation target, output format, and interaction when required.
For each distinct segmentation task, we extract one representative sample from its corresponding instruction pool for illustration, as shown in Table~\ref{tab:segmentation-prompt-templates}.

\begin{table}[htbp]
  \centering
  \footnotesize
  \setlength{\tabcolsep}{3pt}
  \renewcommand{\arraystretch}{0.88}
  \begin{tabular}{p{0.14\textwidth}|p{0.74\textwidth}}
    \toprule
    Task & Instruction \\
    \midrule
    Generic Seg &
    Generate panoptic segmentation masks for categories: \texttt{<p>person</p>}, ..., \texttt{<p>rug</p>}. Locate all instances, assign colors strictly as \texttt{<p>instance-no<color>(R,G,B)</color></p>}, output panoptic masks. \\
    \midrule
    GCGSeg &
    Briefly describe image content, return interleaved segmentation masks matching each part of your description. \\
    \midrule
    RefSeg &
    Visualize binary segmentation of \texttt{<p>person standing</p>} in the image. \\
    \midrule
    ReaSeg &
    Output segmentation mask for \texttt{<p>the ball that can only be hit into the hole at last</p>}. \\
    \midrule
    InterSeg &
    Generate segmentation masks guided by reference regions \texttt{<point><image>}. \\
    \bottomrule
  \end{tabular}
  \caption{Instruction templates for segmentation tasks.}
  \label{tab:segmentation-prompt-templates}
\end{table}

\textbf{Released Datasets}
Since binary-mask datasets can be reconstructed from public annotations using our conversion pipeline, we release the generated and curated resources for generic and GCG segmentation.

\subsection{Multi-view visual geometry}
\textbf{Data Composition}
Multi-view reconstruction and camera pose estimation are built from closely related multi-view geometric data sources.
All adopted datasets are summarized in Table~\ref{tab:multiview-3d-datasets}.
Similar to depth and surface-normal prediction, multi-view reconstruction requires dense geometric supervision.
We leverage LingBot-Depth~\cite{lingbotdepth} to complete sparse depth data and generate dense point maps.

\begin{table}[htbp]
  \centering
  \small
  \renewcommand{\arraystretch}{0.88}
  \setlength{\tabcolsep}{6pt}
    \begin{tabular}{l|l|l|l|l|l}
      \toprule
      Annotation & Dataset & Task & Domain & Frames & Source \\
      \midrule
      \multirow{12}{*}{Dense}
        & Hypersim~\cite{roberts2021hypersimphotorealisticsyntheticdataset}        & Reconstruction/Camera pose & Indoor  & 44K/44K  & Synthetic  \\
        & IRS~\cite{wang2021irslargenaturalisticindoor}             & Reconstruction/Camera pose & Indoor  & 66K/66K & Synthetic \\
        & TartanAir~\cite{wang2020tartanairdatasetpushlimits}       & Reconstruction/Camera pose & Outdoor  & 246K/246K  & Synthetic \\
        & SceneNet RGB-D~\cite{mccormac2017scenenetrgbd5mphotorealistic}  & Reconstruction/Camera pose & Indoor  & 5058K/5058K & Synthetic  \\
        & AriaSyntheticENV~\cite{avetisyan2024scenescriptreconstructingscenesautoregressive} & Reconstruction            & Indoor  & 5919K & Synthetic  \\
        & BlendedMVG~\cite{yao2020blendedmvslargescaledatasetgeneralized}      & Reconstruction             & Outdoor  & 115K & Synthetic \\
        & MegaSynth~\cite{jiang2025megasynthscaling3dscene}       & Reconstruction/Camera pose & Outdoor  & 323K/323K & Synthetic \\
        & MvsSynth~\cite{DeepMVS}        & Reconstruction / Camera pose & Outdoor  & 12K/12K & Synthetic \\
        & OmniObject3D~\cite{wu2023omniobject3dlargevocabulary3dobject}      & Reconstruction/Camera pose & Object-centric  & 595K/595K  & Synthetic \\
        & Objaverse~\cite{deitke2022objaverseuniverseannotated3d}       & Camera pose       & Indoor  & 1400K  & Synthetic \\
        & CO3Dv2~\cite{reizenstein2021commonobjects3dlargescale}          & Camera pose       & Object-centric  & 1500K  & Colmap \\
        & DeMoN-MVE~\cite{Ummenhofer_2017}       & Camera pose       & Outdoor  & 20K  & Synthetic \\
      \midrule
      \multirow{4}{*}{Sparse}
        & ScanNetV2~\cite{dai2017scannetrichlyannotated3dreconstructions} & Reconstruction/Camera pose & Indoor  & 241K/241K & LiDAR \\
        & ScanNet++~\cite{yeshwanth2023scannethighfidelitydataset3d} & Reconstruction/Camera pose & Indoor & 813K/813K & LiDAR \\
        & DL3DV~\cite{ling2023dl3dv10klargescalescenedataset}     & Reconstruction/Camera pose & In-the-wild & 3462K/3462K & Colmap \\
        & WildRGB-D~\cite{xia2024rgbdobjectswildscaling}  & Reconstruction/Camera pose & In-the-wild & 8026K/8026K & LiDAR \\
      \bottomrule
    \end{tabular}
  \caption{
    Summary of datasets used for multi-view visual geometry across reconstruction and camera pose estimation tasks.
  }
  \label{tab:multiview-3d-datasets}
\end{table}

\textbf{Data Processing}
For multi-view reconstruction, we build data examples at the scene level.
Each scene consists of a sequence of views stored in a format convertible to point maps in a shared coordinate frame, including RGB images, depth maps, and camera trajectory.
To provide data suitable for supervision without relying on masking strategies, we ensure that all input views have dense depth maps.
Specifically, we first filter out views whose depth maps contain more than two-thirds invalid pixels.
We then use LingBot-Depth~\cite{lingbotdepth} to complete and refine sparse depth maps.
Additionally, depth values exceeding a dataset-specific threshold are treated as invalid, where the threshold ranges from 30 to 80 meters depending on the source dataset.
For camera pose estimation, we predict camera-to-world extrinsic parameters for all subsequent frames with respect to the initial reference frame.
These parameters are decoupled into rotational and translational components.
Specifically, rotation is represented via quaternions, while translation is formulated as a hybrid representation consisting of a unit direction vector and a scale magnitude.
Furthermore, we adopt reserved numerical tokens from \texttt{<-1000>} to \texttt{<1000>} that span the numerical range of -1000 to 1000.
Following this design, all quaternion components and unit direction vector elements are multiplied by a factor of 1000.
For the scale term, we constrain its valid physical range to [0.01\,m, 10\,m], and define the token \texttt{<1>} to correspond to an actual length of 1\,cm.

\textbf{Prompt Design}
Distinct from other vision tasks, multi-view reconstruction produces training examples in an online manner.
For this reason, we embed scene-level dataset metadata into the prompt, whose detailed structure is elaborated as follows.

\begin{lstlisting}[float=htb]
{
    "id": Scene_id,
    "image": [Image_sequence],
    "depth": [Depth_sequence],
    "trajectory": [Camera_trajectory],
    "conversations": [
      {"from": "human", "value": "Instruction"},
      {"from": "human", "value": "Instruction"},
      ...
    ]
}
\end{lstlisting}

\texttt{Scene\_id} denotes the unique identifier for each scene, while \texttt{Image\_sequence} stores all image paths in that scene.
For paired geometric information, \texttt{Depth\_sequence} holds depth map paths matched to each image entry, and \texttt{Camera\_trajectory} provides frame-wise corresponding camera parameters.

As a multi-image understanding task, camera pose estimation is defined via the prompt template presented below.

\begin{lstlisting}
{
    "id": Sample_id,
    "image": [Image_sequence],
    "conversations": [
      {"from": "human", "value": "<image>...<image>Instruction"},
      {"from": "gpt", "value": "<camerapose_sequence>"}
    ]
}
\end{lstlisting}

 \texttt{Image\_sequence} holds the complete collection of image paths for all frames within the input image sequence.
The repetition count of \texttt{<image>} placeholders in \texttt{<image>}...\texttt{<image>} aligns with the length of \texttt{Image\_sequence}. \texttt{Camerapose\_sequence} contains relative camera poses from the second frame to the final frame, and all entries follow the identical format below.
\begin{lstlisting}
<frame><quat><x><y><z><w></quat><offset><x><y><z></offset><scale><value></scale></frame>
\end{lstlisting}
To improve format consistency and numerical precision, we encode camera-pose records with reserved tokens following the rules specified in Table~\ref{tab:pose_token_markers}.

We present partial instructions adopted for multi-view visual geometry in Table~\ref{tab:camera-pose-instruction-format}.
\begin{table}[htbp]
  \centering
  \footnotesize
  \setlength{\tabcolsep}{3pt}
  \renewcommand{\arraystretch}{0.88}
  \begin{tabular}{p{0.23\textwidth}|p{0.17\textwidth}|p{0.52\textwidth}}
    \toprule
    Marker & Field & Function \\
    \midrule
    \texttt{<-1000>}...\texttt{<1000>} & Numerical values & These dedicated special tokens encode all scalar values within the range $[-1000, 1000]$. \\
    \midrule
    \texttt{<frame>}...\texttt{</frame>} & View entry & Special token delimiting a per-view camera-pose record in multi-view visual geometry tasks. \\
    \midrule
    \texttt{<quat>}...\texttt{</quat>} & Rotation & Special token delimiting quaternion rotation parameters for camera pose. \\
    \midrule
    \texttt{<offset>}...\texttt{</offset>} & Translation & Special token delimiting camera translation or offset parameters. \\
    \midrule
    \texttt{<scale>}...\texttt{</scale>} & Scale & Special token delimiting scale metadata used to decode camera pose or geometry. \\
    \bottomrule
  \end{tabular}
  \caption{Reserved tokens for camera pose encoding.}
  \label{tab:pose_token_markers}
\end{table}

\begin{table}[htbp]
  \centering
  \footnotesize
  \setlength{\tabcolsep}{3pt}
  \renewcommand{\arraystretch}{0.88}
  \begin{tabular}{p{0.16\textwidth}|p{0.74\textwidth}}
    \toprule
    Task & Instruction \\
    \midrule
    Camera pose &
    Use the first frame as the reference frame, and output the relative pose of all subsequent frames with respect to it, following the original input order.
    Rotation is represented by a quaternion in the format \texttt{<quat>}[x,y,z,w]\texttt{</quat>}.
    Translation direction is represented by a unit vector in the format \texttt{<offset>}[x,y,z]\texttt{</offset>}, which only encodes directional information and has no absolute physical meaning.
    Translation scale is represented by a numerical value in the format \texttt{<scale>}value\texttt{</scale>}.
    The result of each target frame is enclosed in \texttt{<frame>}...\texttt{</frame>} tags, with no extra characters, spaces, or line breaks outside the required tags. \\
    \midrule
    \multirow{2}{=}{Multi-view reconstruction}
    & Reconstruct a scene from multiple input images and output one dense 3D coordinate map per view, all aligned to the first camera's perspective. \\
    \cmidrule(lr){2-2}
    & From RGB images, reconstruct a 3D scene and produce XYZ point maps aligned in a shared coordinate system. \\
    \bottomrule
  \end{tabular}
  \caption{Instruction templates for multi-view visual geometry.}
  \label{tab:camera-pose-instruction-format}
\end{table}

\textbf{Released Datasets}
Camera-pose examples can be assembled directly from public multi-view datasets with camera trajectories.
For reconstruction, we publicly release the sparsely annotated datasets preprocessed with LingBot-Depth~\cite{lingbotdepth}, while densely annotated datasets can be reconstructed from the released conversion pipeline.

\subsection{SN-VC-50M Release Summary}
We summarize the released SN-VC-50M resources in Table~\ref{tab:open-source-datasets}.
The release spans four task families and covers 73 dataset-task entries across 10 task types.
Specifically, the released data includes 18.9M frames for structured visual understanding, 1.3M frames for segmentation, 17.3M frames for dense geometric prediction, and 12.5M frames for multi-view visual geometry.
To avoid duplicating raw RGB images from public datasets, the released training examples retain image file paths instead of redistributing the original images.

\begin{table}[htb]
  \centering
  \small
  \renewcommand{\arraystretch}{0.88}
  \setlength{\tabcolsep}{3pt}
  \begin{tabular}{p{0.16\textwidth}|p{0.14\textwidth}|p{0.16\textwidth}||p{0.16\textwidth}|p{0.14\textwidth}|p{0.16\textwidth}}
    \toprule
    Dataset & Task & Frames & Dataset & Task & Frames \\
    \midrule
    \multicolumn{3}{c||}{\textbf{Structured visual understanding}} & \multicolumn{3}{c}{\textbf{Segmentation}} \\
    \cmidrule(r){1-3}\cmidrule(l){4-6}
    APTv2~\cite{yang2023aptv2benchmarkinganimalpose} & Point & 14.7K & COCONut-XL~\cite{coconut2024cvpr} & Generic & 587.3K \\
    BDD100K~\cite{yu2020bdd100k} & Point & 68.0K & Cityscapes~\cite{Cordts2016Cityscapes, Cordts2015Cvprw} & Generic/GCG & 3.0K/2.1K \\
    DOTAv2~\cite{Ding_2022} & Point & 1.7K & Hypersim~\cite{roberts:2021} & Generic/GCG & 46.8K/30.4K \\
    DeepFashion~\cite{liuLQWTcvpr16DeepFashion} & Point & 112.6K & EntityV2~\cite{Qi_2023_EntitySeg} & Generic/GCG & 31.7K/6.1K \\
    EgoObjects~\cite{zhu2023egoobjects} & Point & 49.9K & Trashcan~\cite{hong2020trashcansemanticallysegmenteddatasetvisual} & Generic & 5.9K \\
    FAIR1M~\cite{SUN2022116} & Point & 16.0K & Pidray~\cite{zhang2022pidraylargescalexraybenchmark} & Generic & 29.5K \\
    FSC147~\cite{amininaieni2023openworldtextspecifiedobjectcounting} & BBox/Visual & 1.8K/1.1K & ZeroWaste-f~\cite{Bashkirova_2022_CVPR} & Generic & 2.9K \\
    GroceryStore~\cite{klasson2019hierarchical} & BBox/Visual & 1.8K/1.1K & LVIS~\cite{gupta2019lvis} & Generic & 56.7K \\
    HumanParts~\cite{li2019detector} & Point & 7.0K & IDD-1/2~\cite{varma2019idd} & Generic & 14.0K \\
    ImageNetPart~\cite{he2022partimagenet} & Point & 10.2K & IDDAv3~\cite{alberti2020idda} & Generic/GCG & 6.2K/1.1K \\
    NuImages~\cite{Fong2025nuScenesRP} & Point & 55.0K & Mapillary Vistas~\cite{garg2025mapillaryvistasvalidationfinegrained} & Generic & 18.0K \\
    Objects365~\cite{shao2019objects365} & Point/Referring & 1077.1K/3589.0K & NuScenes~\cite{fong2021panoptic} & Generic & 65.5K \\
    PACO-LVIS~\cite{ramanathan2023paco} & Point & 26.9K & 51WORLD~\cite{51world_dataone_synthetic_2025} & Generic/GCG & 11.6K/3.5K \\
    PixMo-Points~\cite{Deitke2024MolmoAP} & BBox & 0.1K & StreetHazards~\cite{hendrycks2019anomalyseg} & Generic/GCG & 6.2K/1.0K \\
    SA-1B~\cite{sam} & BBox/Point/ & 3119.0K/1949.4K/ & KITTI~\cite{step_2021} & Generic/GCG & 5.0K/3.5K \\
    & Visual & 3116.6K & TAS500~\cite{metzger2021finegraineddatasetefficientsemantic} & Generic/GCG & 0.4K/0.3K \\
    V3Det-OVD~\cite{wang2023v3det} & Point & 60.8K & UDD5/6~\cite{chen2018large} & Generic/GCG & 0.2K/0.1K \\
    VisDrone~\cite{zhu2021detection} & Point & 6.4K & TTPLA~\cite{abdelfattah2020ttpla} & Generic & 1.2K \\
    OpenImages~\cite{OpenImages} & Referring & 4034.0K & LoveDA~\cite{Wang2021LoveDAAR} & Generic/GCG & 2.5K/1.6K \\
    BLIP3-OCR~\cite{Xue2024BLIP3AF} & OCR & 1582.0K & VIPSeg~\cite{miao2021vspw} & Generic/GCG & 66.8K/59.5K \\
    & & & GranDf~\cite{Rasheed_2024_CVPR} & GCG & 1.0K \\
    & & & RefCOCOg~\cite{kazemzadeh-etal-2014-referitgame} & GCG & 19.3K \\
    & & & PSG~\cite{yang2022psg} & GCG & 27.8K \\
    & & & Flickr30k~\cite{plummer2015flickr30k} & GCG & 148.2K \\
    \cmidrule{1-6}
    \multicolumn{3}{c||}{\textbf{Dense geometric prediction}} & \multicolumn{3}{c}{\textbf{Multi-view visual geometry}} \\
    \cmidrule(r){1-3}\cmidrule(l){4-6}
    Taskonomy~\cite{zamir2018taskonomydisentanglingtasktransfer} & Depth/Normal & 2000.0K/2000.0K & ScanNetV2~\cite{dai2017scannetrichlyannotated3dreconstructions} & Reconstruction & 241.0K \\
    ScanNet++~\cite{yeshwanth2023scannethighfidelitydataset3d} & Depth/Normal & 813.0K/813.0K & ScanNet++~\cite{yeshwanth2023scannethighfidelitydataset3d} & Reconstruction & 813.0K \\
    COCO2017~\cite{lin2015microsoftcococommonobjects} & Depth/Normal & 78.0K/79.0K & DL3DV~\cite{ling2023dl3dv10klargescalescenedataset} & Reconstruction & 3462.0K \\
    SA-1B~\cite{sam} & Depth/Normal & 4442.0K/4462.0K & WildRGB-D~\cite{xia2024rgbdobjectswildscaling} & Reconstruction & 8026.0K \\
    Objects365~\cite{shao2019objects365} & Depth/Normal & 1317.0K/1317.0K & & & \\
    \bottomrule
  \end{tabular}
  \caption{
    Overview of the SN-VC-50M Corpus open-source release.
    Datasets are grouped by task family, with each entry reporting the released task type and frame count.
  }
  \label{tab:open-source-datasets}
\end{table}

\section{Additional Qualitative Results}
\label{app:additional_qualitative_results}

This section provides additional qualitative examples that complement the main quantitative evaluation.
The examples are intended to show the breadth of SenseNova-Vision across text, image, and mixed text-and-image tasks, including structured visual understanding, dense geometric prediction, segmentation, and multi-view visual geometry.
They also illustrate how the same model follows different task instructions while keeping outputs decodable under the protocols described in Appendix~\ref{app:data_protocol_construction}.

\begin{figure}[t]
    \centering
    \includegraphics[width=0.9\linewidth]{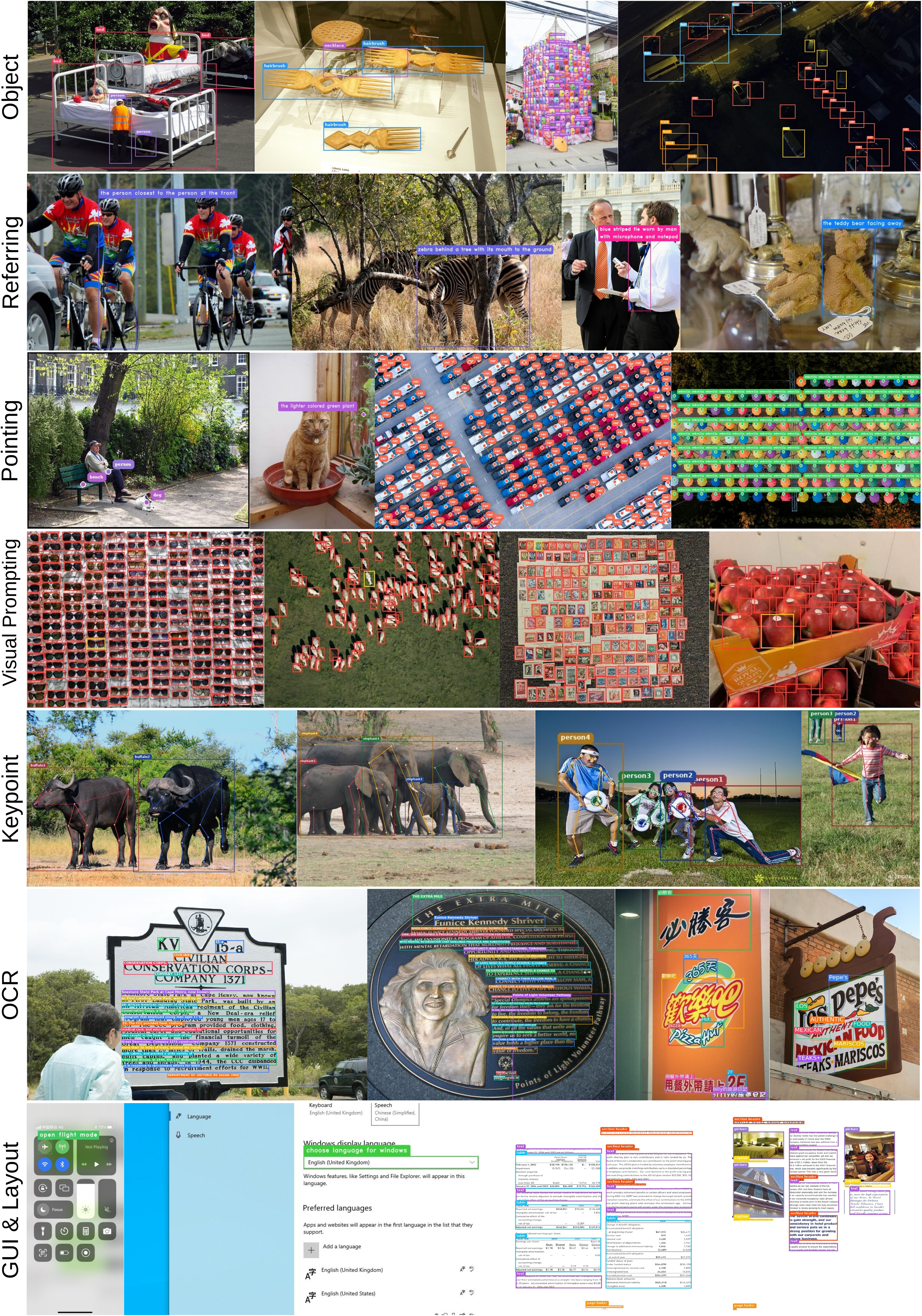}
    \vspace{-3pt}
    \caption{
    Additional qualitative results of structured visual understanding tasks.
    }
    \label{fig:additional-structured-visual-understanding}
\end{figure}

\begin{figure}[t]
    \centering
    \includegraphics[width=0.93\linewidth]{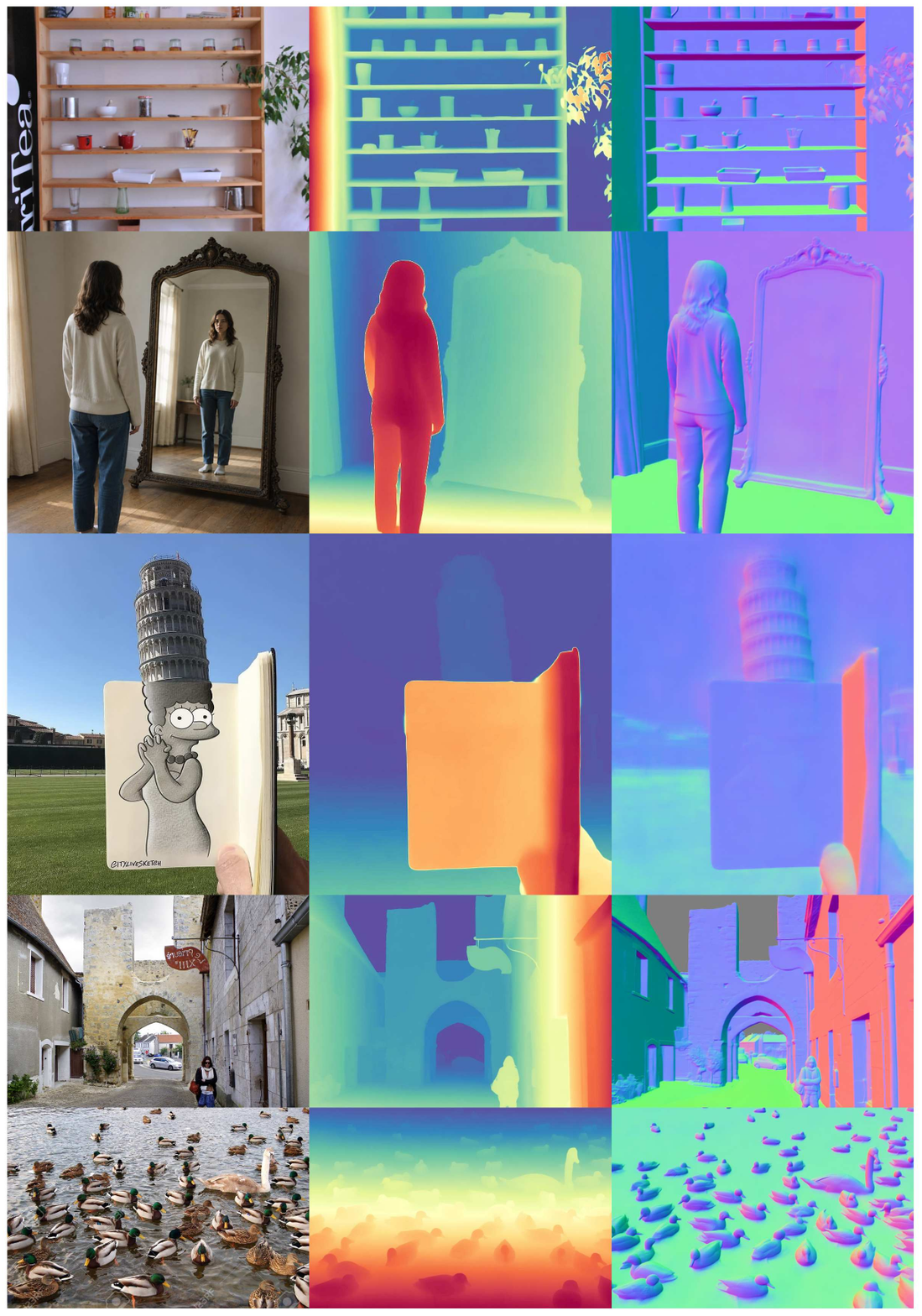}
    \vspace{-3pt}
    \caption{
    Additional qualitative results of dense geometric prediction tasks.
    }
    \label{fig:additional-dense-geometric-prediction}
\end{figure}

\begin{figure}[t]
    \centering
    \includegraphics[width=0.86\linewidth]{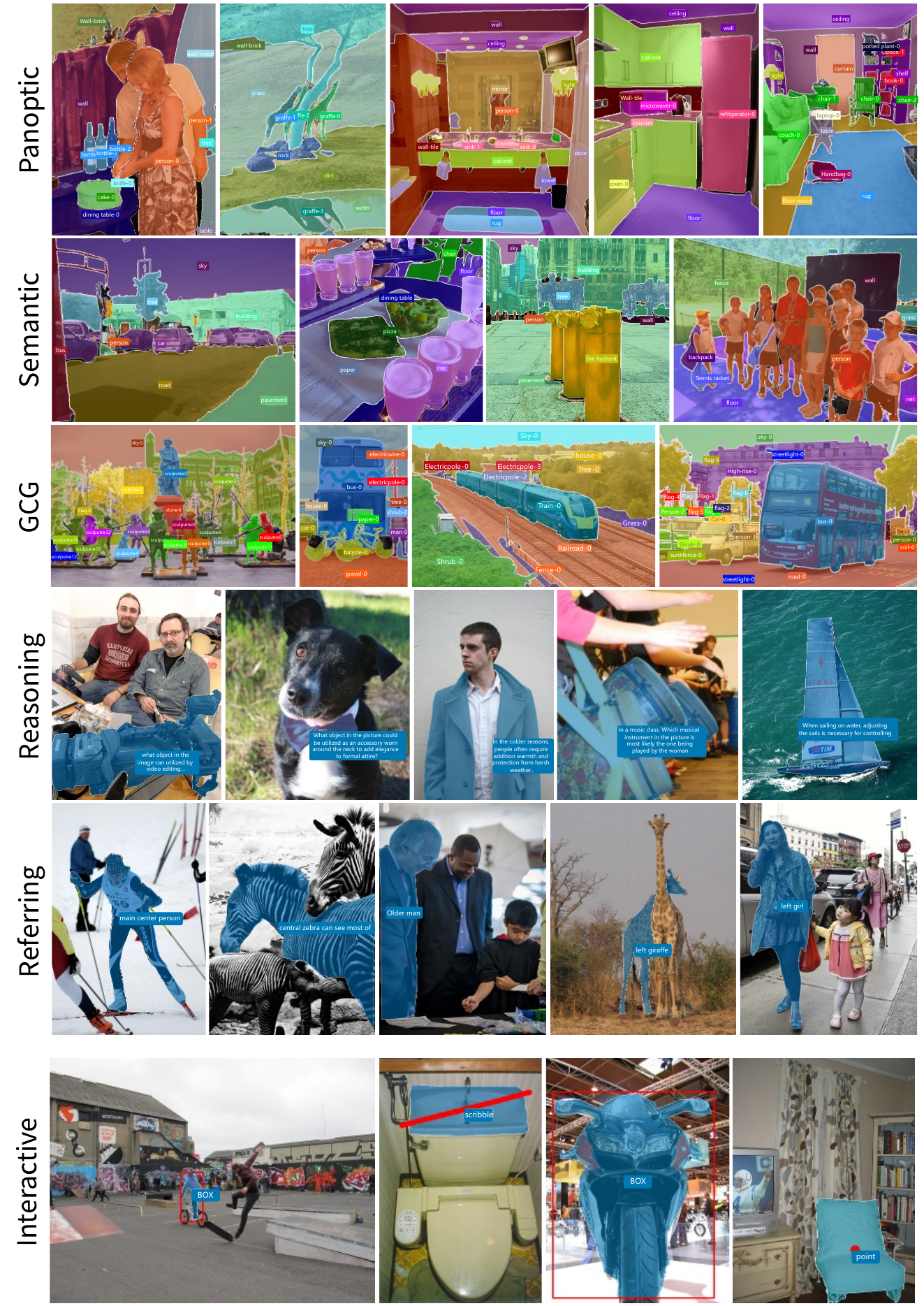}
    \vspace{-3pt}
    \caption{
    Additional qualitative results of segmentation tasks.
    }
    \label{fig:additional-segmentation}
\end{figure}

\FloatBarrier

\begin{figure}[!htbp]
  \centering
  \small
  \setlength{\tabcolsep}{0pt}
  \renewcommand{\arraystretch}{0.5}
  \resizebox{\textwidth}{!}{
    \begin{tabular}{
      m{0.025\linewidth}
      m{0.20\linewidth}
      m{0.28\linewidth}
      @{\hspace{2pt}} m{0.20\linewidth}
      m{0.28\linewidth}
    }
      &
      \multicolumn{1}{c}{Input Views} &
      \multicolumn{1}{c}{Reconstruction Output} &
      \multicolumn{1}{c}{Input Views} &
      \multicolumn{1}{c}{Reconstruction Output} \\
      \noalign{\vskip 4pt} 

      \rotatebox{90}{Indoor} &
      \includegraphics[width=\linewidth]{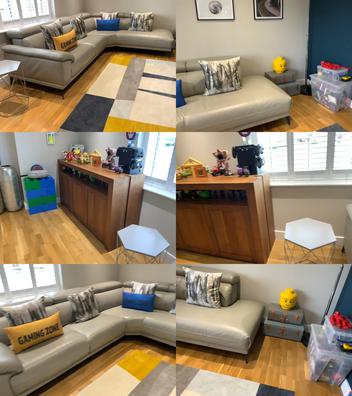} &
      \includegraphics[width=\linewidth]{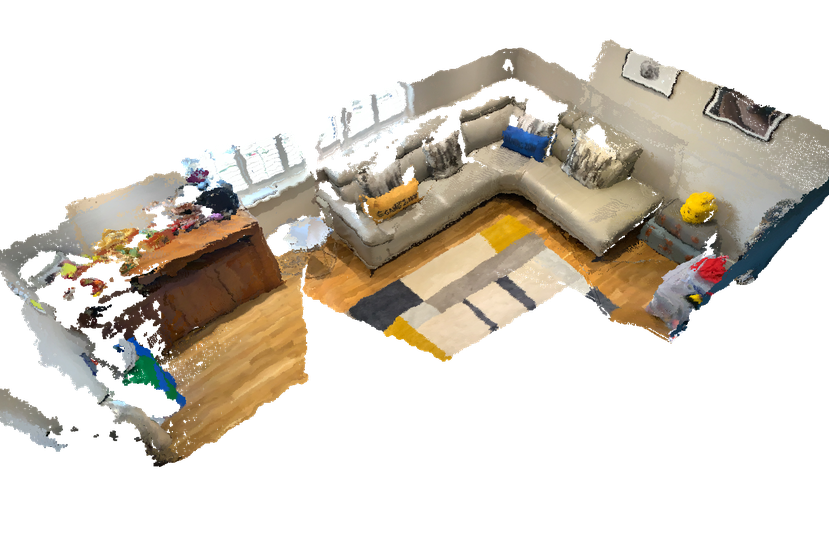} &
      \includegraphics[width=\linewidth]{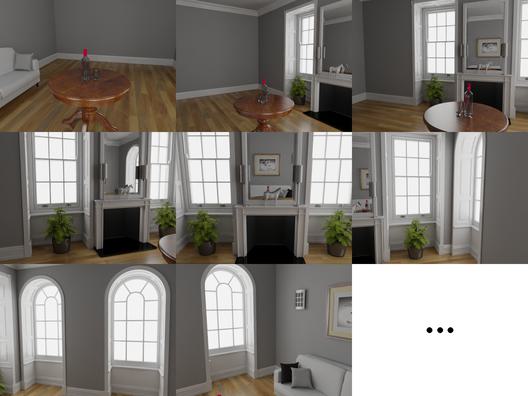} &
      \includegraphics[width=\linewidth]{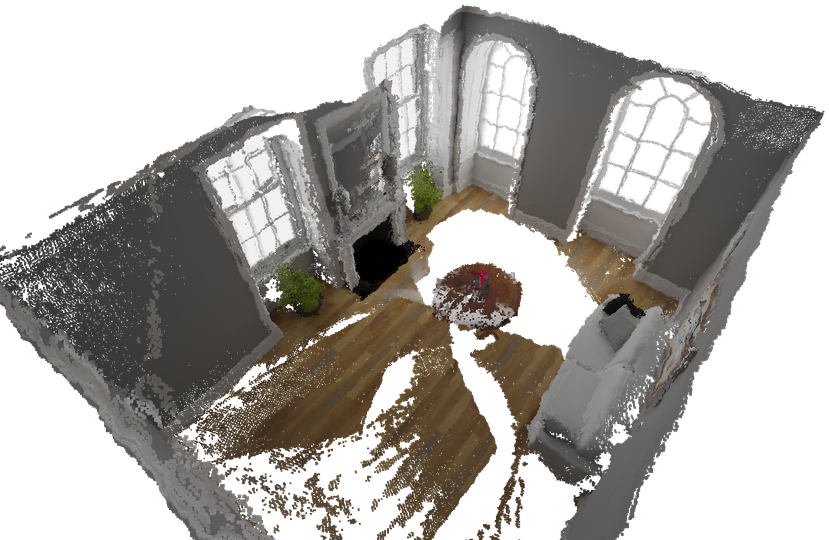} \\

      \rotatebox{90}{Outdoor} &
      \includegraphics[width=\linewidth]{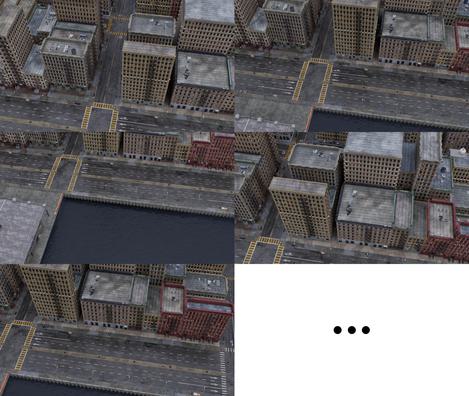} &
      \includegraphics[width=\linewidth]{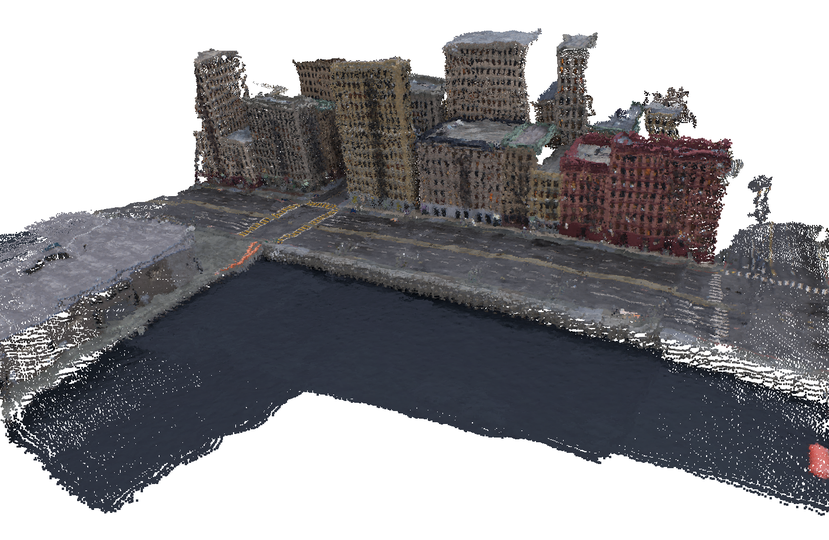} &
      \includegraphics[width=\linewidth]{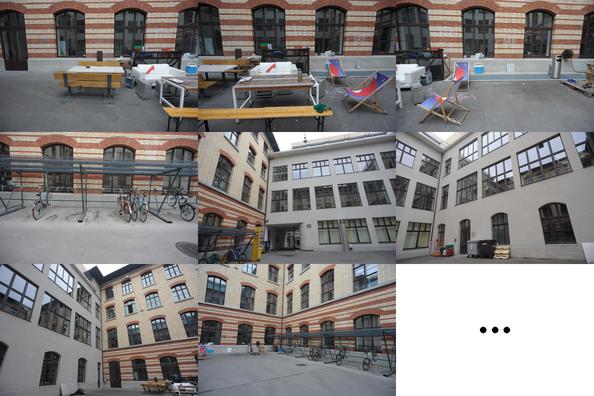} &
      \includegraphics[width=\linewidth]{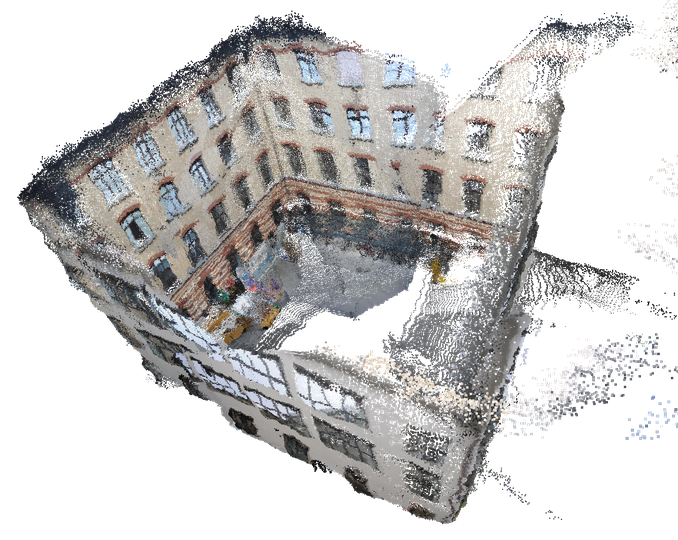} \\

      \rotatebox{90}{Object-Centric} &
      \includegraphics[width=\linewidth]{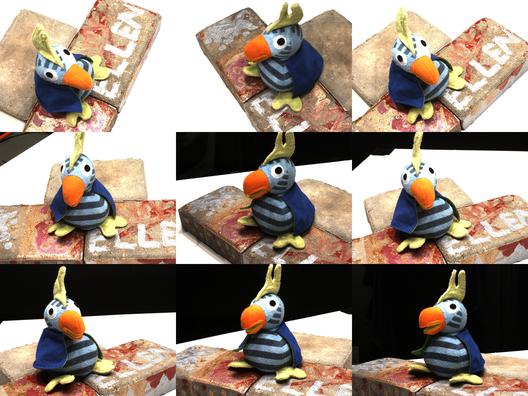} &
      \includegraphics[width=\linewidth]{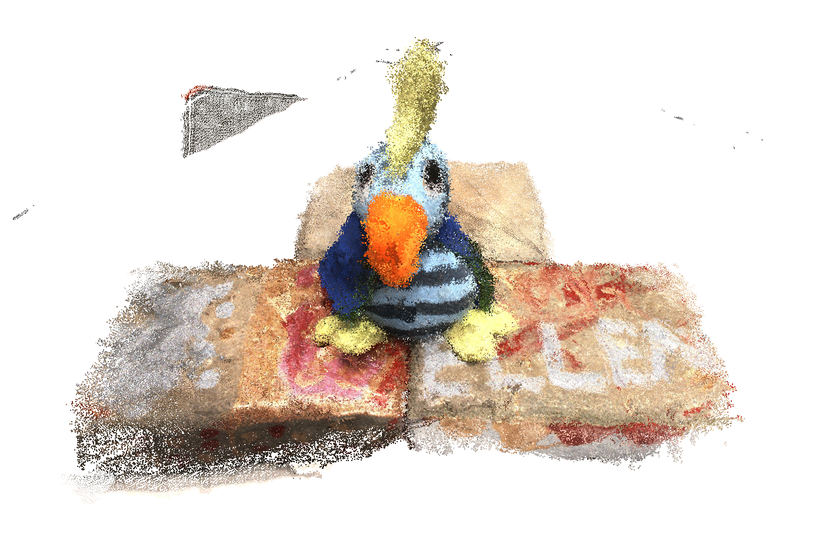} &
      \includegraphics[width=\linewidth]{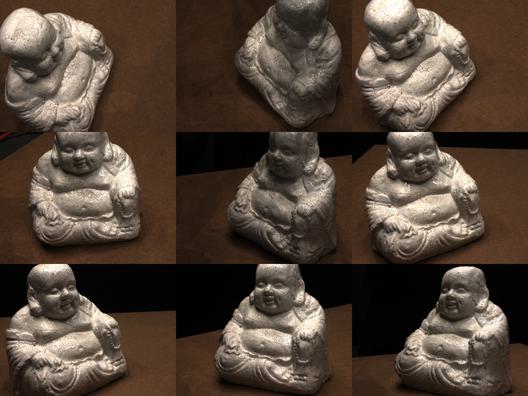} &
      \includegraphics[width=\linewidth]{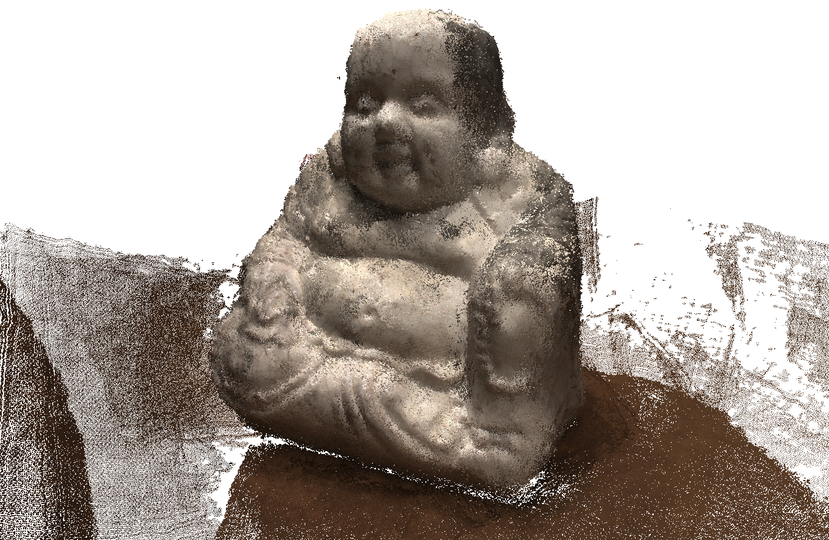} \\

      \rotatebox{90}{Gaming Env. (OOD)} &
      \includegraphics[width=\linewidth]{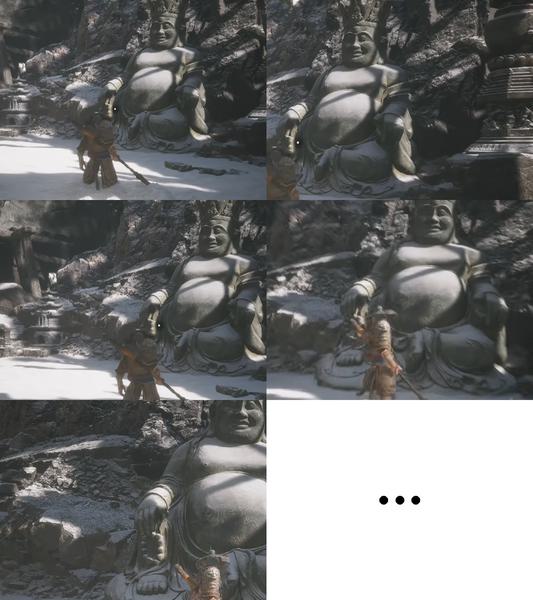} &
      \includegraphics[width=\linewidth]{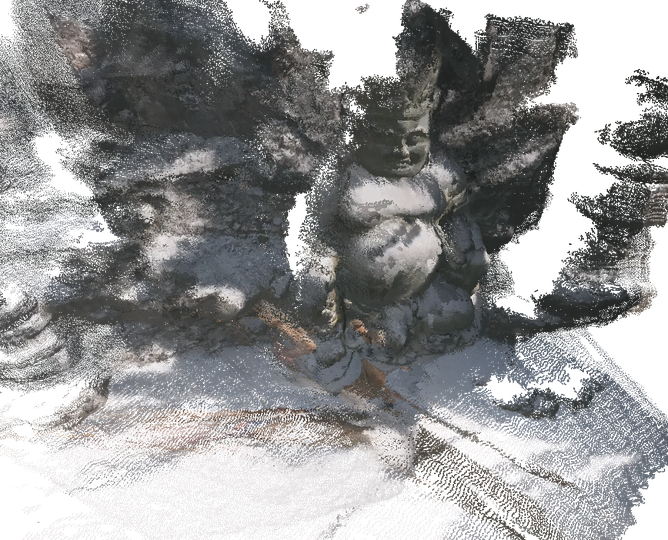} &
      \includegraphics[width=\linewidth]{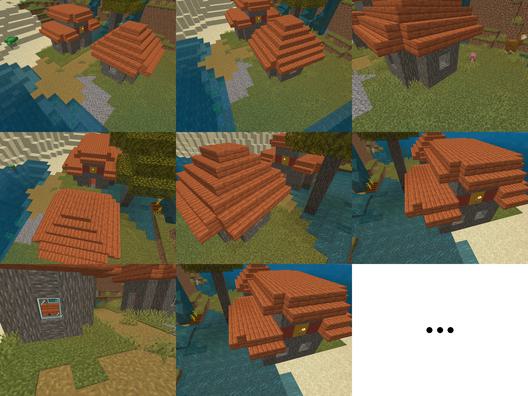} &
      \includegraphics[width=\linewidth]{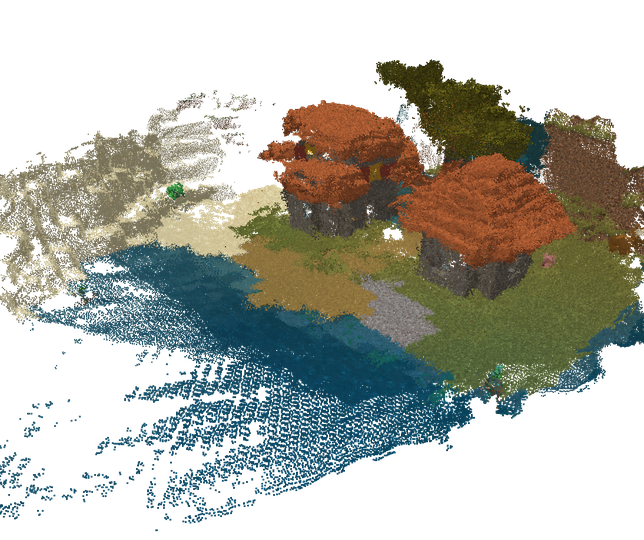}
    \end{tabular}
  }  
  \caption{Additional qualitative results of multi-view reconstruction.}
  \label{fig:additional-multiview-reconstruction}
\end{figure}

\section{Exploratory Results on Challenging and Language-Defined Vision Tasks}
\label{app:language_defined_boundary_cases}

This section presents exploratory qualitative results on challenging cases and language-defined vision tasks.
In our framework, a vision task is specified by language instructions that define both the task objective and the expected output format.
The examples here are exploratory tests beyond the main quantitative evaluation, covering composed dense prediction, specialized mask formats, fine-grained segmentation, and challenging dense geometry.

\begin{figure}[!t]
    \centering

    \includegraphics[height=0.23\textheight,keepaspectratio]{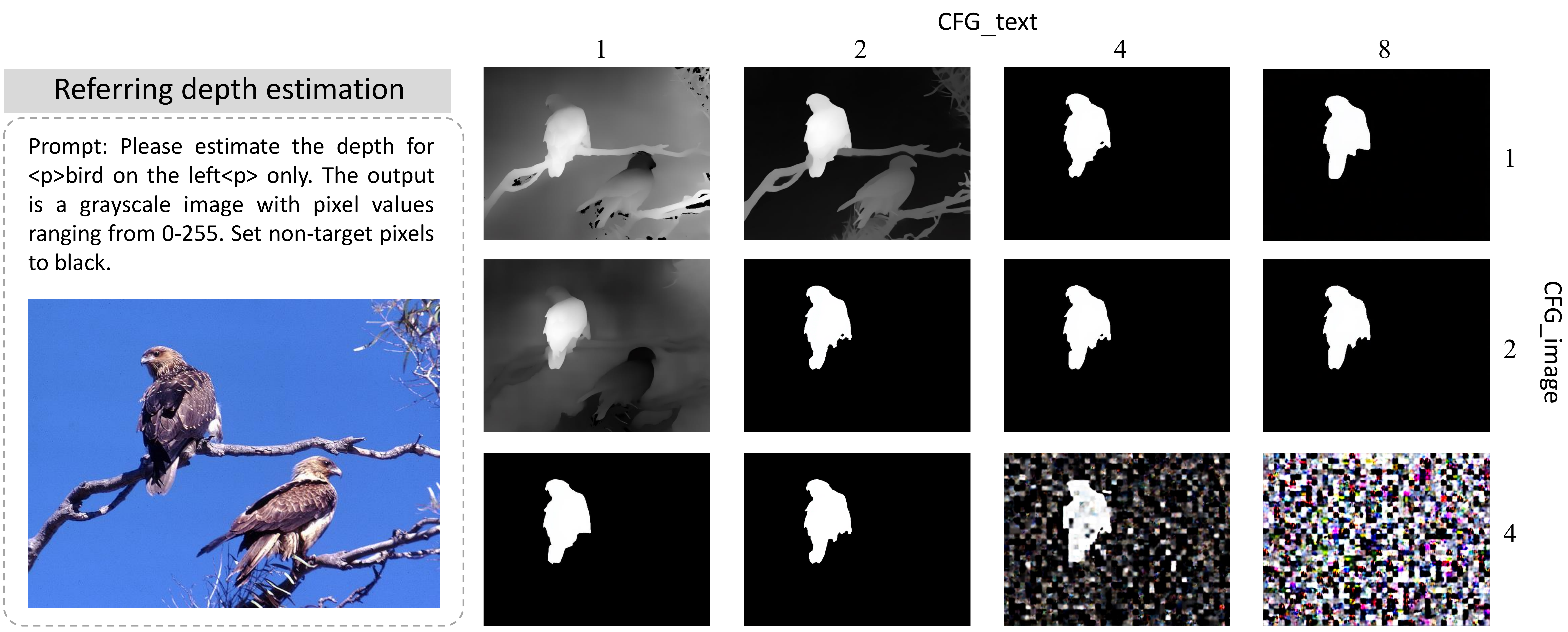}

    \includegraphics[height=0.23\textheight,keepaspectratio]{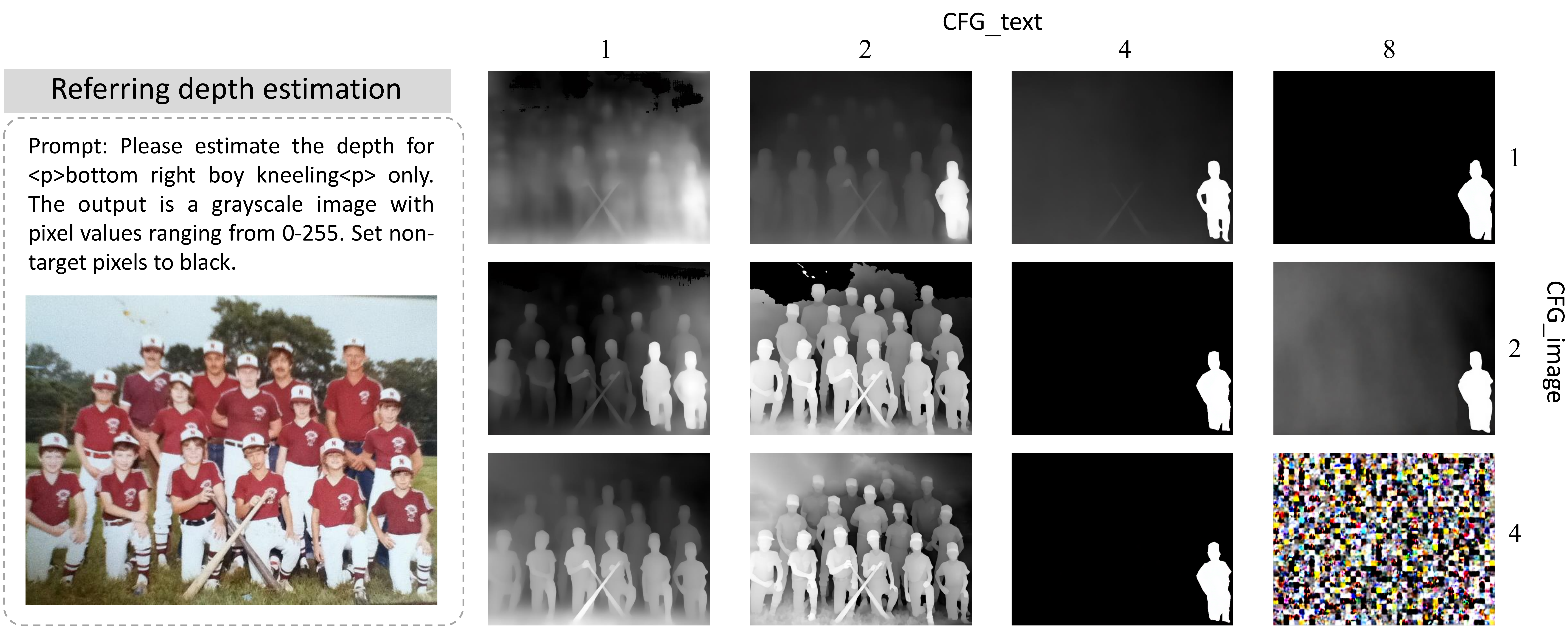}

    \includegraphics[height=0.23\textheight,keepaspectratio]{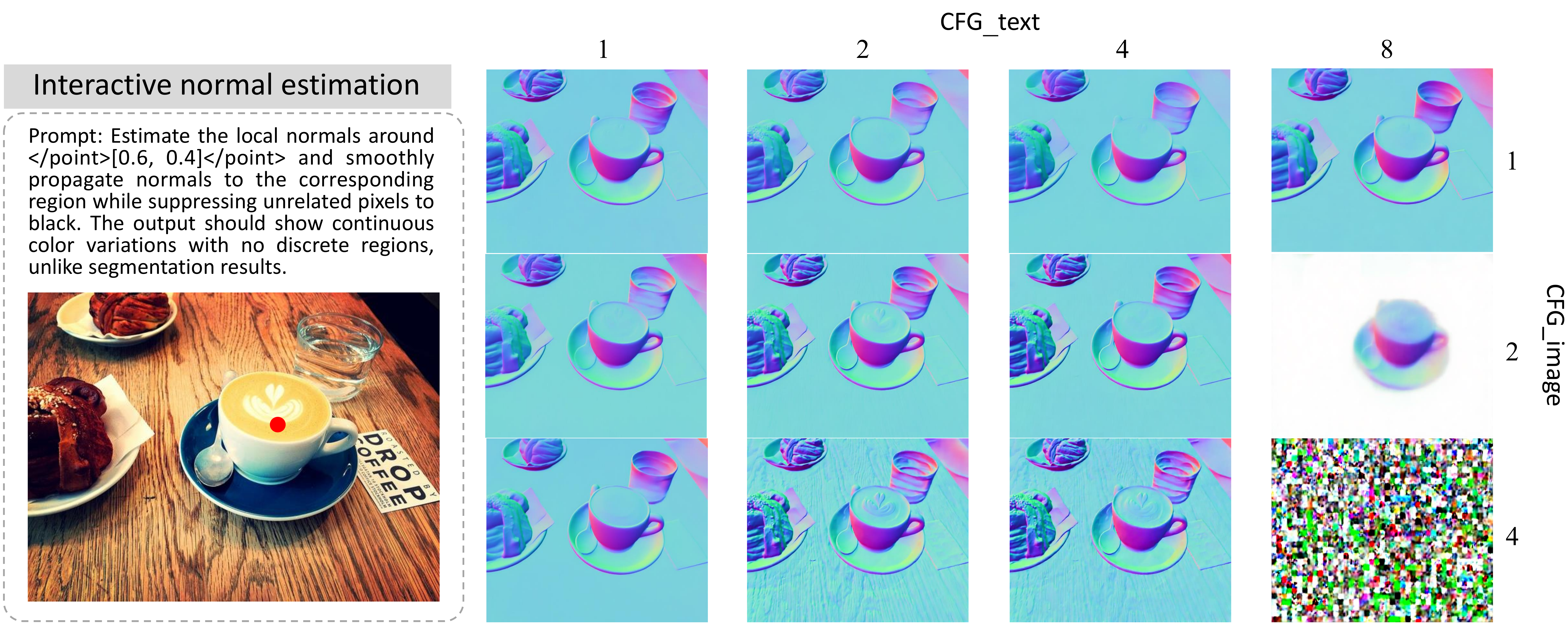}

    \includegraphics[height=0.23\textheight,keepaspectratio]{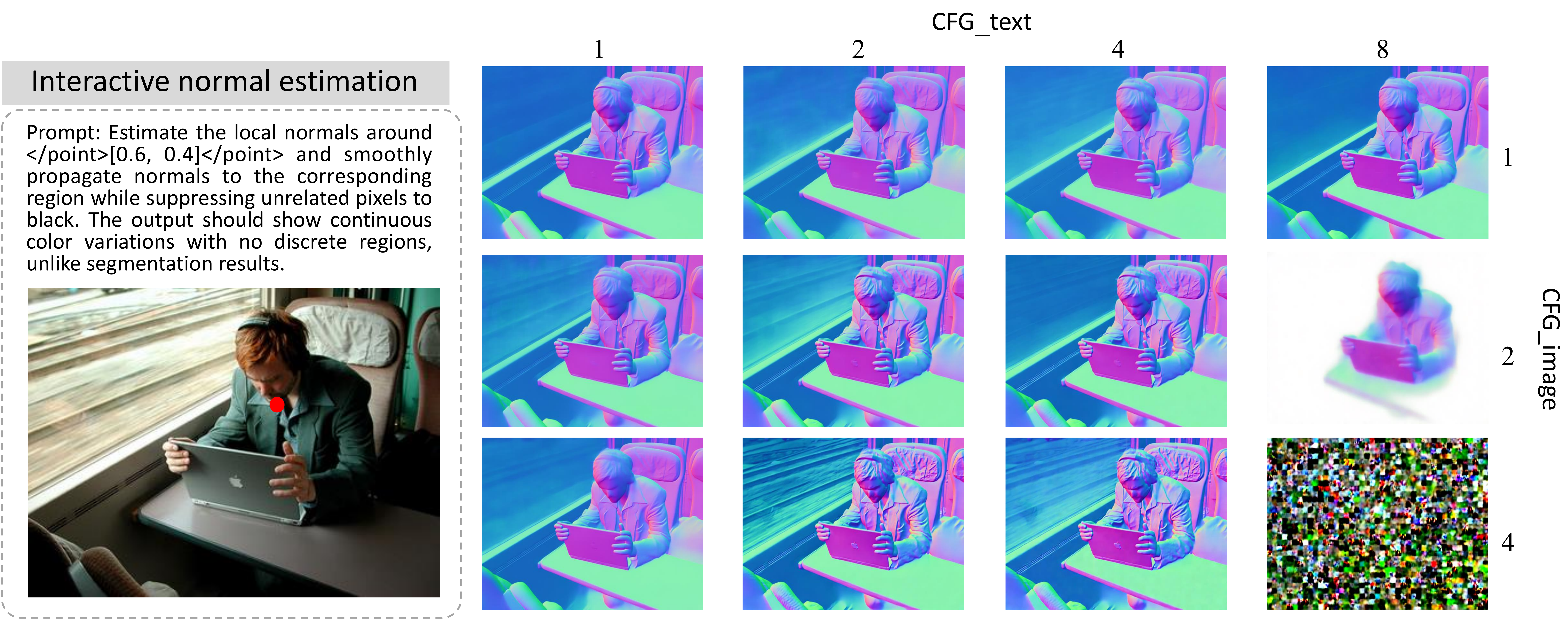}

    \caption{
    Qualitative results of composed dense prediction under different CFG strengths.
    With the input image and instruction fixed, some text-image CFG configurations produce outputs consistent with the composed visual task, while others bias the result toward depth-like, normal-like, or segmentation-like predictions.}
    \label{fig:appendix-cfg-strength}
\end{figure}

\begin{figure}[t]
    \centering

    \includegraphics[width=\linewidth,keepaspectratio]{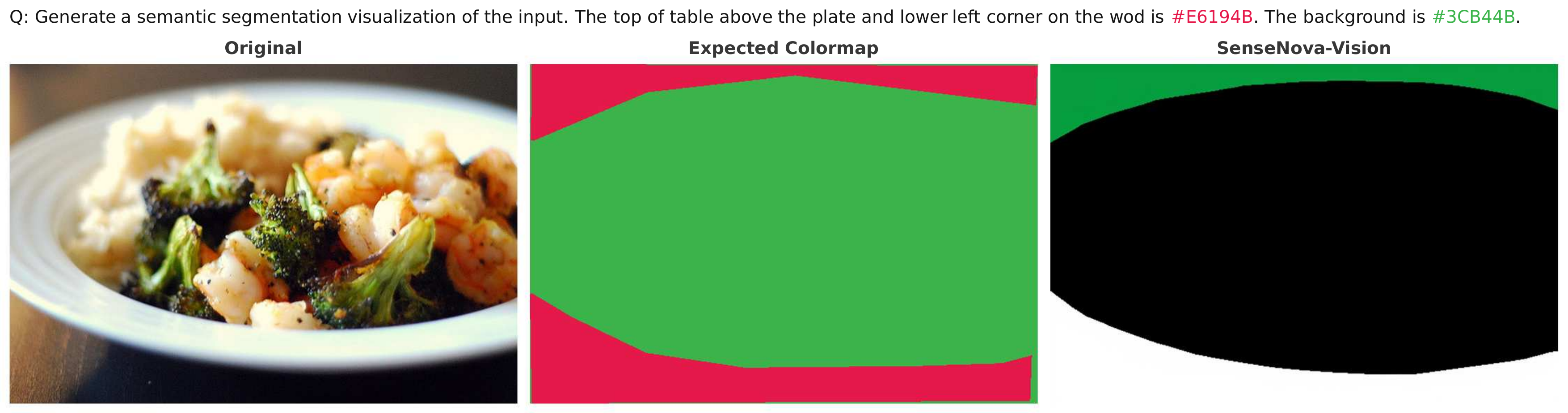}
    \vspace{6pt}

    \includegraphics[width=\linewidth,keepaspectratio]{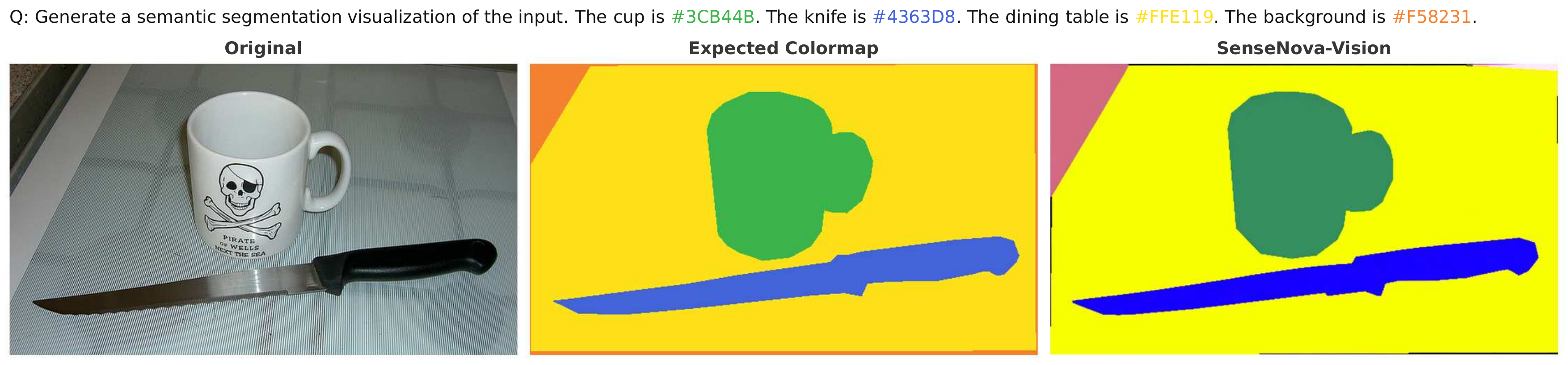}
    \vspace{6pt}

    \includegraphics[width=\linewidth,keepaspectratio]{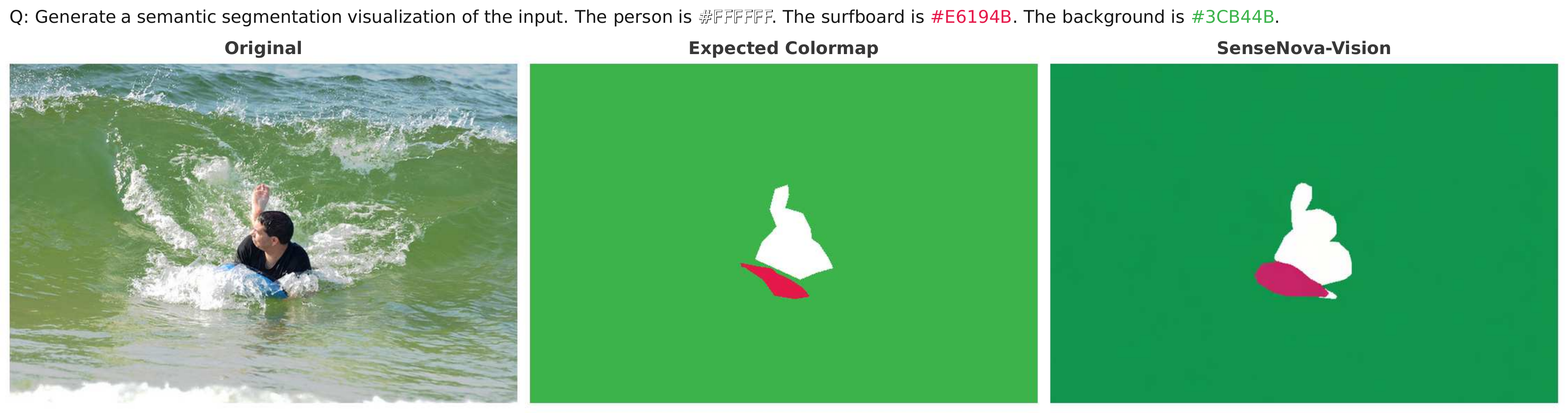}
    \vspace{6pt}

    \caption{
    Qualitative results of hexadecimal-color segmentation.
    These examples test whether the model can jointly satisfy semantic segmentation and language-specified color formatting.
    Although most target regions are correctly localized, the generated masks may deviate from the requested hexadecimal colors in their exact RGB values or assign incorrect colors to regions.
    }
    \label{fig:appendix-hex-color-segmentation}
\end{figure}

\begin{figure}[t]
    \centering
    \includegraphics[width=\linewidth,keepaspectratio]{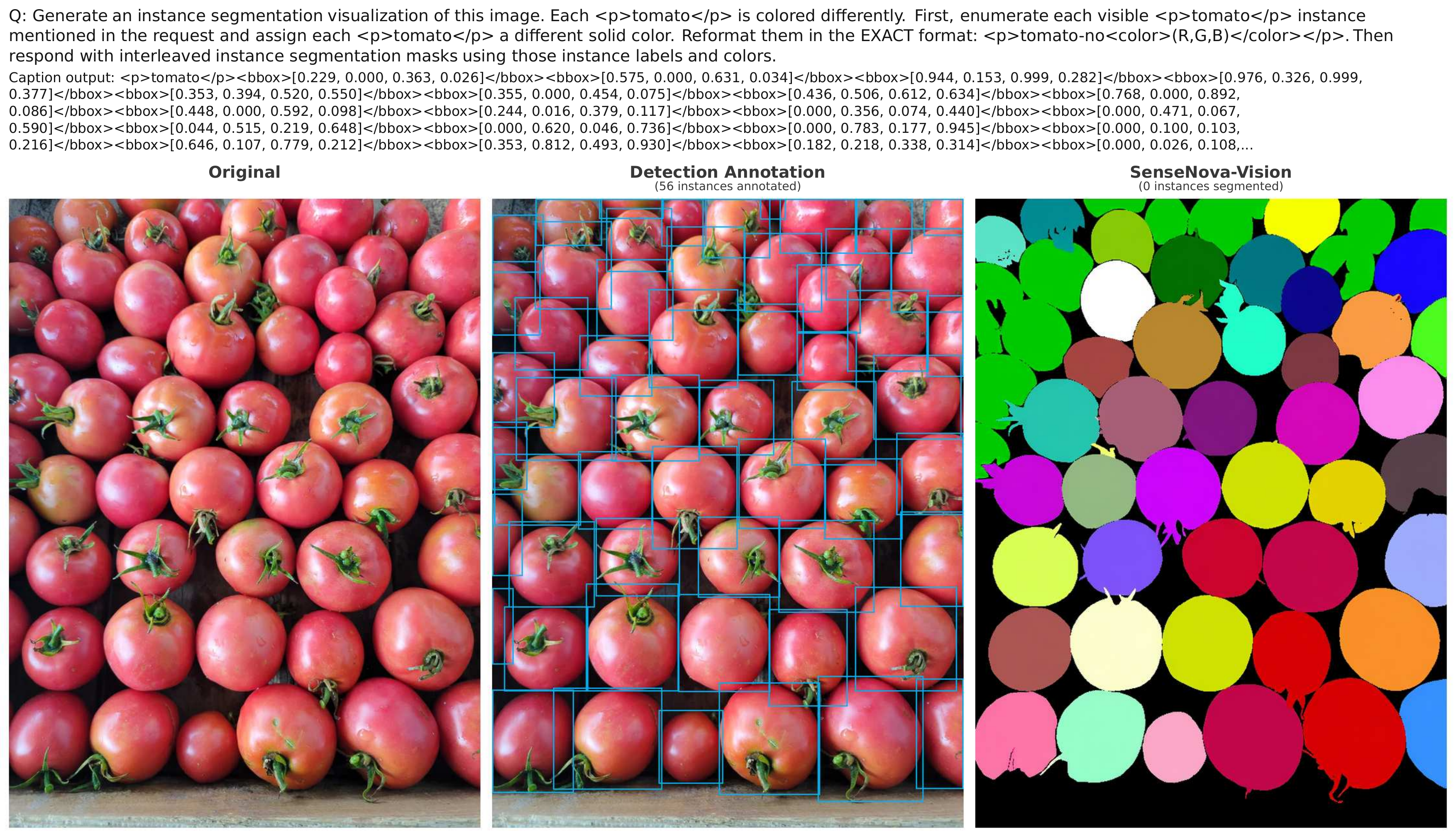}
    \vspace{-2pt}

    \includegraphics[width=\linewidth,keepaspectratio]{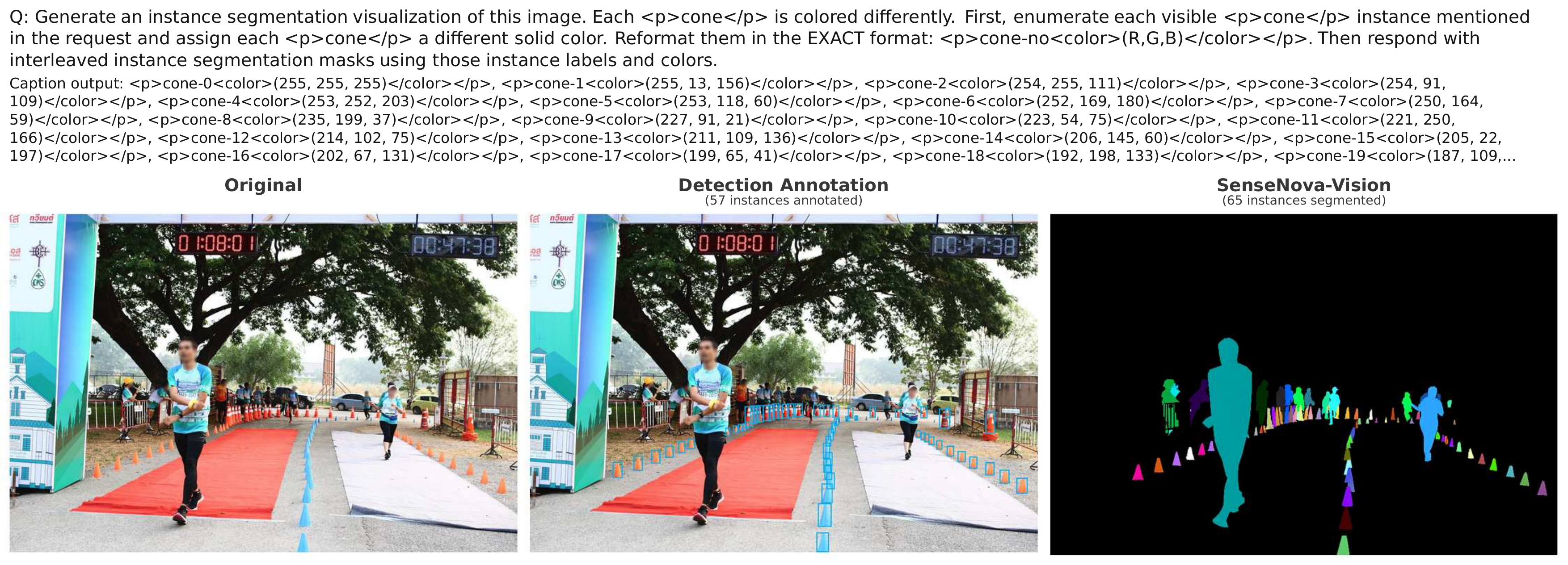}

    \includegraphics[width=0.77\linewidth,keepaspectratio]{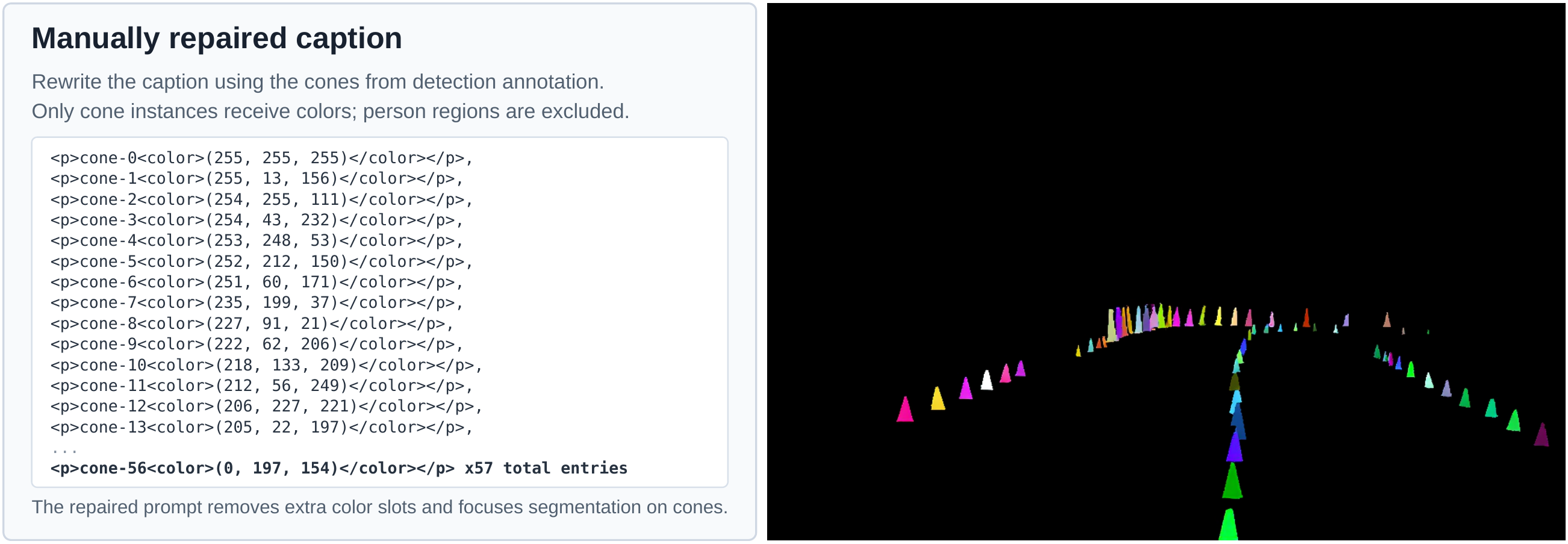}

    \caption{
    Qualitative results of dense instance segmentation.
    These examples require the model to enumerate each instance with a color code in structured text and then render the corresponding color-coded masks.
    The results show that structured text is critical for mask generation: the model may output detection-style records instead of instance color codes, or over-predict textual instances and include non-target regions in the masks; after manually repairing the instance list to include only the annotated target instances, the mask output becomes more focused and accurate.
    }
    \label{fig:appendix-dense-instance-segmentation}
\end{figure}

\begin{figure}[t]
    \centering
    \includegraphics[width=\linewidth,keepaspectratio]{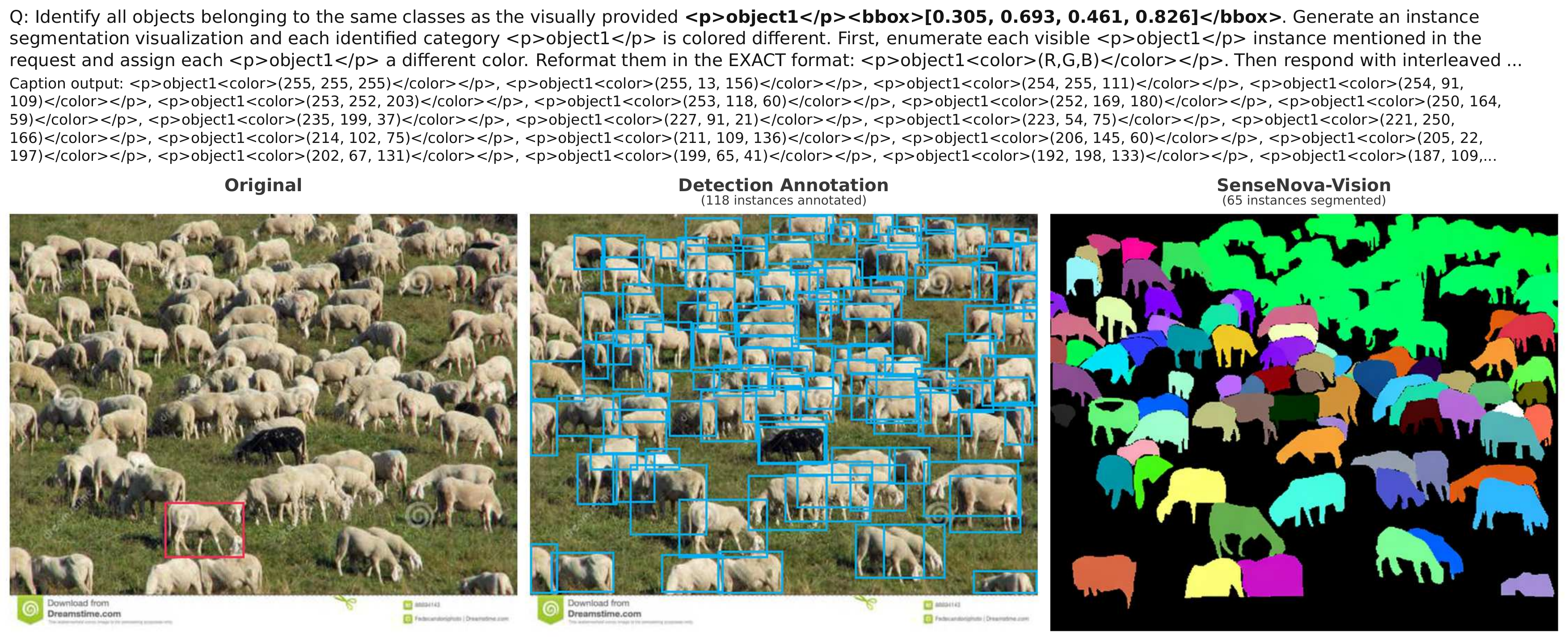}
    \vspace{6pt}

    \includegraphics[width=0.8\linewidth,keepaspectratio]{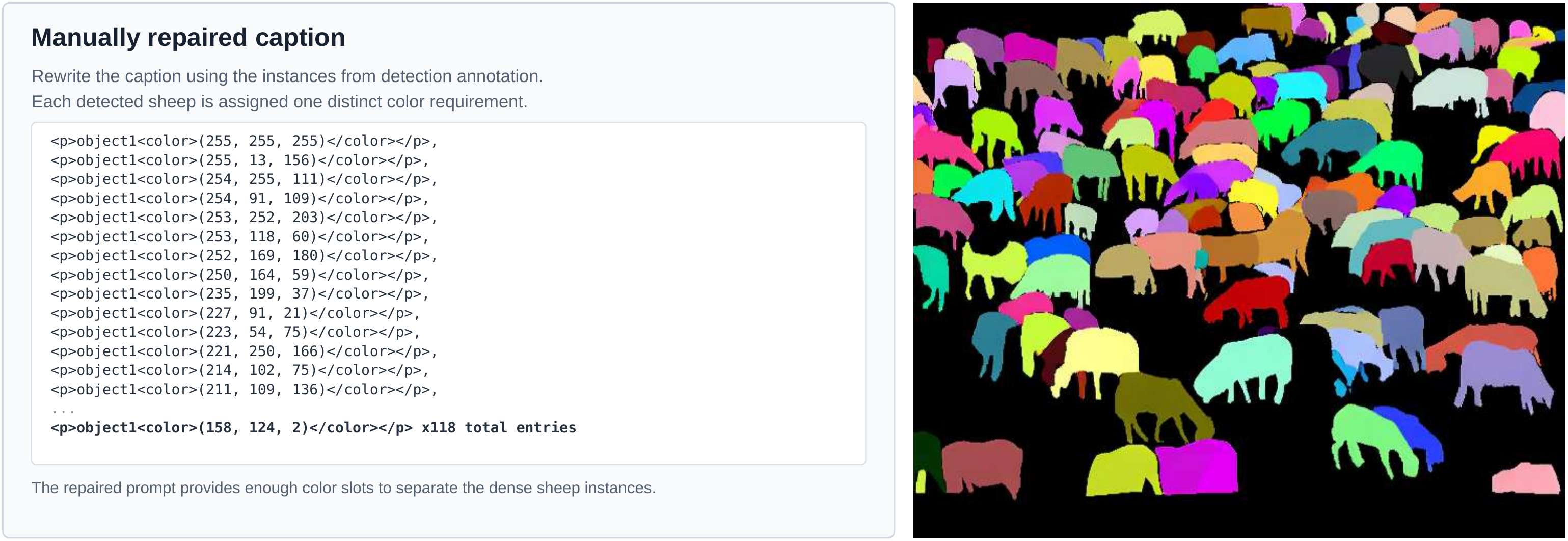}

    \caption{
    Qualitative results of Visual Grounded (VGD) segmentation.
    The visual prompt specifies the target category, and the model is expected to segment all same-category instances.
    When the generated caption provides too few instance color slots, dense objects may be merged in the mask output; manually repairing the caption with sufficient color entries improves instance-level separation.
    }
    \label{fig:appendix-vgd-segmentation}
\end{figure}

\begin{figure}[t]
    \centering
    \includegraphics[width=0.76\linewidth]{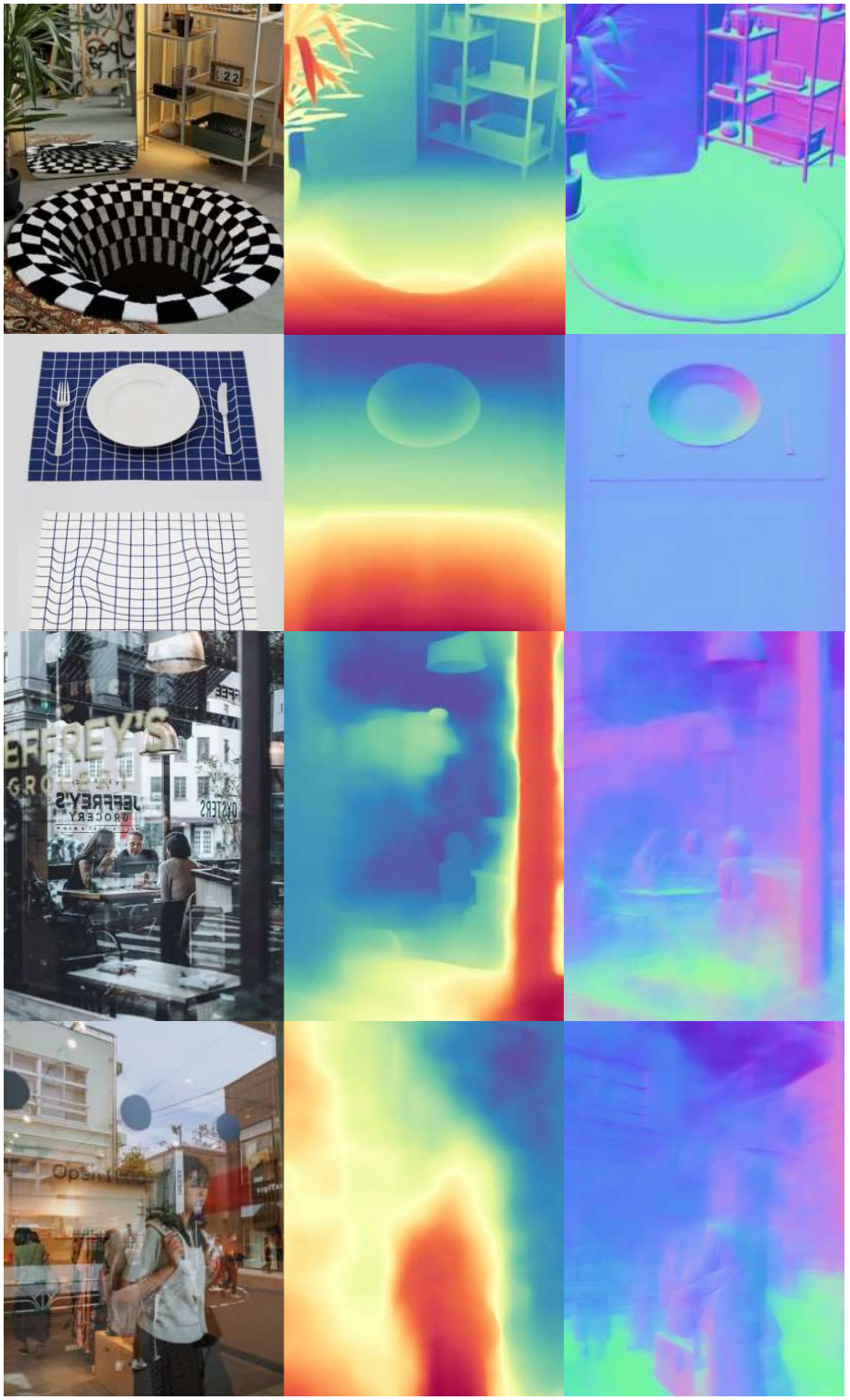}
    \vspace{-3pt}
    \caption{
    Qualitative results of dense geometric prediction.
    The examples cover visual-illusion patterns and transparent or reflective surfaces, where texture cues, reflections, and physical geometry can conflict.
    These cases show that depth or normal predictions may follow misleading appearance cues or become ambiguous around reflective layers and distorted patterns.
    }
    \label{fig:appendix-dense-geometry-challenging-cases}
\end{figure}

Figure~\ref{fig:appendix-cfg-strength} shows composed depth and normal prediction under different classifier-free guidance (CFG) settings.
Figures~\ref{fig:appendix-hex-color-segmentation} and~\ref{fig:appendix-dense-instance-segmentation} show examples with more specialized output requirements, including hexadecimal-color segmentation and dense instance segmentation with structured text-mask outputs.
Figure~\ref{fig:appendix-vgd-segmentation} presents VGD segmentation results under fine-grained object queries.
Figure~\ref{fig:appendix-dense-geometry-challenging-cases} presents dense geometric prediction results on challenging scenes, including visually ambiguous patterns and transparent or refractive surfaces.

\end{document}